%% file: arxiv.tex
\pdfoutput=1
\documentclass[10pt,twocolumn,letterpaper]{article}

\usepackage[pagenumbers]{cvpr}

\input{preamble}

\definecolor{cvprblue}{rgb}{0.21,0.49,0.74}
\definecolor{tableblue}{HTML}{E4F6FF}
\usepackage[pagebackref,breaklinks,colorlinks,allcolors=cvprblue]{hyperref}

\title{Beyond Statistical Priors: Architecture-Agnostic Data-Free Knowledge Distillation via Unified Explicit Semantics}

\author{
Xuewan He \quad Tong Chu \quad Zihan Cheng \quad Yuchen Su \quad Qianxin Xia\\
Guoming Lu \quad Jielei Wang\thanks{Corresponding authors.} \quad Wen Li\footnotemark[1]\\
University of Electronic Science and Technology of China, Chengdu, China\\
{\tt\small hershel@std.uestc.edu.cn, jlwang@uestc.edu.cn, liwenbnu@gmail.com}
}

\begin{document}
\maketitle
\input{sec/0_abstract}
\input{sec/1_intro}
\input{sec/2_related}
\input{sec/3_method}
\input{sec/4_experiments}
\input{sec/5_conclusion}

{
    \small
    \bibliographystyle{ieeenat_fullname}
    \bibliography{main}
}

\input{sec/X_suppl}

\end{document}

%% file: preamble.tex
\usepackage{microtype}
\usepackage{multirow}
\usepackage{array}
\usepackage{tabularx}
\usepackage{adjustbox}
\usepackage{diagbox}
\usepackage{pifont}
\usepackage{colortbl}
\usepackage{tikz}
\usetikzlibrary{arrows.meta,decorations.pathmorphing}

\newcommand{\cmark}{\ding{51}}
\newcommand{\xmark}{\ding{55}}
\newcommand{\gmark}{\textcolor{black!45}{\xmark}}

\newtheorem{proposition}{Proposition}[section]
\newtheorem{lemma}[proposition]{Lemma}
\newtheorem{theorem}[proposition]{Theorem}
\makeatletter
\def\@begintheorem#1#2{\trivlist\item[\hskip\labelsep{\bfseries #1\ #2.}]\itshape}
\def\@opargbegintheorem#1#2#3{\trivlist\item[\hskip\labelsep{\bfseries #1\ #2}]{\bfseries (#3).}\ \itshape}
\makeatother

\makeatletter
\renewcommand\paragraph{\@startsection{paragraph}{4}{\z@}{1ex \@plus .3ex \@minus .2ex}{-1em}{\normalfont\normalsize\bfseries}}
\makeatother

\newif\ifmergedanalysisfig
\mergedanalysisfigtrue

%% file: sec/0_abstract.tex
\begin{abstract}
Data-Free Knowledge Distillation (DFKD) transfers knowledge from a pretrained teacher model to a compact student model by synthesizing semantically informative data, eliminating the need for access to the original training dataset. 
Existing DFKD methods rely heavily on architecture-specific statistical priors (e.g., Batch Normalization statistics) to guide data synthesis;
however, such architecture-dependent priors are often absent in modern architectures such as Vision Transformers (ViTs), resulting in degraded semantic quality of the synthesized data and consequently catastrophic performance degradation. 
In this paper, we propose \mbox{\emph{UniDFKD}}, a unified data-free knowledge distillation framework that replaces architecture-specific statistics with explicit, architecture-agnostic semantic priors. UniDFKD governs the entire synthesis--distillation pipeline along three dimensions: (1) Categorical Semantic Conditioning (CSC) defines \emph{what} to synthesize by persistently modulating the generator with language-derived embeddings to capture semantic diversity; (2) Spatial Semantic Anchoring (SSA) regularizes \emph{where} evidence belongs by anchoring the teacher's spatial attributions to a Gaussian prior; and (3) Spatial Semantic Distillation (SSD) controls \emph{how} knowledge is transferred by explicitly aligning teacher--student spatial evidence alongside predictions.
Extensive experiments across CNNs and ViTs demonstrate that UniDFKD sets a new state of the art. It outperforms the strongest baseline NAYER by average absolute margins of 17.23 and 24.49 in top-1 accuracy under homogeneous and heterogeneous settings, respectively.
\end{abstract}

\begin{figure}[t]
\centering
\includegraphics[width=1.\columnwidth]{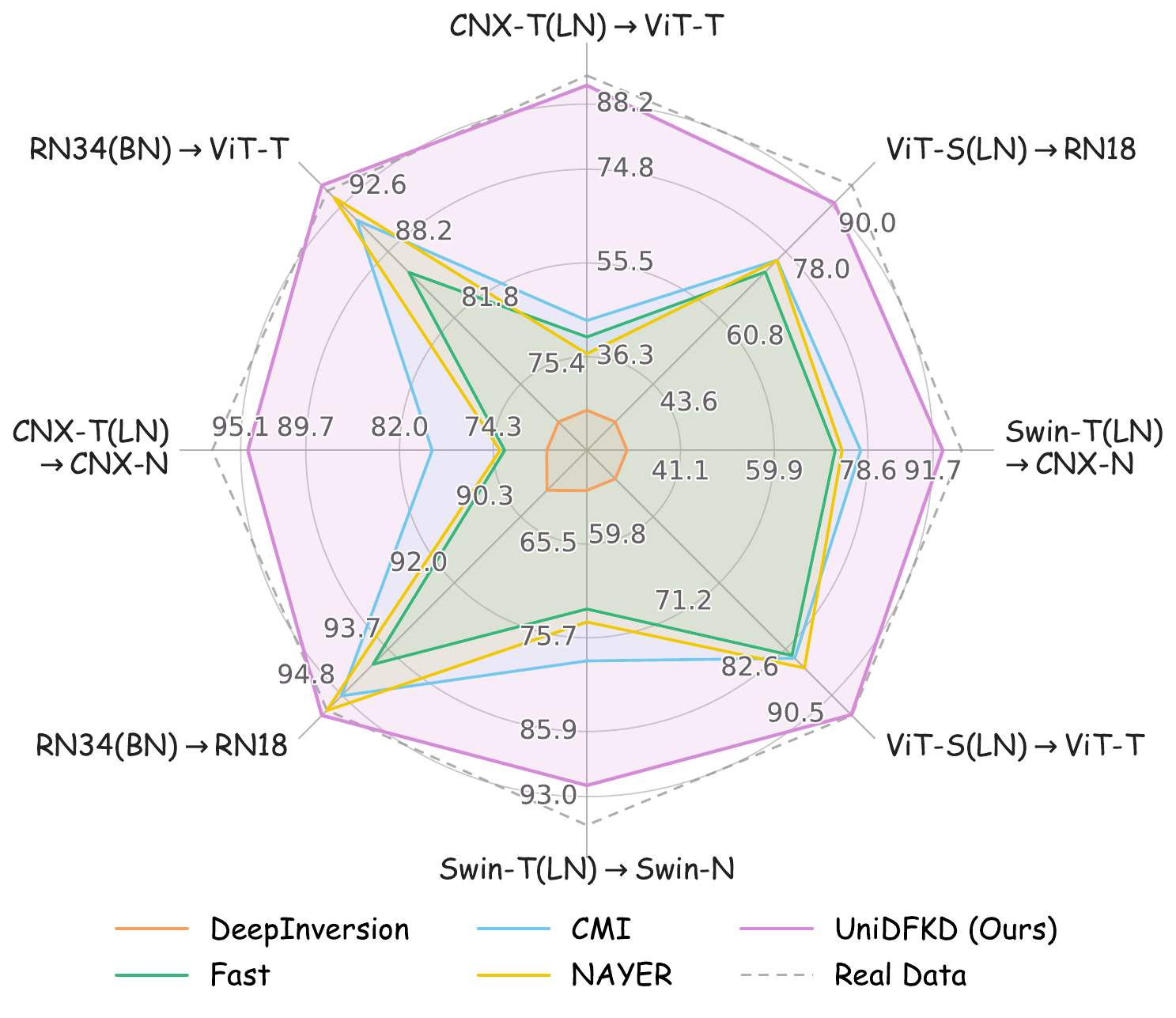}
\caption{
Student accuracy (\%) on CIFAR-10 across eight homogeneous and heterogeneous teacher--student settings.
BN/LN denotes the teacher normalization, and the dashed polygon marks students trained on real data.
UniDFKD consistently delivers stronger performance across diverse transfer settings and remains the closest to the real-data reference.
}
\label{fig:c10_distillation_radar}
\vspace{-5pt}
\end{figure}

%% file: sec/1_intro.tex
\section{Introduction}
\begin{figure*}[t]
    \centering
    \begin{subfigure}[b]{0.57\linewidth}
        \centering
        {\sffamily\scriptsize\makebox[.25\linewidth]{Block 2}\makebox[.25\linewidth]{Block 6}\makebox[.25\linewidth]{Block 12}\makebox[.25\linewidth]{Block 15}}\\[-1pt]
        \includegraphics[width=\linewidth,trim={0 4bp 0 11.6bp},clip]{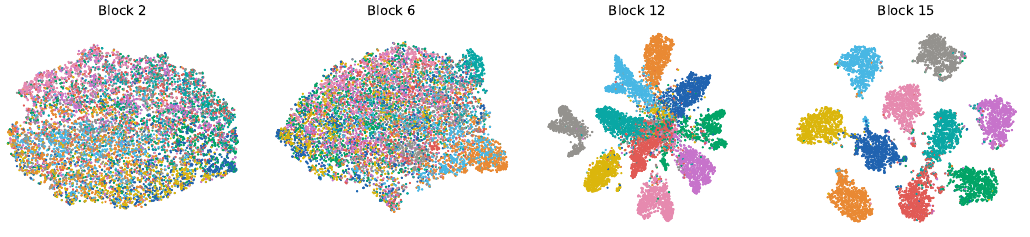}
        \caption{t-SNE of RN34 BN statistics.}
    \label{fig:resnet_bn_statistics_tsne}
    \end{subfigure}\hfill
    \begin{minipage}[b]{0.425\linewidth}
        \centering
        \includegraphics[width=\linewidth]{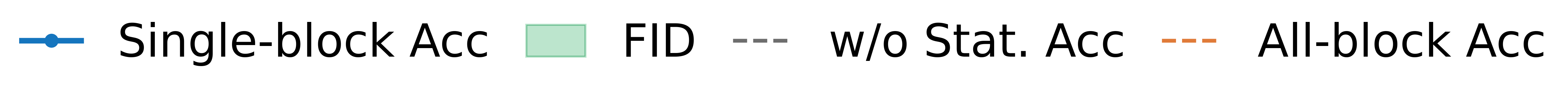}
        \\
        \begin{subfigure}[b]{0.495\linewidth}
            \centering
            \includegraphics[width=\linewidth]{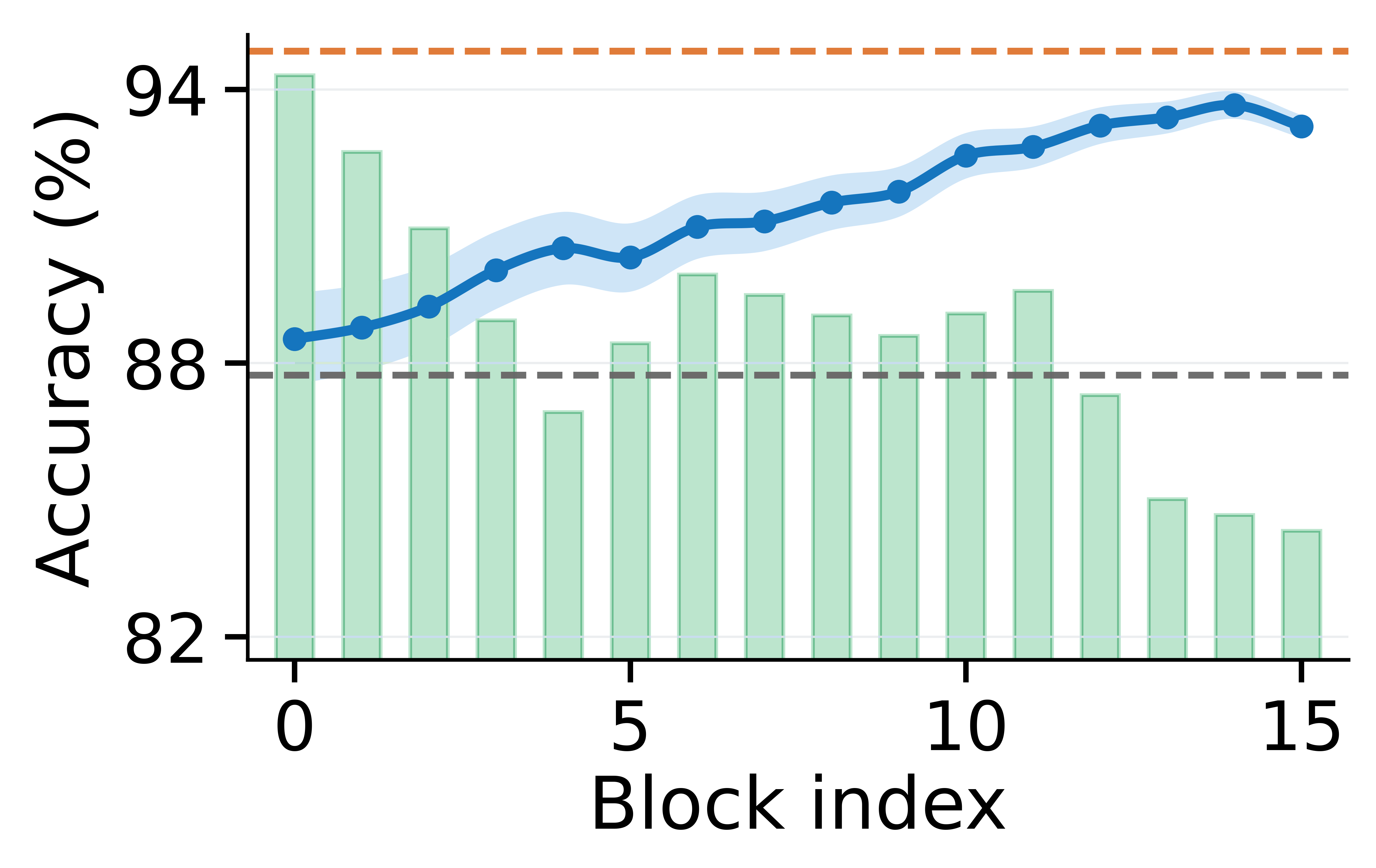}
            \caption{RN34$\to$RN18.}
        \label{fig:cmi_bn_layerwise_ablation_cnn}
        \end{subfigure}\hfill
        \begin{subfigure}[b]{0.495\linewidth}
            \centering
            \includegraphics[width=\linewidth]{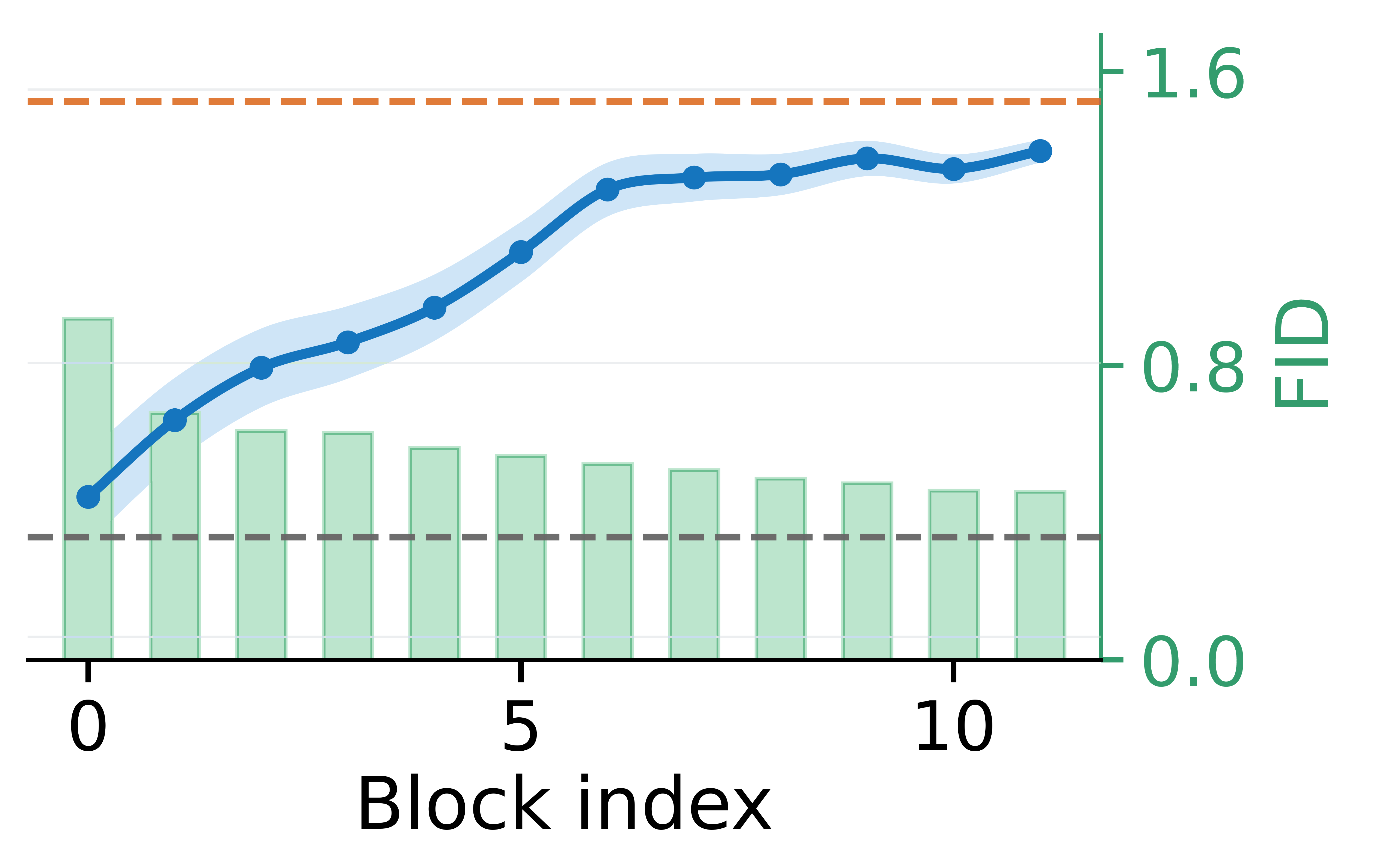}
            \caption{ViT-S$\to$ViT-T.}
         \label{fig:cmi_bn_layerwise_ablation_vit}
        \end{subfigure}
    \end{minipage}
    \caption{(a) t-SNE of BN statistics at the last block of each RN34 stage on CIFAR-10, each point representing a sample's concatenated channel-wise mean and standard deviation, colored by class. Different classes overlap in shallow blocks but form distinct clusters in deeper ones. (b, c) Single-block feature-statistic matching on CIFAR-10.
    Student acc. ($\uparrow$) and T-FID ($\downarrow$) are reported for analysis.}
    \label{fig:cmi_bn_layerwise_ablation}
    \vspace{-5pt}
\end{figure*}

Knowledge distillation is a cornerstone technique for deploying highly performant, large-scale vision models onto resource-constrained edge devices~\cite{hinton2015distilling,mansourian2025comprehensive}. In many real-world applications, however, the original training data used to optimize the teacher model is rendered inaccessible due to stringent privacy regulations, proprietary restrictions, or formidable transmission costs~\cite{chen2019data}. Data-Free Knowledge Distillation (DFKD) mitigates this bottleneck by synthesizing surrogate training samples directly from the pretrained teacher, effectively hallucinating the missing data distribution to facilitate downstream student training~\cite{liu2021data,liu2024small,chen2025cpnet}.

Existing DFKD methodologies rely heavily on the teacher's Batch Normalization (BN) statistics~\cite{ioffe2015batch} as a strong data prior to constrain the generator, since data synthesis in a high-dimensional space is severely underdetermined~\cite{yin2020dreaming}. Specifically, the running mean and variance stored within the teacher's BN layers serve as the \emph{de facto} surrogate for the inaccessible data distribution. By aligning the feature statistics of synthetic batches with these stored real-data statistics, inversion on BN-based CNNs can remarkably approach the accuracy of students trained on real data (cf. RN34$\to$RN18 and RN34$\to$ViT-T settings in Fig.~\ref{fig:c10_distillation_radar}).

Unfortunately, this reliance on BN statistics is merely an architectural accident. 
This critical prior simply vanishes as the field rapidly pivots toward BN-free architectures that rely on Layer Normalization (LN)~\cite{huang2023normalization}, such as Vision Transformer (ViT) and ConvNeXt.
Consequently, state-of-the-art DFKD methods fail to generalize to BN-free teachers, especially across heterogeneous architectures. 
Under LN-based settings (e.g., Swin-T$\to$Swin-N and CNX-T$\to$ViT-T), their performance falls substantially short of that achieved by students trained on real data. (Fig.~\ref{fig:c10_distillation_radar}).
Such a pronounced drop compels us to ask a fundamental question: \emph{what makes BN statistics effective for DFKD in the first place?}

To answer this question, we examine BN statistics block by block across RN and ViT teachers (Sec.~\ref{sec:motivation}).
Specifically, we constrain synthesis with the feature statistics of a single block at a time and evaluate the student distilled from the synthetic data (Figs.~\ref{fig:cmi_bn_layerwise_ablation_cnn} and~\ref{fig:cmi_bn_layerwise_ablation_vit}).
On both teachers, the gain concentrates in deep blocks. Matching only the deepest block nearly recovers the accuracy of matching all blocks, whereas matching shallow blocks improves little over unconstrained synthesis.
To understand this depth dependence, we visualize per-sample BN statistics at different depths (Fig.~\ref{fig:resnet_bn_statistics_tsne}).
Statistics of different classes overlap in shallow blocks and form distinct class clusters in deep blocks, in line with the shift from low-level textures to high-level semantics along network depth~\cite{gatys2016image,zhong2022fine}.
We therefore conjecture that \emph{the effectiveness of BN statistics stems from an implicit semantic prior in deep representations}.

This conjecture motivates a paradigm shift in DFKD: from implicitly matching architecture-specific statistics to enforcing explicit, architecture-agnostic semantic priors (Fig.~\ref{fig:Comparison}). 
To govern the full distillation pipeline, an effective semantic prior must systematically address three critical dimensions: \emph{what} semantic content to generate, \emph{where} to spatially localize it, and \emph{how} to transfer it to the student. 
Yet existing methods address these dimensions in isolation~\cite{tran2024nayer,tran2024muse,zhong2024semantics,yu2023data}, and no DFKD framework unifies them throughout synthesis and distillation.

To this end, we introduce \textbf{UniDFKD}, which injects explicit semantic priors into both stages to \emph{unify} DFKD across diverse architectures.
During synthesis, UniDFKD answers the \emph{what} via Categorical Semantic Conditioning (CSC), which leverages large language models (LLMs) to construct a diverse semantic bank, persistently modulating every block of the generator so that synthesized images preserve inter-class relations and rich intra-class diversity. To determine the \emph{where}, Spatial Semantic Anchoring (SSA) projects the teacher's deepest responses into a unified class-discriminative spatial attribution (CDSA) map and anchors its batch-level average to a center-biased Gaussian prior to match the spatial layout of real images.
Finally, to address the \emph{how}, Spatial Semantic Distillation (SSD) complements logit matching by aligning the per-sample CDSA maps between the teacher and student, so that the student inherits the spatial evidence behind the teacher's predictions.

Our contributions are three-fold:

\begin{itemize}
\item We reveal that the effectiveness of BN statistics concentrates in deep blocks whose statistics encode high-level semantics, which reframes the challenge of DFKD as the construction of architecture-agnostic semantic priors.
\item We propose UniDFKD, which instantiates such priors via CSC for depth-persistent categorical semantics, SSA for a spatial prior on class-discriminative semantics, and SSD for transferring the semantics from teacher to student.
\item Comprehensive experiments across CNNs and ViTs show that UniDFKD sets a new state of the art, outperforming the strongest baseline NAYER by average margins of 17.23 and 24.49 under homogeneous and heterogeneous settings.
\end{itemize}

%% file: sec/2_related.tex
\section{Related Work}
\begin{figure*}[t]
    \centering
    \begin{subfigure}[t]{0.42\linewidth}
        \centering
        \includegraphics[width=\linewidth]{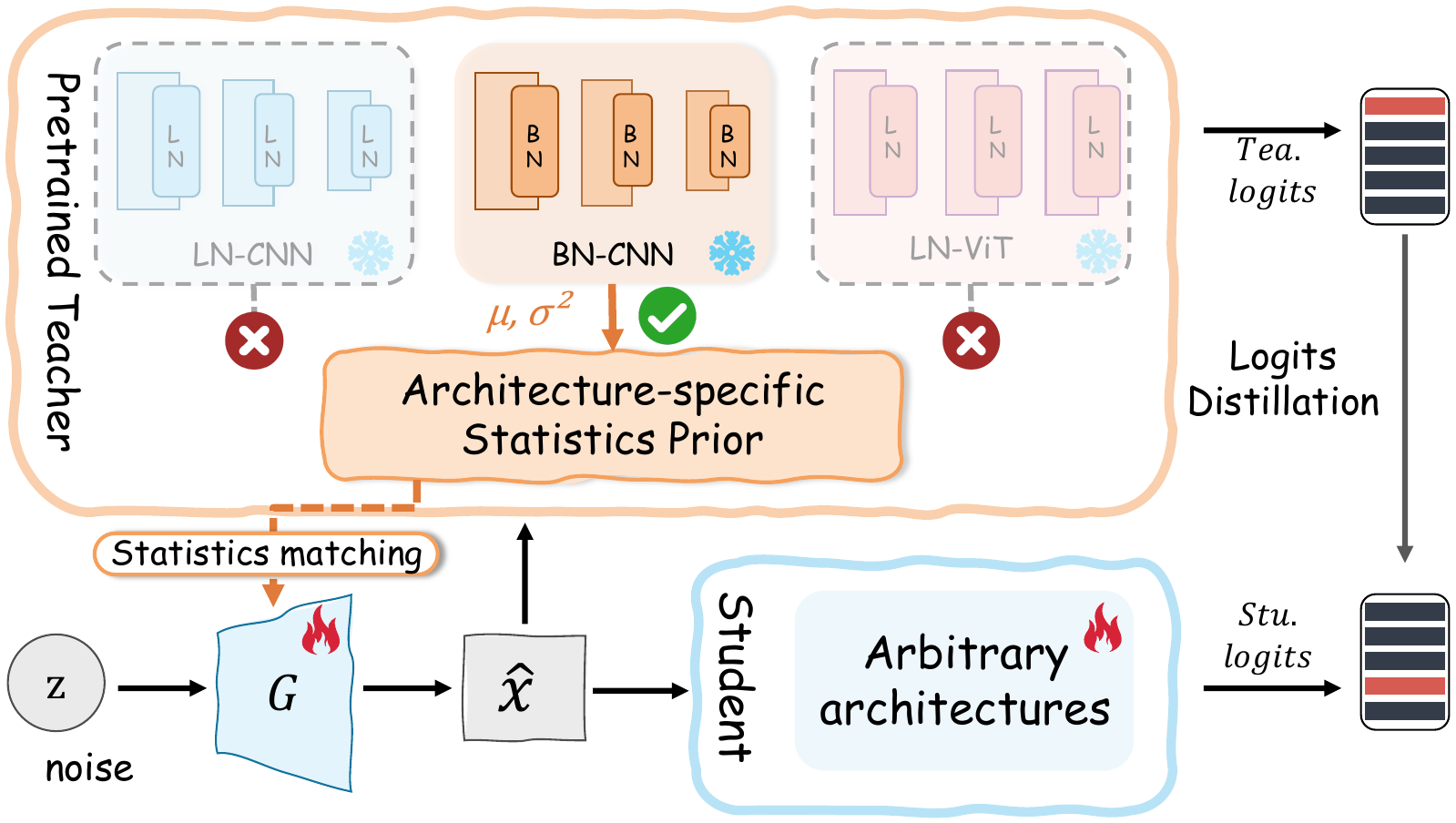}
        \caption{Previous BN-dependent DFKD.}
    \end{subfigure}
    \qquad
    \begin{subfigure}[t]{0.52\linewidth}
        \centering
        \includegraphics[width=\linewidth]{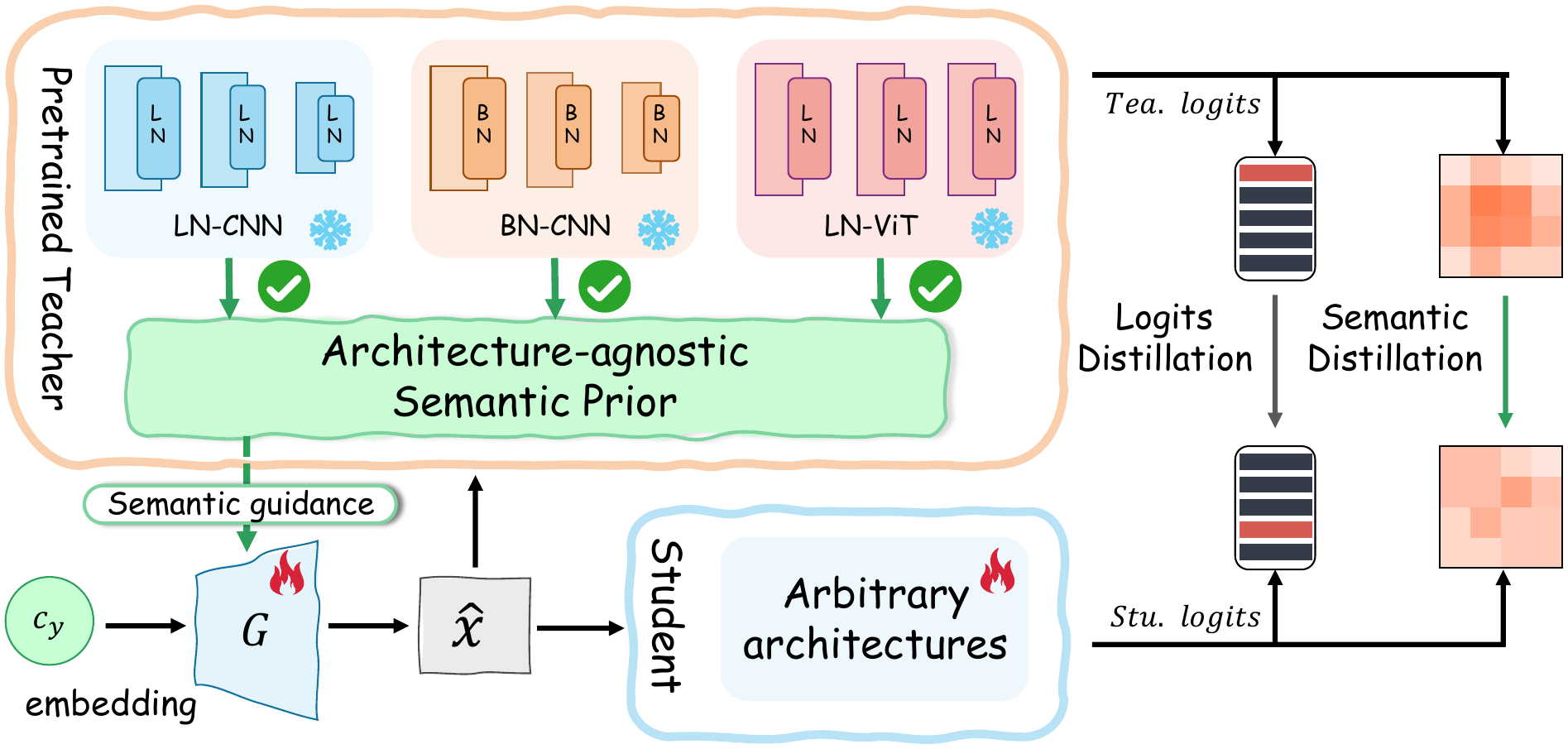}
        \caption{UniDFKD (ours).}
    \end{subfigure}
    \vspace{-3pt}
    \caption{Comparison of DFKD paradigms.
    (a)~Existing methods adopt BN statistics as an implicit semantic prior, confining synthesis to BN-based teachers.
    (b)~UniDFKD replaces this implicit prior with an explicit, architecture-agnostic semantic prior applicable to arbitrary teachers and augments distillation with spatial semantic transfer.
    }
    \label{fig:Comparison}
    \vspace{-5pt}
\end{figure*}

\subsection{Data-Free Knowledge Distillation}
Data-free knowledge distillation (DFKD) aims to recover a surrogate training distribution from a pretrained teacher alone~\cite{lopes2017data,liu2021data}.
Early methods primarily extracted supervision from the teacher's output space via confidence, class-similarity, or adversarial belief matching~\cite{chen2019data,nayak2019zero,micaelli2019zero}.
However, these output-level signals provide limited guidance for intermediate features, often yielding semantically incomplete samples.
A major turning point was DeepInversion~\cite{yin2020dreaming}, which established Batch Normalization (BN) running statistics as the de facto data prior.
Subsequent studies have improved synthesis diversity, efficiency, scalability, and robustness atop this common anchor~\cite{fang2021contrastive,yu2023data,fang2022up,liu2024small,tran2024muse,binici2022robust}.
However, using BN statistics as the primary distributional prior strongly couples data synthesis to the teacher's architectural design. 
As a result, such approaches fail to generalize to BN-free teachers (e.g., Vision Transformers and LayerNorm-based models), where this critical prior is absent.

\subsection{Semantic Priors in DFKD}
Beyond the BN prior, early alternatives imposed pixel-space regularizations like total variation and $\ell_2$ smoothness~\cite{mahendran2015understanding}, which are too weak to independently recover task-relevant distributions.
Recent methods introduce semantic knowledge but typically isolate a single facet.
To specify \emph{what} to synthesize, NAYER~\cite{tran2024nayer} conditions the generator on text embeddings. However, its input-level conditioning leaves categorical fidelity underconstrained in deeper layers and overlooks intra-class diversity.
To specify \emph{where} evidence appears, other works extract spatial cues via activation amplitudes~\cite{tran2024muse} or self-attention~\cite{zhong2024semantics}. Yet, these cues rely on architecture-specific operators and serve only as supplements to the BN prior.
To specify \emph{how} knowledge is transferred, CAT-KD \cite{guo2023class} aligns spatial activation responses between teacher and student networks on real data, whereas SpaceshipNet \cite{yu2023data} introduces activation region constraints into DFKD. Nevertheless, both methods formulate spatial response constraints exclusively for CNNs.
UniDFKD bridges these gaps by unifying categorical identity, intra-class diversity, and spatial organization.
By persistently injecting diverse categorical semantics throughout the generator and aligning class-discriminative spatial responses via an architecture-agnostic interface, UniDFKD establishes a holistic semantic prior.
This eliminates the reliance on BN statistics, enabling effective DFKD and spatial evidence transfer across heterogeneous architectures.

%% file: sec/3_method.tex
\section{Revisiting the Statistical Priors in DFKD}
\label{sec:motivation}
\ifmergedanalysisfig\else
\begin{figure}[t]
    \centering
    \includegraphics[width=\linewidth]{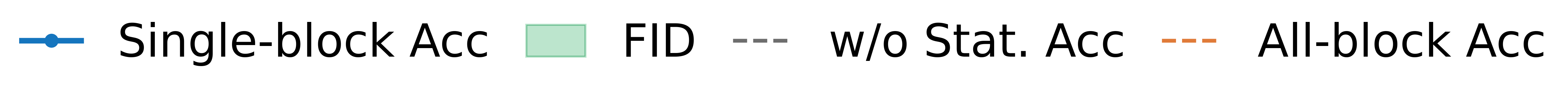}
    \\
    \begin{subfigure}[t]{0.495\linewidth}
        \centering
        \includegraphics[width=\linewidth]{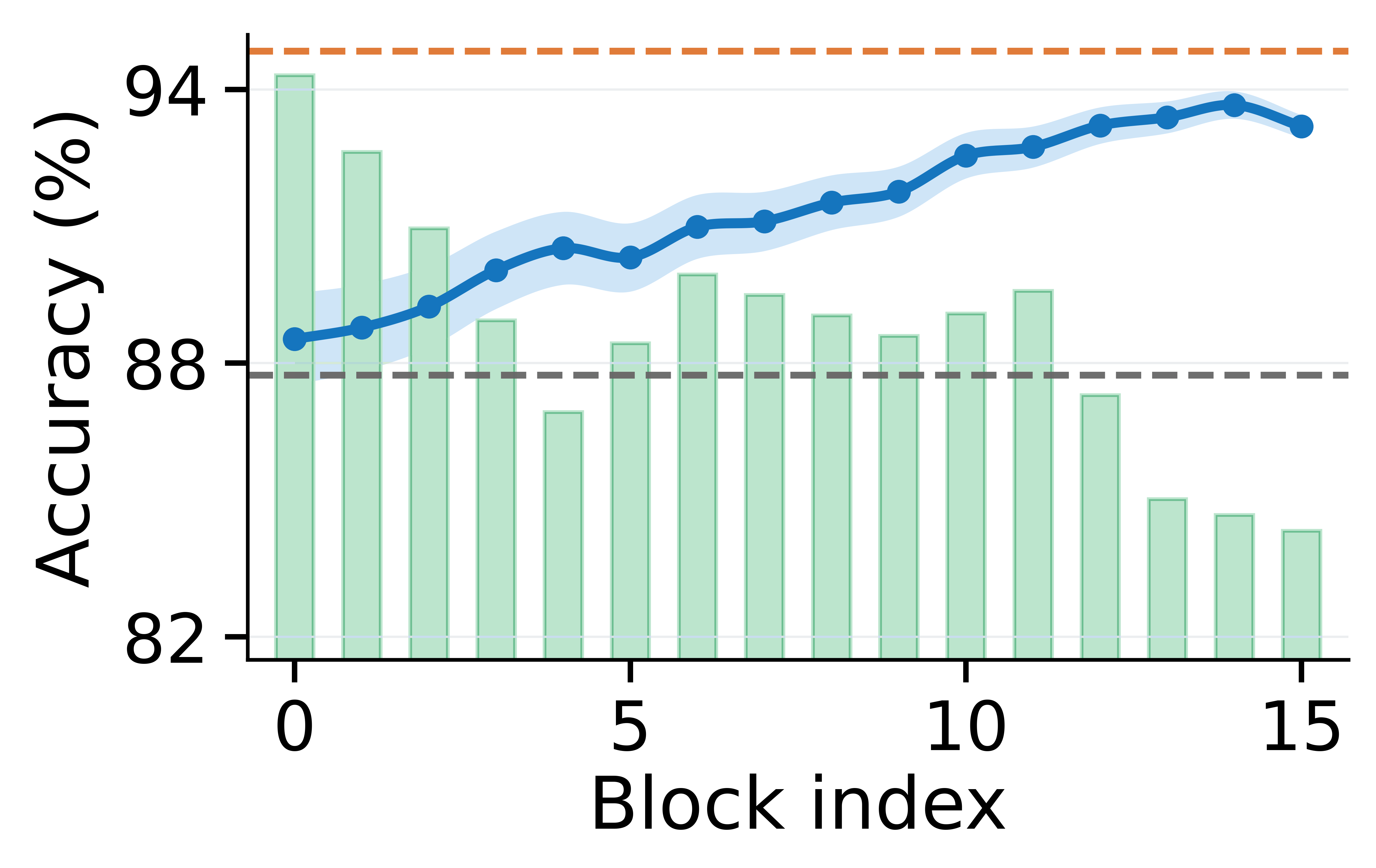}
        \caption{RN34$\to$RN18.}
        \label{fig:cmi_bn_layerwise_ablation_cnn}
    \end{subfigure}
    \begin{subfigure}[t]{0.495\linewidth}
        \centering
        \includegraphics[width=\linewidth]{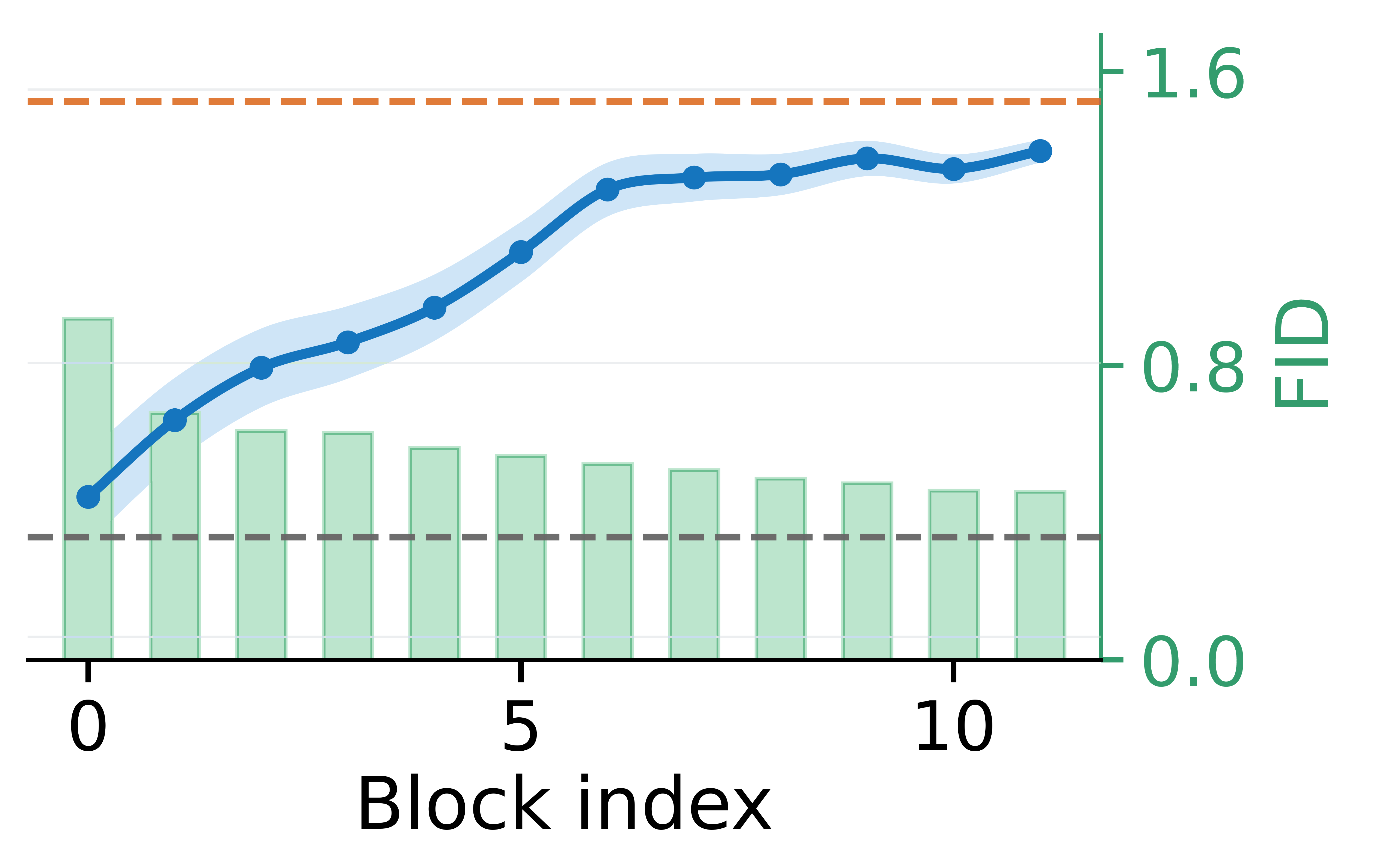}
        \caption{ViT-S$\to$ViT-T.}
        \label{fig:cmi_bn_layerwise_ablation_vit}
    \end{subfigure}
    \caption{Layer-wise feature-statistic matching on CIFAR-10.
    Student acc. ($\uparrow$) and T-FID ($\downarrow$) are reported for analysis.
    }
    \label{fig:cmi_bn_layerwise_ablation}
\end{figure}
\fi

\subsection{The Role of BN Statistics in DFKD}
Given a pretrained teacher $\mathcal{T}$ but no access to its training data $\mathcal{D}$, DFKD trains a compact student $\mathcal{S}$ using surrogate images $\hat{x}$ synthesized by a generator $\mathcal{G}$ initialized from scratch.
To regularize the synthetic distribution, existing methods align the intermediate feature statistics of $\hat{x}$ with the running statistics stored in all BN layers of the teacher~\cite{yin2020dreaming}: 
\begin{equation}
\mathcal{L}_{\mathrm{BN}}=\sum_{l=1}^L\bigl(\|\mu_l(\hat{x})-\mu_l^{\mathrm{BN}}\|_2+\|\sigma_l^2(\hat{x})-(\sigma_l^{\mathrm{BN}})^2\|_2\bigr).
\label{eq:bn-stat}
\end{equation}
Modern architectures such as ViT and ConvNeXt replace BN with LayerNorm~\cite{huang2023normalization}, which normalizes each sample independently and stores no running statistics, so $\mathcal{L}_{\mathrm{BN}}$ has no alignment target on BN-free teachers.
As Fig.~\ref{fig:c10_distillation_radar} shows, existing DFKD methods therefore degrade substantially under BN-free or heterogeneous settings. This bottleneck directly requires an architecture-agnostic prior for BN-free models.

\subsection{Why Do BN Statistics Work for DFKD?}
\label{subsec:why-bn}
To determine what such a prior should preserve, we examine what the BN prior encodes at each depth and which part of it DFKD relies on.
We first visualize per-sample BN statistics across four depths of the RN34 teacher (Fig.~\ref{fig:resnet_bn_statistics_tsne}). In shallow blocks, statistics from different classes intermingle because they reflect class-agnostic local textures and appearance~\cite{gatys2016image,zhou2021domain}. In deep blocks, they separate into distinct class-wise clusters as representations shift toward class-discriminative semantic structures~\cite{zhong2022fine,gao2022towards}. Deep BN statistics thus primarily encode high-level semantics.

We then isolate each block in turn, using only its statistics to guide synthesis and measuring the resulting distillation accuracy as that block's standalone contribution.
Experiments are conducted on both RN and ViT teachers, where ViT uses equivalent channel-wise activation moments as the target, serving as an oracle (details are in Sec.~\ref{app:layerwise} of the supplementary material).
Figs.~\ref{fig:cmi_bn_layerwise_ablation_cnn} and~\ref{fig:cmi_bn_layerwise_ablation_vit} show a consistent depth-wise trend across both RN and ViT teachers.
Matching shallow blocks barely improves over unconstrained synthesis whereas deeper blocks yield stronger synthetic distributional consistency (T-FID $\downarrow$) and steadily higher distillation performance.
Notably, matching the deepest block alone approaches full-layer matching.
Taken together, these two findings resolve our central question: shallow texture constraints are largely redundant, while deep semantic alignment accounts for nearly all distributional consistency and distillation gains regardless of architecture. Consequently, \emph{what makes BN statistics effective for DFKD is not the BN layer itself, but the deep semantic prior it implicitly enforces.}

Given the absence of running statistics in BN-free models, extending DFKD to arbitrary architectures calls for replacing implicit statistic matching with explicit semantic priors. We thus introduce UniDFKD to inject such architecture-agnostic priors directly into synthesis and distillation.
\begin{figure*}[t]
\centering
\includegraphics[width=0.97\linewidth]{figures/overall_1.pdf}
\caption{Overview of UniDFKD.
\textbf{Synthesis (left).} CSC builds a semantic bank $\mathcal{E}_{y}$ from LLM-generated class descriptions and modulates every generator block with a sampled embedding $\mathbf{c}_{y}$ via CBN, while SSA aligns the teacher's batch-averaged attribution $\bar{\mathbf{M}}^{\mathcal{T}}$ with a Gaussian prior; iterative generation accumulates the surrogate pool $\widehat{\mathcal{D}}$.
\textbf{Distillation (right).} SSD complements logit matching with attribution alignment, so the student also inherits the teacher's spatial evidence.
}
\label{fig:framework}
\vspace{-7pt}
\end{figure*}
\section{Unified Semantic Priors for DFKD}
\label{sec:method}
\subsection{Overview of UniDFKD}
\label{subsec:overview}

We decompose the construction of such a prior into three perspectives:
\emph{what} the semantics span---the class identity, inter-class relations, and intra-class diversity to be encoded in the synthesized distribution;
\emph{where} in the image plane they localize---the spatial locations where class-discriminative evidence should concentrate rather than scattering into backgrounds and textures; and
\emph{how} they transfer to the student---the explicit distillation of the semantics instilled in the synthetic images beyond logit matching alone.
The first two establish the structural and spatial priors for data synthesis, while the third seamlessly bridges this synthesized data with knowledge transfer.
Together, they form a closed loop that advances DFKD into a unified, semantics-driven paradigm.

As illustrated in Fig.~\ref{fig:framework}, UniDFKD operationalizes these within a unified two-stage synthesis--distillation pipeline.
During synthesis, Categorical Semantic Conditioning (CSC) addresses \emph{what} by conditioning generator $\mathcal{G}$ on language-derived embeddings to preserve inter-class relations and intra-class diversity. 
Meanwhile, Spatial Semantic Anchoring (SSA) addresses \emph{where} by aligning class-discriminative attribution with a Gaussian prior, concentrating semantic evidence in compact, center-biased regions.

Given the synthesized class-conditional images $\hat{x}$, the generator is updated by minimizing a combined objective comprising the cross-entropy loss $\mathcal{L}_{\mathrm{CE}}$, which optimizes $\hat{x}$ toward the designated class label, and the SSA loss $\mathcal{L}_{\mathrm{SSA}}$:
\begin{equation}
  \mathcal{L}_\mathcal{G}
  =\mathcal{L}_{\mathrm{CE}}
  +\lambda_{\mathrm{SSA}}\mathcal{L}_{\mathrm{SSA}}.
  \label{eq:generator-objective}
\end{equation}

During distillation, we address \emph{how} the student learns through a joint knowledge transfer objective. This incorporates the KL divergence $\mathcal{L}_{\mathrm{KL}}$ between the teacher and student predictions, and the Spatial Semantic Distillation (SSD) loss $\mathcal{L}_{\mathrm{SSD}}$, which supplements prediction matching by transferring the underlying spatial evidence:
\begin{equation}
\mathcal{L}_{\mathrm{KD}}
=\mathcal{L}_{\mathrm{KL}}
+\lambda_{\mathrm{SSD}}\,\mathcal{L}_{\mathrm{SSD}}.
\label{eq:distillation-objective}
\end{equation}

\subsection{Categorical Semantic Conditioning}
\label{subsec:csc}
Categorical Semantic Conditioning (CSC) leverages natural-language descriptions in a shared vision--language space as a surrogate for real image geometry, capturing both inter-class relatedness and intra-class diversity~\cite{radford2021learning,pratt2023what}. 
We implement this via a two-stage semantic conditioning pipeline that first refines captions into a compact per-class semantic bank, and then conditions the generator on the sampled embeddings.
\paragraph{Diverse Embedding.} For each class $y$, we first leverage large language models (LLMs) to generate a set of candidate prompts that describe complementary semantic modes such as attributes, viewpoints, and contexts~\cite{pratt2023what}.
A frozen text encoder~\cite{radford2021learning} then maps these prompts into text embeddings.
Prompts derived without strict constraints inevitably contain noisy descriptions that act as out-of-distribution conditions for class $y$.
Therefore, we apply spherical $k$-means~\cite{dhillon2001concept} to partition these embeddings into $K$ clusters $\{\mathcal{C}_{y,k}\}_{k=1}^{K}$, retaining only the $r$ nearest members to each normalized centroid $\mathbf{v}_{y,k}$ to yield the diverse semantic bank $\mathcal{E}_y$.
\begin{equation}
\mathcal E_y
=
\left\{
\mathbf e\in\mathcal C_{y,k}
\;\middle|\;
1\le k\le K,\;
\operatorname{rank}_{\mathcal C_{y,k}}
\left(
\mathbf e^\top\mathbf v_{y,k}
\right)
\le r
\right\},
\label{eq:semantic-filtering}
\end{equation}
where $\operatorname{rank}_{\mathcal C_{y,k}}(\cdot)$ denotes the descending rank of the similarity score within $\mathcal C_{y,k}$. 
$\mathcal{E}_y$ keeps $K{\times}r$ strictly in-distribution embeddings that still comprehensively cover the semantic structures and intra-class diversity of class $y$. During synthesis, each sample of class $y$ draws its semantic embedding $\mathbf{c}_y$ uniformly from $\mathcal{E}_y$. This sampled embedding is then projected to construct the initial feature $\mathbf{h}_0 = \mathrm{MLP}(\mathbf{c}_y)$, which serves as the foundational input for the generator.
\paragraph{Hierarchical Embedding Injection.}
Although input-level conditioning establishes the desired semantic mode in $\mathbf{h}_0$, it doesn't preserve this information throughout the synthesis hierarchy. 
In the deeper generators required for $224$-scale synthesis, the influence of $\mathbf{c}_y$ progressively weakens across successive upsampling blocks, leaving intermediate features insufficiently constrained by the sampled categorical semantics.
To sustain these semantics along the entire synthesis hierarchy, we inject $\mathbf{c}_y$ into every generator block through Conditional Batch Normalization (CBN)~\cite{devries2017modulating}.
Formally, the generator stacks $J$ upsampling blocks, and within each block the static affine of its normalization layer is replaced by linear projections of the embedding \(\mathbf{c}_y\) sampled from \(\mathcal{E}_y\):
\begin{equation}
\begin{aligned}
\hat{x}&=(G_J\circ\cdots\circ G_1)(\mathbf{h}_0),\\
\mathbf{h}_j&=\boldsymbol{\gamma}_j(\mathbf{c}_y)\odot\operatorname{Norm}(\tilde{\mathbf{h}}_j)+\boldsymbol{\beta}_j(\mathbf{c}_y),
\end{aligned}
\label{eq:persistent-categorical-modulation}
\end{equation}
where $\mathbf{h}_j$ is the intermediate feature output by block $G_j$, $\tilde{\mathbf{h}}_j$ is its pre-normalization activation, and $\boldsymbol{\gamma}_j,\boldsymbol{\beta}_j$ are block-specific linear projections broadcast over spatial locations. 
Persistent conditioning on semantic embeddings encourages the synthesized distribution to retain the inter-class organization of the shared text space, whereas sampling diverse embeddings from $\mathcal{E}_y$ expands the intra-class semantic modes covered for each category (Fig.~\ref{fig:dist-quality}b).

\subsection{Spatial Semantic Anchoring}
\label{subsec:ssa}
Spatial Semantic Anchoring (SSA) draws on the observation that objects in natural images tend to occupy compact regions near the center~\cite{schauerte2015distribution}, inducing a center-biased distribution of class-discriminative responses.
Accordingly, attribution maps (e.g., Grad-CAM~\cite{selvaraju2017grad}) aggregated across samples typically exhibit a center-peaked spatial profile (Fig.~\ref{fig:cdsa_vis}).
We approximate this regularity with a Gaussian prior and align the class-discriminative attribution of synthetic images to it.

To decouple SSA from architecture-specific internals, we abstract the core principles of existing interpretability techniques~\cite{zhou2016learning,chefer2021transformer} to define a unified class-discriminative spatial attribution (CDSA) interface.
Given an input $\hat{x}$ and a classifier $f$ over $C$ categories, we first recover a spatial response map
$\mathbf{R}^{f}(\hat{x})\in\mathbb{R}^{H_p\times W_p\times D_f}$
from its deepest task-related representation before the spatial information is collapsed into a global prediction.
We then project each spatial response through the classification-head weights
$\mathbf{W}^{C,f}\in\mathbb{R}^{D_f\times C}$:
\begin{equation}
\mathbf{M}^{f}(\hat{x})
=
\mathbf{R}^{f}(\hat{x})\mathbf{W}^{C,f}
\in\mathbb{R}^{H_p\times W_p\times C},
\label{eq:attribution-interface}
\end{equation}
where $\mathbf{M}^{f}_{c}[i,j]$ denotes the CDSA score assigned to spatial location $(i,j)$ for category $c$.
Although architectures differ in how $\mathbf{R}^{f}$ is extracted, all maps consistently yield spatially resolved class-discriminative responses within a common $C$-dimensional category space.
Specifically, for CLS-based models (e.g., ViT), $\mathbf{R}^{f}$ is assembled from the attention-weighted patch values before CLS token aggregation, whereas for architectures that retain
spatial feature maps (e.g., ConvNeXt), it corresponds directly to the final feature map before pooling.
Detailed constructions for each architecture are provided in Sec.~\ref{app:cdsa_instantiation} of the supplementary material.\looseness=-1

Over a synthetic batch, we aggregate the per-sample maps into a batch statistic via ReLU,
$\bar{\mathbf{M}}=\bigl(\mathbb{E}_{\hat{x}}[\sigma\left(\mathbf{M}(\hat{x})\right)]\bigr)\in\mathbb{R}^{H_p\times W_p\times C}$,
to retain positive attribution responses.
We then align it against a Gaussian prior that encodes the center-biased regularity of natural images.
Given that the spatial tendency is a broadly observed statistical regularity across natural images, we analytically construct the prior on the $H_p\times W_p$ patch grid, requiring neither real data nor statistics stored in the teacher:
\begin{equation}
\boldsymbol{\pi}[i,j]
= B + A\,\exp\!\left(
  -\frac{1}{2}\left[
    \frac{(i-\mu_x)^2}{\sigma_x^2}
    +\frac{(j-\mu_y)^2}{\sigma_y^2}
  \right]
\right),
\label{eq:structural-prior}
\end{equation}
where 
the peak location $(\mu_x,\mu_y)$ is resampled for each batch within a small window around the grid center, $(\sigma_x,\sigma_y)$ control the expected spatial extent of discriminative evidence, and $A$, $B$ set the peak strength and the residual background response.
The sampling ranges of these parameters are provided in Sec.~\ref{app:exp_settings} of the supplementary material.
We then align the aggregated CDSA of each category with the Gaussian prior:\looseness=-1
\vspace{-5pt}
\begin{equation}
\label{eq:spatial-alignment}
\mathcal{L}_{\mathrm{SSA}}=
\frac{1}{C}\sum_{c=1}^{C}
(
1-\cos(
\bar{\mathbf{M}}^{\mathcal T}_{c},
\boldsymbol{\pi})
).
\vspace{-5pt}
\end{equation}
The cosine objective aligns only the spatial shape with the prior while remaining agnostic to the teacher-dependent magnitude of $\bar{\mathbf{M}}$, consistent with the design of center bias.

\begin{table*}[t]
\centering
\caption{Student accuracy (\%) in homogeneous and heterogeneous data-free distillation across three datasets.
$T.$/$S.$ acc.: teacher/student trained from scratch on real data; ``Avg.'': mean across the 12 configurations.
Blue rows: UniDFKD; \textbf{bold}/\underline{underlined}: best/second best.}
\label{tab:main}
\vspace{-4pt}
\footnotesize
\setlength{\tabcolsep}{5.5pt}
\renewcommand{\arraystretch}{1.03}
\begin{tabular}{l cccc cccc cccc @{\hspace{12pt}}>{\columncolor{gray!10}}c}
\toprule
& \multicolumn{4}{c}{CIFAR-10}
& \multicolumn{4}{c}{CIFAR-100}
& \multicolumn{4}{c}{ImageNet-100}
& \\
\cmidrule(lr){2-5}
\cmidrule(lr){6-9}
\cmidrule(lr){10-13}

\multicolumn{14}{l}{\textit{Homogeneous distillation}} \\[1pt]

$Teacher$
  & RN34  & CNX-T  & ViT-S & Swin-T
  & RN34  & CNX-T  & ViT-S & Swin-T
  & RN34  & CNX-T  & ViT-S & Swin-T
  & \\
$Student$
  & RN18  & CNX-N  & ViT-T & Swin-N
  & RN18  & CNX-N  & ViT-T & Swin-N
  & RN18  & CNX-N  & ViT-T & Swin-N
  & \multirow{-2}{*}{Avg.} \\

\midrule

{\color{gray}T. acc.}
  & {\color{gray}95.70} & {\color{gray}97.54} & {\color{gray}95.25} & {\color{gray}96.43}
  & {\color{gray}78.05} & {\color{gray}83.10} & {\color{gray}77.93} & {\color{gray}81.24}
  & {\color{gray}85.92} & {\color{gray}88.98} & {\color{gray}85.06} & {\color{gray}90.83}
  & {\color{gray}88.00} \\
{\color{gray}S. acc.}
  & {\color{gray}95.20} & {\color{gray}97.41} & {\color{gray}94.01} & {\color{gray}96.08}
  & {\color{gray}77.10} & {\color{gray}82.50} & {\color{gray}73.80} & {\color{gray}79.22}
  & {\color{gray}84.72} & {\color{gray}86.27} & {\color{gray}83.20} & {\color{gray}88.30}
  & {\color{gray}86.48} \\

\cmidrule{1-14}

DeepInversion
  & 89.69 & 69.84 & 53.34 & 59.62
  & 61.32 & 17.69 & 14.51 & 17.77
  & 55.98 & 19.36 & 3.60 & 3.28
  & 38.83 \\
Fast
  & 94.05 & 73.31 & 83.71 & 72.53
  & 74.34 & 58.75 & 23.65 & 56.17
  & 69.84 & 34.53 & 13.34 & 22.81
  & 56.42 \\
CMI
  & 94.84 & \underline{79.28} & 84.19 & \underline{78.18}
  & 77.04 & \underline{70.09} & 42.42 & \underline{68.41}
  & 70.42 & \underline{39.50} & 23.56 & 18.30
  & 62.19 \\
NAYER
  & \underline{95.21} & 73.69 & \underline{85.86} & 73.94
  & \textbf{77.54} & 66.48 & \underline{57.51} & 66.32
  & \underline{78.60} & 39.10 & \underline{39.30} & \underline{29.76}
  & \underline{65.28}\\
\rowcolor{tableblue}
UniDFKD
  & \textbf{95.33} & \textbf{94.47} & \textbf{93.91} & \textbf{91.75}
  & \underline{77.37} & \textbf{80.23} & \textbf{70.41} & \textbf{77.94}
  & \textbf{83.08} & \textbf{83.40} & \textbf{71.16} & \textbf{71.00}
  & \textbf{82.50} \\

\midrule[0.08em]

\multicolumn{14}{l}{\textit{Heterogeneous distillation}} \\[1pt]

$Teacher$
  & RN34  & CNX-T & ViT-S & Swin-T
  & RN34  & CNX-T & ViT-S & Swin-T
  & RN34  & CNX-T & ViT-S & Swin-T
  & \\
$Student$
  & ViT-T & ViT-T & RN18  & CNX-N
  & ViT-T & ViT-T & RN18  & CNX-N
  & ViT-T & ViT-T & RN18  & CNX-N
  & \multirow{-2}{*}{Avg.} \\

\midrule

{\color{gray}T. acc.}
  & {\color{gray}95.70} & {\color{gray}97.54} & {\color{gray}95.25} & {\color{gray}96.43}
  & {\color{gray}78.05} & {\color{gray}83.10} & {\color{gray}77.93} & {\color{gray}81.24}
  & {\color{gray}85.92} & {\color{gray}88.98} & {\color{gray}85.06} & {\color{gray}90.83}
  & {\color{gray}88.00} \\
{\color{gray}S. acc.}
  & {\color{gray}94.01} & {\color{gray}94.01} & {\color{gray}95.20} & {\color{gray}97.41}
  & {\color{gray}73.80} & {\color{gray}73.80} & {\color{gray}77.10} & {\color{gray}82.50}
  & {\color{gray}83.20} & {\color{gray}83.20} & {\color{gray}84.72} & {\color{gray}86.27}
  & {\color{gray}85.44} \\

\cmidrule{1-14}

DeepInversion
  & 71.72 & 25.31 & 33.81 & 30.36
  & 52.53 & 7.02  & 11.86 & 33.02
  & 15.58 & 5.06  & 6.07  & 10.99
  & 25.28 \\
Fast
  & 86.15 & 40.35 & 72.73 & 72.07
  & 63.15 & 11.79 & 15.70 & 41.98
  & 49.83 & 7.65  & 23.73 & 48.91
  & 44.50 \\
CMI
  & 91.19 & \underline{43.72} & \underline{75.79} & \underline{77.18}
  & 69.83 & 25.30 & 35.55 & \underline{65.70}
  & 57.30 & 9.16  & 24.95 & 37.83
  & 51.13 \\
NAYER
  & \underline{93.38} & 36.90 & 75.73 & 73.46
  & \underline{70.92} & \underline{34.46} & \underline{44.66} & 61.88
  & \underline{66.54} & \underline{11.74} & \underline{28.52} & \underline{57.86}
  & \underline{54.67} \\
\rowcolor{tableblue}
UniDFKD
  & \textbf{94.54} & \textbf{91.99} & \textbf{90.64} & \textbf{93.64}
  & \textbf{74.59} & \textbf{73.15} & \textbf{71.67} & \textbf{75.58}
  & \textbf{70.32} & \textbf{68.36} & \textbf{66.72} & \textbf{78.72}
  & \textbf{79.16} \\

\bottomrule
\end{tabular}%
\vspace{-8pt}
\end{table*}

\subsection{Spatial Semantic Distillation}
\label{subsec:SSD}

Spatial Semantic Distillation (SSD) focuses on effectively transferring both categorical and spatial semantics from synthetic images to the student.
Although the KL loss relays the teacher's categorical knowledge, it discards the class-discriminative spatial semantics behind it.
We therefore align the per-sample CDSA maps of teacher and student (Eq.~\eqref{eq:attribution-interface}) via channel-wise similarity, upsampling the smaller map when patch resolutions differ:
\vspace{-5pt}
\begin{equation}
\begin{aligned}
\mathcal{L}_{\mathrm{SSD}}&=
\mathbb E_{\hat{x}\sim\widehat{\mathcal D}}
\left[
\frac{1}{C}\sum_{c=1}^{C}
(
1-\cos(
\mathbf M^{\mathcal T}_{c},
\mathbf M^{\mathcal S}_{c})
)
\right].
\end{aligned}
\label{eq:student-objective}
\vspace{-3pt}
\end{equation}
Note that $\mathcal{L}_{\mathrm{SSD}}$ and $\mathcal{L}_{\mathrm{SSA}}$ play complementary roles: the latter constrains the spatial statistics of synthesized images, while the former aligns per-sample attribution during distillation, so the student inherits both the teacher's class predictions and the spatial discriminative evidence.

%% file: sec/4_experiments.tex
\section{Experiments}
\label{sec:experiments}
\label{subsec:setup}
\paragraph{Datasets and architectures.}
We evaluate UniDFKD on CIFAR-10, CIFAR-100~\cite{krizhevsky2009learning}, and ImageNet-100~\cite{tian2020contrastive} (a 100-class subset of ILSVRC-2012~\cite{deng2009imagenet}). 
Our experiments span four architecture families: the BN-based ResNet (RN)~\cite{he2016deep}, alongside the LN-based ConvNeXt (CNX)~\cite{liu2022convnet}, Vision Transformer (ViT)~\cite{dosovitskiy2021image}, and Swin Transformer (Swin)~\cite{liu2021swin}.
To comprehensively assess our method, we construct distillation pairs under two regimes: homogeneous (e.g., within CNNs or ViTs) and heterogeneous (e.g., ViT$\to$RN). 

\paragraph{Baselines.}
We compare our approach against the representative DeepInversion~\cite{yin2020dreaming}, Fast~\cite{fang2022up}, CMI~\cite{fang2021contrastive}, and NAYER~\cite{tran2024nayer}; comparisons with further methods on more conventional CNNs are provided in Sec.~\ref{app:general_cnn} of the supplementary material. 
For fair benchmarking, all methods share the same teachers and apply $\mathcal{L}_{\mathrm{BN}}$ only for BN-based teachers. We adopt their official implementations and train all methods with an architecture-optimized recipe of optimizer, learning rate, data augmentation and distillation iterations. We further sweep the method-specific hyperparameters of each baseline (e.g., the loss weights) to establish strong baselines.\looseness=-1

\paragraph{Implementation details.}
For CSC, 200 captions per class are generated by DeepSeek-V4~\cite{deepseekai2026deepseekv4} from class names alone and encoded by a frozen OpenCLIP ViT-H/14 text encoder~\cite{cherti2023reproducible}; as with the LTE of NAYER~\cite{tran2024nayer}, no real or external images are involved.
We upsample CIFAR images to 224$\times$224 for all pairs except RN34$\to$RN18 to match the input size of modern architectures. 
The generator thus synthesizes at this resolution.
We report the best top-1 validation accuracy, averaged over three seeds for all methods.
Full hyperparameters, augmentation recipes, and teacher training protocols are provided in Sec.~\ref{app:exp_settings} of the supplementary material.\looseness=-1

\subsection{Main Results}
\label{subsec:main-results}

As shown in Tab.~\ref{tab:main}, UniDFKD achieves state-of-the-art accuracy in 23 out of 24 evaluated settings, including 11 of 12 homogeneous and all 12 heterogeneous scenarios. It attains average accuracies of 82.50 and 79.16, outperforming the strongest baseline NAYER by 17.23 and 24.49, respectively. 

\paragraph{BN-based Teachers.}
When the teacher is equipped with BN, $\mathcal{L}_{\mathrm{BN}}$ already supplies sufficient distributional guidance on CIFAR, enabling existing methods to approach the real-data reference in homogeneous RN34$\to$RN18 distillation, where UniDFKD matches NAYER within a 0.17 margin on CIFAR-100. However, UniDFKD still surpasses baselines on the more complex ImageNet-100 and yields clear gains in cross-architecture RN34$\to$ViT transfer across all benchmarks, outperforming NAYER by 3.67 on CIFAR-100 and exceeding the 73.80 real-data reference. 

\paragraph{BN-free Homogeneous Distillation.}
The advantage grows substantially with BN-free teachers. While baselines maintain moderate accuracy on CIFAR, even the strongest baselines drop by more than 43 below the real-data reference on ImageNet-100 across all three BN-free families. In stark contrast, UniDFKD bridges this gap across both ViTs and CNNs, outperforming the strongest baselines by 31.86 on ViT, 41.24 on Swin, and 43.90 on CNX, where it reaches 83.40 and comes within 2.87 of the 86.27 real-data reference.

\paragraph{BN-free Heterogeneous Distillation.}
Fixing the student to ViT-T while switching the teacher from RN34 to CNX-T directly highlights our robustness to teacher selection. Across all three benchmarks, NAYER suffers an average drop of 49.25 when moving to the CNX-T teacher, whereas UniDFKD varies by only 1.98 on average. Specifically, on CNX-T$\to$ViT-T, UniDFKD recovers within 0.65 of the real-data reference on CIFAR-100 and outperforms NAYER by 56.62 on ImageNet-100. Together, these results validate that our proposed explicit semantic priors enable reliable data-free transfer across all evaluated BN-free teachers.
\subsection{Ablation Study and Component Discussion}
\label{subsec:ablation}

\begin{table}[t]
\centering
\caption{Component ablation on ViT-S$\to$ViT-T for CIFAR-10. }
\label{tab:ablation}
\vspace{-4pt}
\footnotesize
\setlength{\tabcolsep}{2.5pt}
\renewcommand{\arraystretch}{1.00}
\begin{tabular*}{\columnwidth}{@{\extracolsep{\fill}}lcccccccc@{}}
\toprule
& \multicolumn{1}{c}{Base} & \multicolumn{3}{c}{Single} & \multicolumn{3}{c}{Pair} & \multicolumn{1}{c}{Full}\\
\cmidrule(lr){2-2}\cmidrule(lr){3-5}\cmidrule(lr){6-8}\cmidrule(lr){9-9}
CSC & \gmark & \cmark & \gmark & \gmark & \cmark & \cmark & \gmark & \cmark\\
SSA & \gmark & \gmark & \cmark & \gmark & \cmark & \gmark & \cmark & \cmark\\
SSD & \gmark & \gmark & \gmark & \cmark & \gmark & \cmark & \cmark & \cmark\\
\midrule
Acc. & 82.45 & 89.77 & 87.51 & 88.29 & 90.51 & 92.53 & 92.69 & \textbf{93.91}\\
\bottomrule
\end{tabular*}
\vspace{-2pt}
\end{table}

\begin{table}[tb]
\centering
\caption{SSA spatial priors on CIFAR-10.}
\label{tab:ssa-priors}
\vspace{-4pt}
\footnotesize
\setlength{\tabcolsep}{2.5pt}
\begin{tabularx}{\columnwidth}{@{}l*{3}{>{\centering\arraybackslash}X}@{}}
\toprule
Teacher & RN34 & ViT-S & Swin-T \\
Student & RN18 & ViT-T & CNX-N \\
\midrule
Uniform ($U(0.5)$)                    & 95.01 & 93.20 & 93.15 \\
Inverted Gaussian ($\max(\pi)-\pi$) & 92.20 & 90.60 & 89.20 \\
Gaussian (fixed center)             & 95.12 & 93.30 & 93.55 \\
\rowcolor{gray!15}
Gaussian (sampled center)           & \textbf{95.33} & \textbf{93.91} & \textbf{93.64} \\
\bottomrule
\end{tabularx}%
\vspace{-5pt}
\end{table}

\paragraph{Component Effectiveness.}
Tab.~\ref{tab:ablation} reports the ablation results of CSC, SSA, and SSD. 
With all three modules disabled, the student achieves an accuracy of 82.45. 
Individually enabling CSC, SSA, and SSD boosts accuracy by 7.32, 5.06, and 5.84. 
When CSC is disabled so that synthesis uses no extra LLM-generated descriptions or text embeddings, SSA+SSD alone reaches 92.69, outperforming the Base model by 10.24 and text-prior NAYER (85.86). 
Adding CSC further improves accuracy by 1.22. 
Finally, ablating spatial semantic prior generation and distillation thoroughly confirms that these modules address complementary aspects of the data-free synthesis bottleneck.

\begin{table}[!t]
\centering
\caption{CSC analysis on ViT-S$\to$ViT-T for CIFAR-10.}
\label{tab:csc}
\vspace{-4pt}
\footnotesize
\begin{subtable}[b]{0.585\linewidth}
\setlength{\tabcolsep}{3.5pt}
\renewcommand{\arraystretch}{1.00}
\begin{tabular}{lcccc}
\toprule
& Sem. & Div. & Filt. & Acc.\\
\midrule
Noise    & \gmark & \gmark & \gmark & 82.45\\
Prototype   & \cmark & \gmark & \gmark & 87.56\\
Diverse  & \cmark & \cmark & \gmark & 89.09\\
\rowcolor{gray!15}
Filtered & \cmark & \cmark & \cmark & \textbf{89.77}\\
\bottomrule 
\end{tabular}
\caption{Conditioning type}
\label{tab:csc_input}
\end{subtable}%
\hfill
\begin{subtable}[b]{0.40\linewidth}
\centering
\setlength{\tabcolsep}{4.5pt}
\renewcommand{\arraystretch}{1.00}
\begin{tabular}{lc}
\toprule
Mechanism & Acc.\\
\midrule
w/o inject  & 87.61\\
Additive    & 87.58\\
SE attn.    & 89.13\\
\rowcolor{gray!15}
CBN & \textbf{89.77}\\
\bottomrule 
\end{tabular}
\caption{Injection mechanism}
\label{tab:csc_inject}
\end{subtable}
\vspace{-6pt}
\end{table}

\begin{figure}[!t]
\centering
\includegraphics[width=\linewidth]{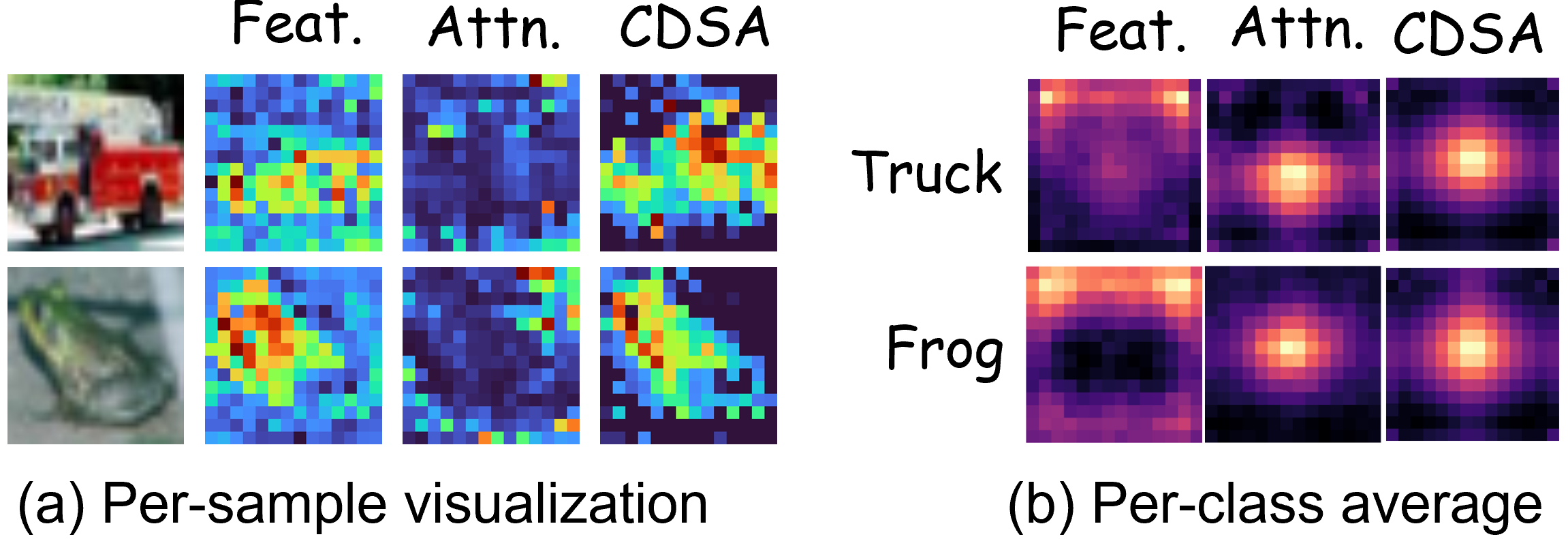}
\par\vspace{-4pt}
\caption{
Visual comparison of our class-discriminative spatial attribution (CDSA) map against alternative attention mechanisms.
$Feat.$ and $Attn.$ denote the $\ell_2$ norm of patch tokens and the CLS attention of the last layer, respectively. 
}
\label{fig:cdsa_vis}
\vspace{-8pt}
\end{figure}

\paragraph{Exploration of CSC Design Choices.}
Tab.~\ref{tab:csc} analyzes CSC in isolation on the Base model from Tab.~\ref{tab:ablation} with SSA and SSD disabled. 
As shown in Tab.~\ref{tab:csc_input}, replacing noise with a single class-name prototype raises accuracy from 82.45 to 87.56, confirming that categorical semantics provide a strong generation prior. 
This setting uses the exact same external information as NAYER (85.86) yet surpasses it by 1.70, demonstrating the effectiveness of our hierarchical CBN injection over NAYER's input-only injection.
Diversifying intra-class embeddings with LLM-generated descriptions and filtering OOD samples further maximizes accuracy to 89.77, aligning the synthetic distribution with the ground-truth categorical structure.

For feature injection (Tab.~\ref{tab:csc_inject}), direct additive fusion falls below the no-injection baseline by collapsing generation diversity. 
While SE attention~\cite{hu2018squeeze} helps via channel-wise reweighting, our CBN performs best by modulating normalization layers, achieving the optimal trade-off between semantic consistency and generation diversity.

\paragraph{Analysis of SSA Design Choices.}
Designing SSA requires both an architecture-agnostic spatial representation and an appropriate target distribution. 

First, we justify CDSA as the alignment representation by comparing it with raw attention $Attn.$ and feature maps $Feat.$ from a ViT-S teacher in Fig.~\ref{fig:cdsa_vis}. Instance-wise in Fig.~\ref{fig:cdsa_vis}a, both $Attn.$ and $Feat.$ suffer from severe background activation, whereas CDSA localizes the target object to provide compact discriminative evidence. Distribution-wise under real data in Fig.~\ref{fig:cdsa_vis}b, $Feat.$ collapses into a near-uniform profile, while $Attn.$ relies on self-attention that precludes generalization to CNNs. In contrast, CDSA produces a smooth, architecture-agnostic profile that peaks near the image center, naturally motivating our Gaussian prior in Eq.~\eqref{eq:structural-prior}.\looseness=-1

Next, we compare four target spatial priors in Tab.~\ref{tab:ssa-priors}. 
Averaged over the three settings, even the structureless uniform prior $U(0.5)$ reaches 93.79, and on ViT-S$\to$ViT-T it lifts the variant without SSA from 92.53 to 93.20, showing that aligning CDSA with a spatially regular target already benefits synthesis. Inverting the Gaussian so that evidence is pushed toward image borders instead lowers accuracy to 90.67, which shows that SSA effectively steers where class evidence emerges and that its gains come from agreeing with the center-peaked CDSA profile of real data (Fig.~\ref{fig:cdsa_vis}b). Matching this profile with a Gaussian and further sampling its peak coordinates $(\mu_x, \mu_y)$ performs best at 94.29, exceeding the fixed-center Gaussian and the uniform prior by 0.30 and 0.51, respectively. 
Together, these choices ensure that SSA aligns compact, architecture-agnostic object evidence while balancing structural regularity with positional diversity.

\subsection{Further Analysis}
\label{subsec:analysis}
\begin{figure}[tb]
\centering
\begin{subfigure}[t]{0.495\linewidth}
    \centering
    \includegraphics[width=\linewidth]{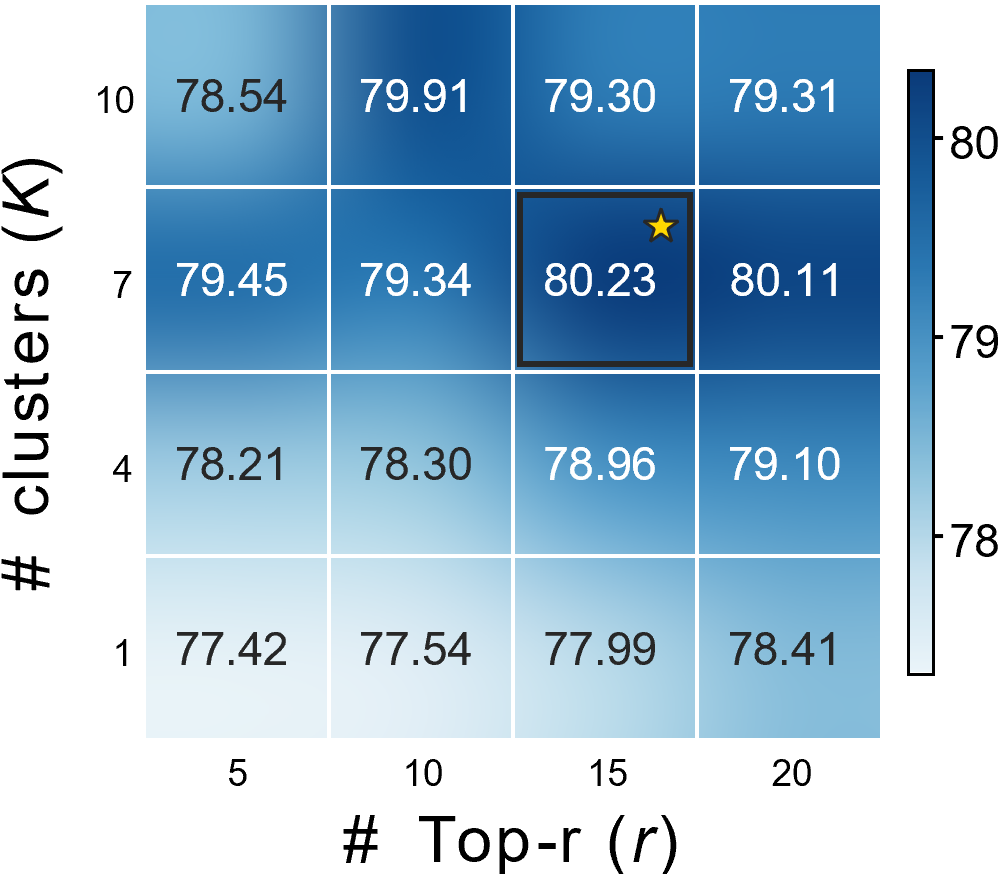}
    \caption{Semantic bank.}
    \label{fig:hyper-cluster}
\end{subfigure}
\begin{subfigure}[t]{0.495\linewidth}
    \centering
    \includegraphics[width=\linewidth]{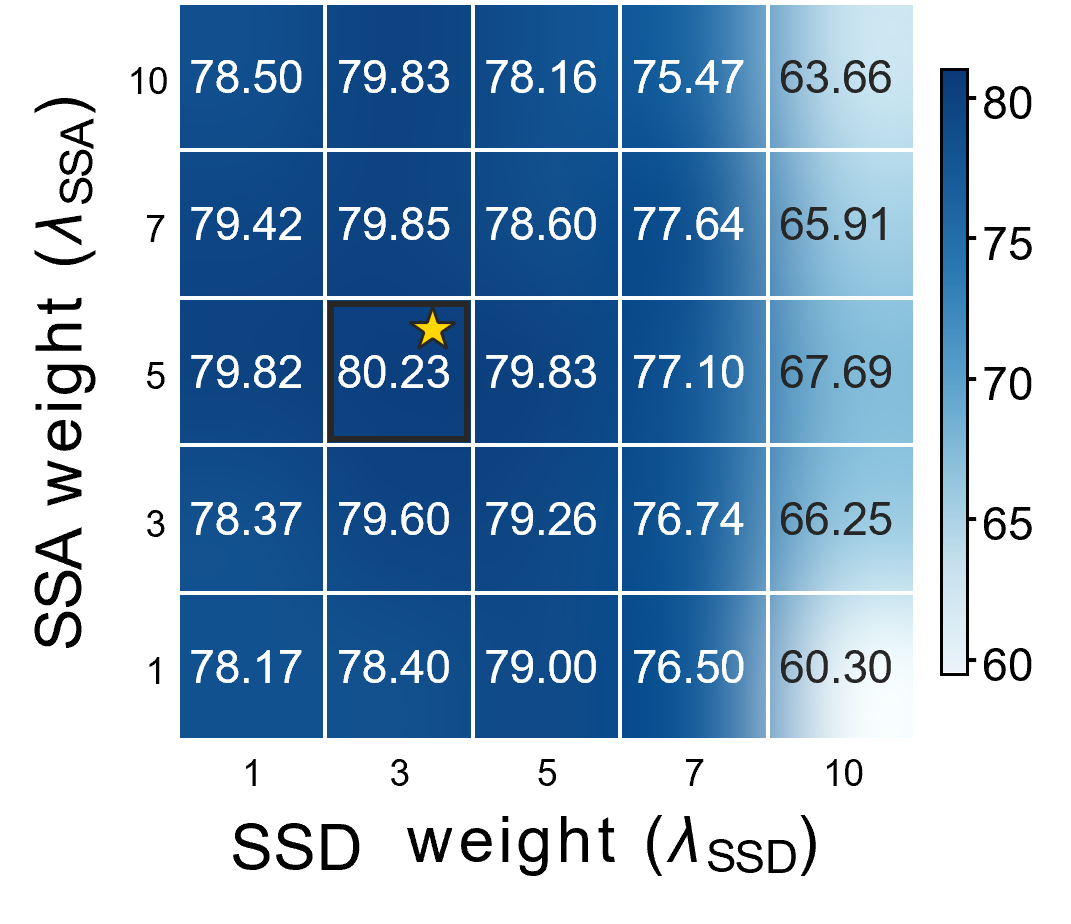}
    \caption{Loss weights.}
    \label{fig:hyper-lambda}
\end{subfigure}
\par\vspace{-6pt}
\caption{Hyperparameter analysis on CIFAR-100 (CNX-T$\to$CNX-N).
(a) Joint effects of cluster number and top-$r$ in the diverse semantic bank.
(b) Joint effects of synthesis weight $\lambda_{\mathrm{SSA}}$ and distillation weight $\lambda_{\mathrm{SSD}}$.
}
\label{fig:hyper}
\vspace{-5pt}
\end{figure}
\paragraph{Hyperparameter Sensitivity.}
Fig.~\ref{fig:hyper} evaluates hyperparameter sensitivity on CNX-T$\to$CNX-N, demonstrating broad stability across both hyperparameter pairs. For semantic bank sizing (Fig.~\ref{fig:hyper}a), filtering 200 LLM prompts per class down to roughly half ($K{=}7$, $r{=}15$) best balances intra-class diversity against prompt noise. For loss weighting (Fig.~\ref{fig:hyper}b), batch-level $\mathcal{L}_{\mathrm{SSA}}$ remains robust across all tested $\lambda_{\mathrm{SSA}}$ values, whereas sample-wise $\mathcal{L}_{\mathrm{SSD}}$ competes with $\mathcal{L}_{\mathrm{KL}}$ so an excessive $\lambda_{\mathrm{SSD}}$ ($>5$) over-emphasizes spatial layout and degrades classification. Overall, UniDFKD remains stable across a wide range of hyperparameter settings.

\begin{figure}[t]
    \centering
    \begin{subfigure}[t]{0.43\linewidth}
        \centering
        \includegraphics[width=\linewidth]{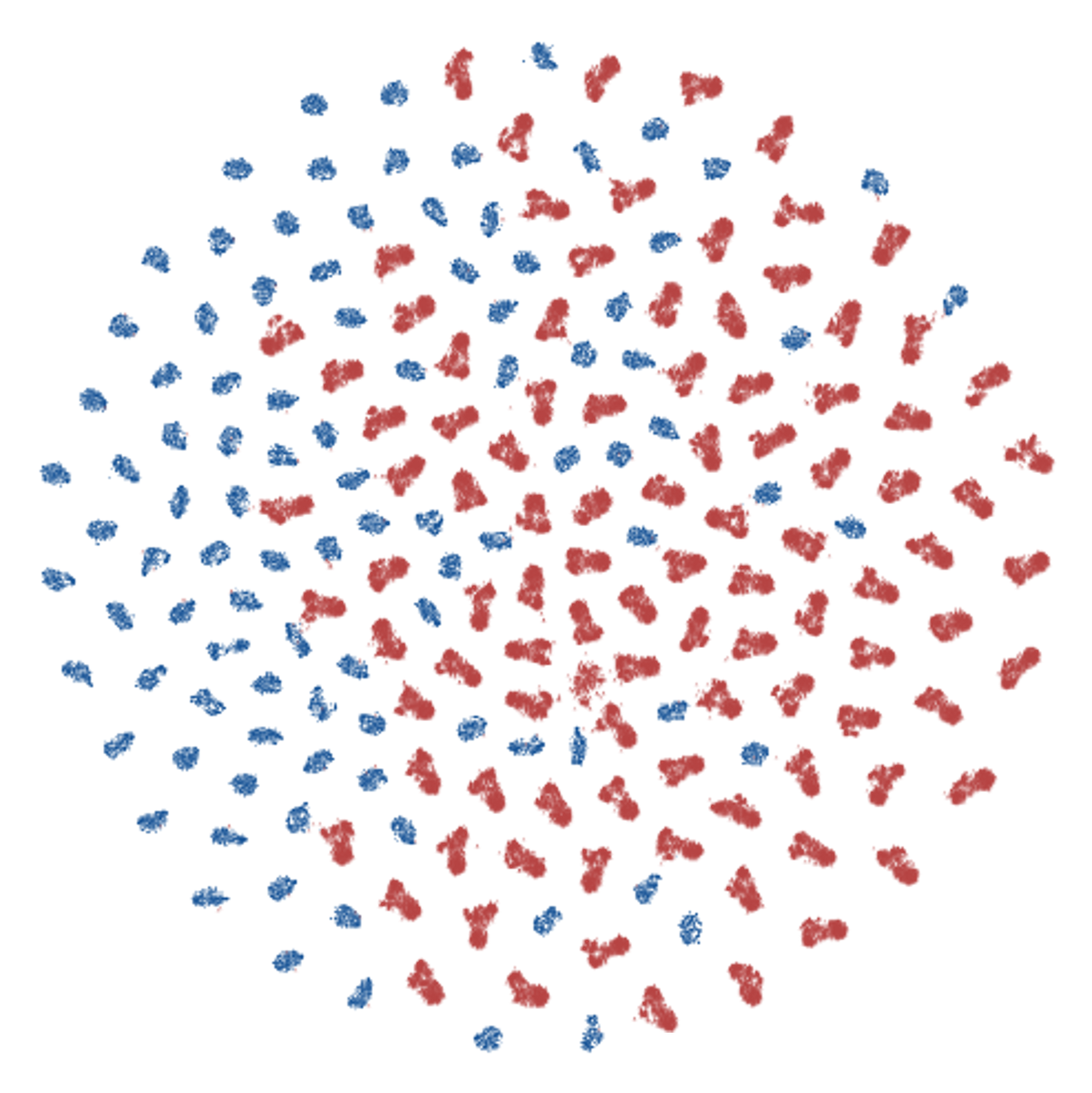}
        \caption{NAYER.}
        \label{fig:tsne-nayer}
    \end{subfigure}
     \hspace{0.05\linewidth}
    \begin{subfigure}[t]{0.43\linewidth}
        \centering
        \includegraphics[width=\linewidth]{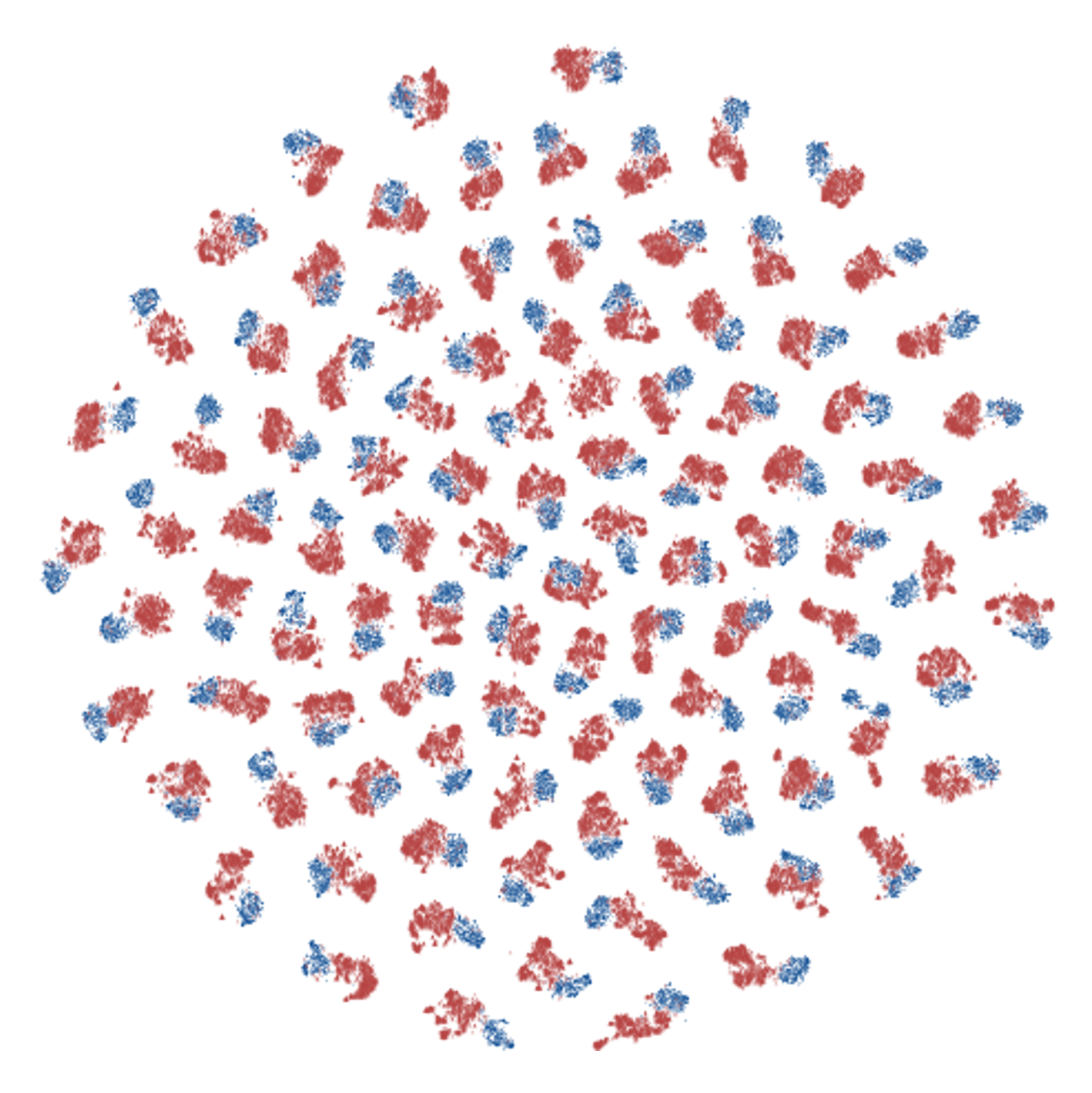}
        \caption{UniDFKD.}
        \label{fig:tsne-ours}
    \end{subfigure}
\par\vspace{-6pt}
    \caption{Coverage of synthetic distribution on CIFAR-100 (CNX-T$\to$ViT-T), visualized by t-SNE of teacher features on training (blue) and synthetic (red) data.}
    \label{fig:tsne}
\vspace{-9pt}
\end{figure}

\begin{figure}[t]
\centering
\includegraphics[width=.35\linewidth]{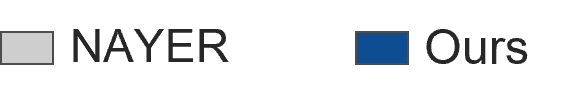}
\\
\begin{subfigure}[t]{0.495\linewidth}
    \centering
    \includegraphics[width=\linewidth]{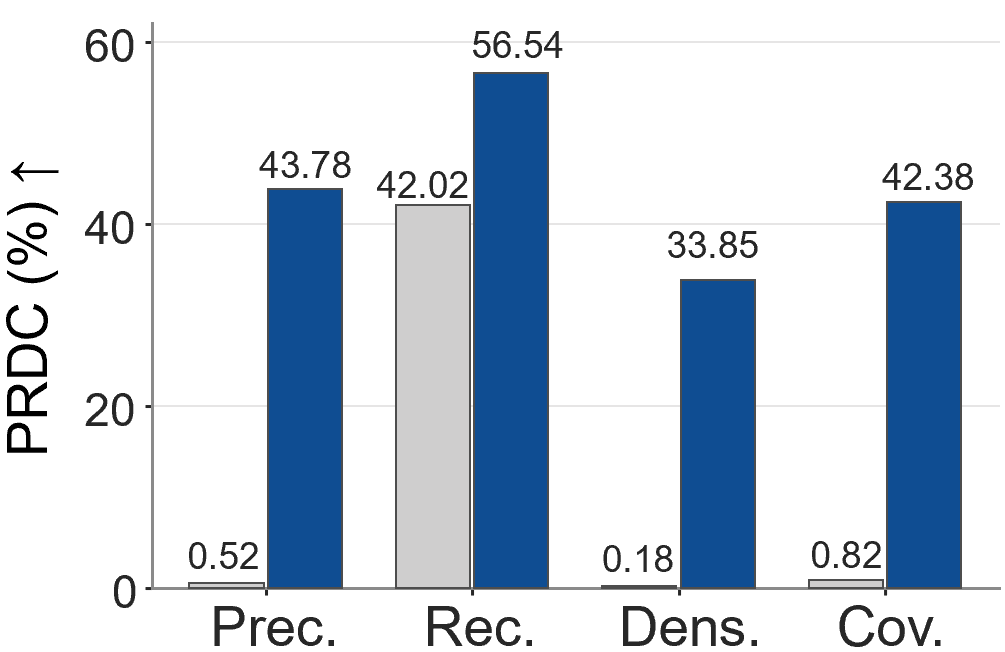}
    \caption{Feature-manifold coverage.}
    \label{fig:PRDC}
\end{subfigure}
\begin{subfigure}[t]{0.495\linewidth}
    \centering
    \includegraphics[width=\linewidth]{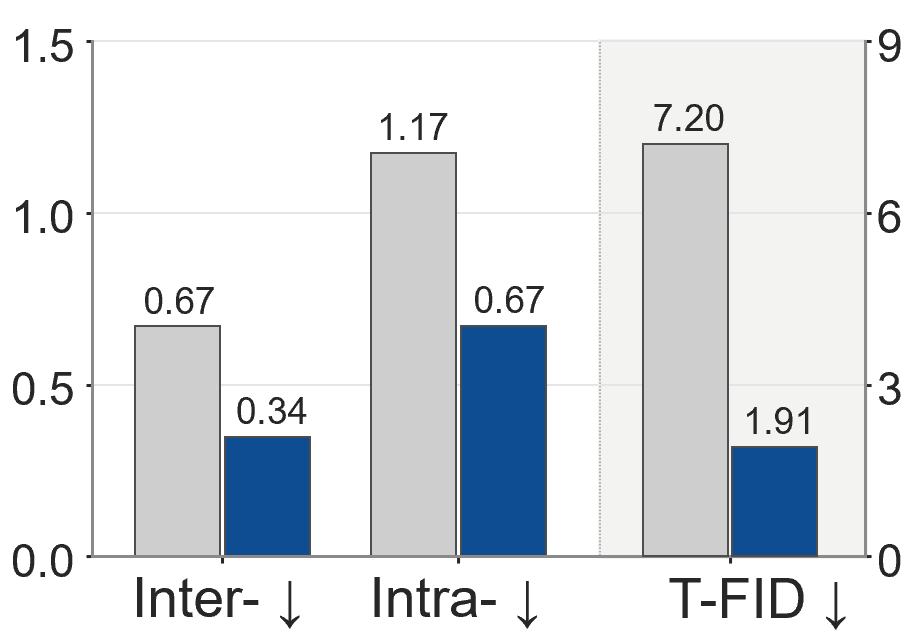}
    \caption{Distributional consistency.}
    \label{fig:FID}
\end{subfigure}
\par\vspace{-6pt}
\caption{Synthetic distribution quality on CIFAR-100 (CNX-T$\to$ViT-T).
(a) PRDC (precision, recall, density and coverage);
(b) inter-class MAE between the inter-class correlation matrices computed from real and synthetic data, intra-class Vendi MAE between their per-class Vendi scores, and T-FID between their teacher-feature distributions.}
\label{fig:dist-quality}
\vspace{-12pt}
\end{figure}

\paragraph{Synthetic Distribution Consistency.}
To verify that our gains stem from a better-aligned synthetic distribution, we evaluate the real and synthetic CIFAR-100 features across visual, manifold, and hierarchical dimensions (Figs.~\ref{fig:tsne} and~\ref{fig:dist-quality}).
Visually, t-SNE projections show that our synthetic clusters closely overlap with and broadly cover the real distribution, correcting the deviations seen in NAYER.
At the manifold level, UniDFKD consistently improves all four PRDC metrics~\cite{naeem2020reliable}, confirming higher fidelity and broader spatial coverage.
At the hierarchical level, UniDFKD achieves substantially lower MAE in both inter-class correlation matrices (0.34 vs.\ 0.67) and per-class Vendi scores~\cite{friedman2023vendi} (0.67 vs.\ 1.17), indicating better preservation of inter-class relations and intra-class diversity, respectively.
Finally, the global discrepancy, measured by the teacher-feature adapted FID (T-FID)~\cite{heusel2017gans}, significantly decreases (1.91 vs.\ 7.20).
Together, these results confirm that UniDFKD synthesizes a distribution that semantically mirrors the real distribution.

%% file: sec/5_conclusion.tex
\section{Conclusion}
This paper unifies data-free knowledge distillation across homogeneous and heterogeneous architectures by shifting from implicit BN statistics to explicit architecture-agnostic semantic priors.
Our layer-wise analysis reveals that the effective content of the BN prior resides in deep semantic representations, motivating the design of \emph{UniDFKD}.
It leverages categorical semantics to dictate \emph{what} to synthesize and spatial semantics to regularize \emph{where} the discriminative evidence is located.
Consequently, semantic distillation resolves \emph{how} to effectively transfer this preserved knowledge from the teacher to the student.
Experiments across CNNs and ViTs under homogeneous and heterogeneous settings demonstrate substantial improvements over prior methods.

%% file: sec/X_suppl.tex
\clearpage
\maketitlesupplementary
\appendix
\setcounter{figure}{0}
\setcounter{table}{0}
\setcounter{equation}{0}
\renewcommand{\thefigure}{S\arabic{figure}}
\renewcommand{\thetable}{S\arabic{table}}
\renewcommand{\theequation}{S\arabic{equation}}
\renewcommand{\theHfigure}{S\arabic{figure}}
\renewcommand{\theHtable}{S\arabic{table}}
\renewcommand{\theHequation}{S\arabic{equation}}
\input{sec/appendix}

%% file: sec/appendix.tex
\section*{Overview}
This supplementary material complements the main paper with the supplementary analysis of the layer-wise motivation, further methodological details, the complete implementation setup, extensive additional quantitative results, qualitative visualizations and extended discussions.
Our source code will be made publicly available upon acceptance.

The contents are organized as follows:

\begin{itemize}
    \item \textbf{Sec.~\ref{app:motivation} (Supplementary Analysis of the Layer-wise Motivation):} Presents the full protocol of the layer-wise diagnosis reported in Fig.~\ref{fig:cmi_bn_layerwise_ablation} of the main paper, extends it to additional architectures and datasets together with t-SNE visualizations of ViT-S and CNX-T statistics on CIFAR-10 and CIFAR-100, and proves that matching the channel statistics of the deepest layer, together with the classification loss, concentrates the class-conditional representation of the synthetic data around that of the real data, whereas a single shallow layer leaves this representation undetermined.
    \item \textbf{Sec.~\ref{app:instantiation} ({Methodology Theoretical Foundations}):} Completes the description of the class-discriminative spatial attribution (CDSA) interface by detailing how the spatial response is recovered for every teacher family considered in this work, i.e., feature-map-based backbones (ResNet, ConvNeXt, Swin) and CLS-token-based backbones (plain ViT) (Sec.~\ref{app:cdsa_instantiation}), and establishes the theoretical guarantees of UniDFKD: the orthogonal ``What''--``Where'' decomposition, the double degeneracy of the classification loss, and how SSA and CSC act on the two components (Sec.~\ref{app:theory_what_where}), together with the synthetic-to-real generalization bound and constructive necessity of SSD (Sec.~\ref{app:theory_ssd}).
    \item \textbf{Sec.~\ref{app:implementation} (Extended Implementation Details):} Provides comprehensive experimental settings, including hyperparameter configurations, optimization strategies, and data augmentation recipes.
    \item \textbf{Sec.~\ref{app:quantitative} (Extensive Quantitative Results):} Reports supplementary quantitative evaluations, including a comparison with state-of-the-art DFKD methods on conventional CNNs, an analysis of computational efficiency (speedup), results on the large-scale ImageNet-1K benchmark under tight synthesis budgets, and the robustness of Categorical Semantic Conditioning (CSC) to vision-agnostic text encoders.
    \item \textbf{Sec.~\ref{app:qualitative} (Qualitative Visualizations):} Showcases visual comparisons of the synthesized images (Sec.~\ref{app:qualitative_synth}) and student spatial attribution alignment under SSD (Sec.~\ref{app:qualitative_student_cdsa}).
    \item \textbf{Sec.~\ref{app:discussion} (Extended Discussions and Limitations):} Clarifies the relationship between our framework and alternative paradigms (e.g., BN-free model inversion, data-free quantization, and diffusion-based synthesis), and frankly analyzes the potential limitations of the center-biased spatial prior.
\end{itemize}

\section{Supplementary Analysis of the Layer-wise Motivation}
\label{app:motivation}
\subsection{Protocol of the Layer-wise Diagnosis in \texorpdfstring{Fig.~\ref{fig:cmi_bn_layerwise_ablation}}{Fig. 2}}
\label{app:layerwise}
We select the representative method CMI~\cite{fang2021contrastive} to conduct our baseline layer-wise experiments.
Unlike NAYER~\cite{tran2024nayer}, which inherently introduces the confounding effects of semantic embeddings, CMI enhances the diversity of synthesized data primarily through an auxiliary generator and contrastive objectives, independent of the BN prior.
By keeping these generative components fixed across experiments during the layer-wise evaluation, we uniquely isolate the matched feature statistics.
Consequently, any variation in student accuracy can be directly attributed to the effectiveness of the statistical priors at different network depths.

In the feature matching process, we strictly define a ``stat hook'' as a channel-wise mean and variance statistical pair on a normalization layer.
To accommodate the intrinsic differences of normalization layers across various network architectures, we design specific feature extraction and matching strategies for ResNet-34 and ViT-S.

\paragraph{Statistic Matching in ResNet-34.}
ResNet-34 consists of 16 BasicBlocks with a hierarchical layout of 3+4+6+3.
For a standard BasicBlock, the main branch contains two BN layers (\texttt{bn1} and \texttt{bn2}), providing 2 stat hooks.
For the first block of Layer 2 to Layer 4, where spatial downsampling occurs, an additional projection BN layer is appended to the shortcut branch, yielding 3 stat hooks each.
Furthermore, the \texttt{stem bn1} at the network entrance is treated as an independent hook in the global (all-stat) mode.
Overall, ResNet-34 comprises 36 BN stat hooks, as detailed in Table~\ref{tab:resnet_hooks}.
\begin{table}[htbp]
\centering
\caption{Distribution of BN-stat hooks in ResNet-34.}
\label{tab:resnet_hooks}
\vspace{-4pt}
\resizebox{\linewidth}{!}{
\begin{tabular}{lcccc}
\toprule
Network Part & \# Blocks & Channels ($C_l$) & Hooks per Block & Subtotal Hooks \\
\midrule
Stem & - & 64 & 1 & 1 \\
Layer 1 & 3 & 64 & 2, 2, 2 & 6 \\
Layer 2 & 4 & 128 & 3, 2, 2, 2 & 9 \\
Layer 3 & 6 & 256 & 3, 2, 2, 2, 2, 2 & 13 \\
Layer 4 & 3 & 512 & 3, 2, 2 & 7  \\
\midrule
\textbf{Total} & \textbf{16} & - & - & \textbf{36}\\
\bottomrule
\end{tabular}
}
\end{table}

To ensure a fair comparison during the preliminary block-wise diagnosis, we employ a non-cumulative constraint: when analyzing a specific block, we exclusively activate the two stat (\texttt{bn1} and \texttt{bn2}) hooks from its main branch, regardless of whether it contains a shortcut branch, while disabling constraints from all other layers.
Isolated nodes, such as the \texttt{stem bn1}, are also excluded from this single-block evaluation.
Conversely, for the global all-stat experiment, we adhere to the standard baseline protocol by utilizing all 36 hooks.

For a given BN input activation $A_l \in \mathbb{R}^{B \times C_l \times H_l \times W_l}$, we calculate the channel-wise moments across the batch and spatial dimensions:
\begin{equation}
\begin{aligned}
\hat{\mu}_{l,c} &= \frac{1}{B H_l W_l} \sum_{b=1}^{B} \sum_{h=1}^{H_l} \sum_{w=1}^{W_l} A_{l,b,c,h,w} \\
\hat{\sigma}_{l,c}^2 &= \frac{1}{B H_l W_l} \sum_{b=1}^{B} \sum_{h=1}^{H_l} \sum_{w=1}^{W_l} (A_{l,b,c,h,w} - \hat{\mu}_{l,c})^2. 
\end{aligned}
\end{equation}

These statistics are directly aligned with the BN running moments saved in the teacher model's checkpoint.
The matching function for the ResNet architecture is directly applied to the raw variance, formulated as follows:
\begin{equation}
\mathcal{L}_{\mathrm{stat}}^{\mathrm{BN}} = \sum_{l\in\mathcal{H}} \Bigl( \| \hat{\mu}_{l} - \mu_l^{\mathrm{BN}} \|_2 + \| \hat{\sigma}_l^2 - {(\sigma_l^{\mathrm{BN}})}^2 \|_2 \Bigr)
\end{equation}
where $\mathcal{H}$ denotes the set of currently activated stat hooks (e.g., all 36 hooks in the global matching mode, or a specific subset during the block-wise diagnostic experiments).

\paragraph{Statistic Matching in ViT-S.}
Unlike BN, LN does not track running statistics during training.
Therefore, we utilize real data to pre-compute the reference moments offline.
Specifically, the pre-trained ViT-S teacher model is set to \texttt{eval()} mode, and forward propagation is performed on the CIFAR-10 training set (50,000 images).
Given an input resolution of $224 \times 224$ and a patch size of 16, a single image yields 197 tokens (196 patch tokens and 1 CLS token).
We register a hook at each LN layer to intercept its input features $A_l \in \mathbb{R}^{B \times N \times D}$ and flatten them into $A_l^{\mathrm{flat}} \in \mathbb{R}^{(BN) \times D}$.
The population mean and variance are computed along the sample and token dimensions: \begin{equation}
\begin{aligned}
\mu_{l,d}^{\mathrm{real}} &= \frac{1}{B N} \sum_{b=1}^{B} \sum_{n=1}^{N} A_{l,b,n,d} \\
(\sigma_{l,d}^{\mathrm{real}})^2 &= \frac{1}{B N} \sum_{b=1}^{B} \sum_{n=1}^{N} (A_{l,b,n,d} - \mu_{l,d}^{\mathrm{real}})^2.
\end{aligned}
\end{equation}
ViT-S contains 12 Transformer blocks, and each block corresponds to 2 stat hooks (\texttt{norm1} and \texttt{norm2}), totaling 24 hooks.
Including the final \texttt{norm} layer, the global mode comprises 25 LN hooks.

Similarly, we apply a strict non-cumulative constraint for ViT-S: when analyzing a specific Transformer block, we exclusively activate its two corresponding stat hooks (\texttt{norm1} and \texttt{norm2}) while disabling constraints from all other layers.
The isolated \texttt{final norm} layer is consequently excluded from this single-block evaluation.
Conversely, for the global all-stat experiment, we use all 25 LN hooks.

Consistent with the BN alignment approach in ResNet, we directly match the raw variance and formulate the statistical loss for ViT-S as follows:
\begin{equation}\mathcal{L}_{\mathrm{stat}}^{\mathrm{LN}} = \sum_{l\in\mathcal{H}} \Bigl( \| \hat{\mu}_l - \mu_l^{\mathrm{real}} \|_2 + \| \hat{\sigma}_l^2 - (\sigma_l^{\mathrm{real}})^2 \|_2 \Bigr).
\end{equation}

\paragraph{t-SNE of Per-image Statistics.}
For the t-SNE visualizations, real test images of CIFAR-10 and CIFAR-100 are resized to $224\times224$ and passed through the frozen ViT-S teacher in \texttt{eval()} mode.
At the \texttt{norm1} input of Blocks 2, 5, 8, and 11, each image yields the channel-wise mean and population standard deviation over its 197 tokens (196 patch tokens and the CLS token), and the two vectors are concatenated into one descriptor.
The descriptors of each block are embedded independently by t-SNE with Euclidean distance, perplexity 30, 1,000 iterations, PCA initialization, and random seed 42, without further normalization.
The CIFAR-10 panel uses all 10,000 test images.
The CIFAR-100 panel uses 100 test images from each of its 100 classes, 10,000 in total.
The same procedure is applied to the CNX-T teacher, whose statistics are taken at the \texttt{norm} input of the last block of each of its four stages and computed over the spatial positions of the feature map.
Likewise, for the RN34 teacher in Fig.~\ref{fig:resnet_bn_statistics_tsne} of the main paper, native $32\times32$ CIFAR-10 test images are passed through the frozen model and the same procedure is applied to the spatial channel-wise BN statistics at the last block of each stage (Blocks 2, 6, 12, 15).

\subsection{Extended Layer-wise Analysis}
\begin{figure*}[t]
    \centering
    \begin{minipage}[t]{0.485\linewidth}
        \centering
        \begin{subfigure}{0.78\linewidth}
            \centering
            \includegraphics[width=\linewidth]
            {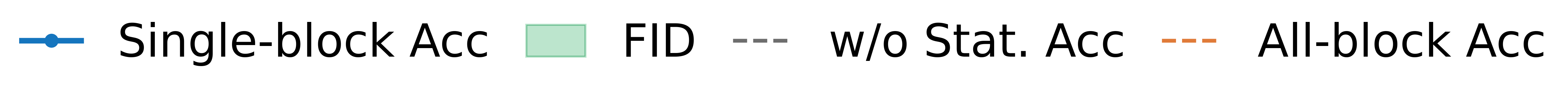}
        \end{subfigure}

        \vspace{0.3em}

        \begin{subfigure}[t]{0.45\linewidth}
            \centering
            \includegraphics[width=\linewidth]
            {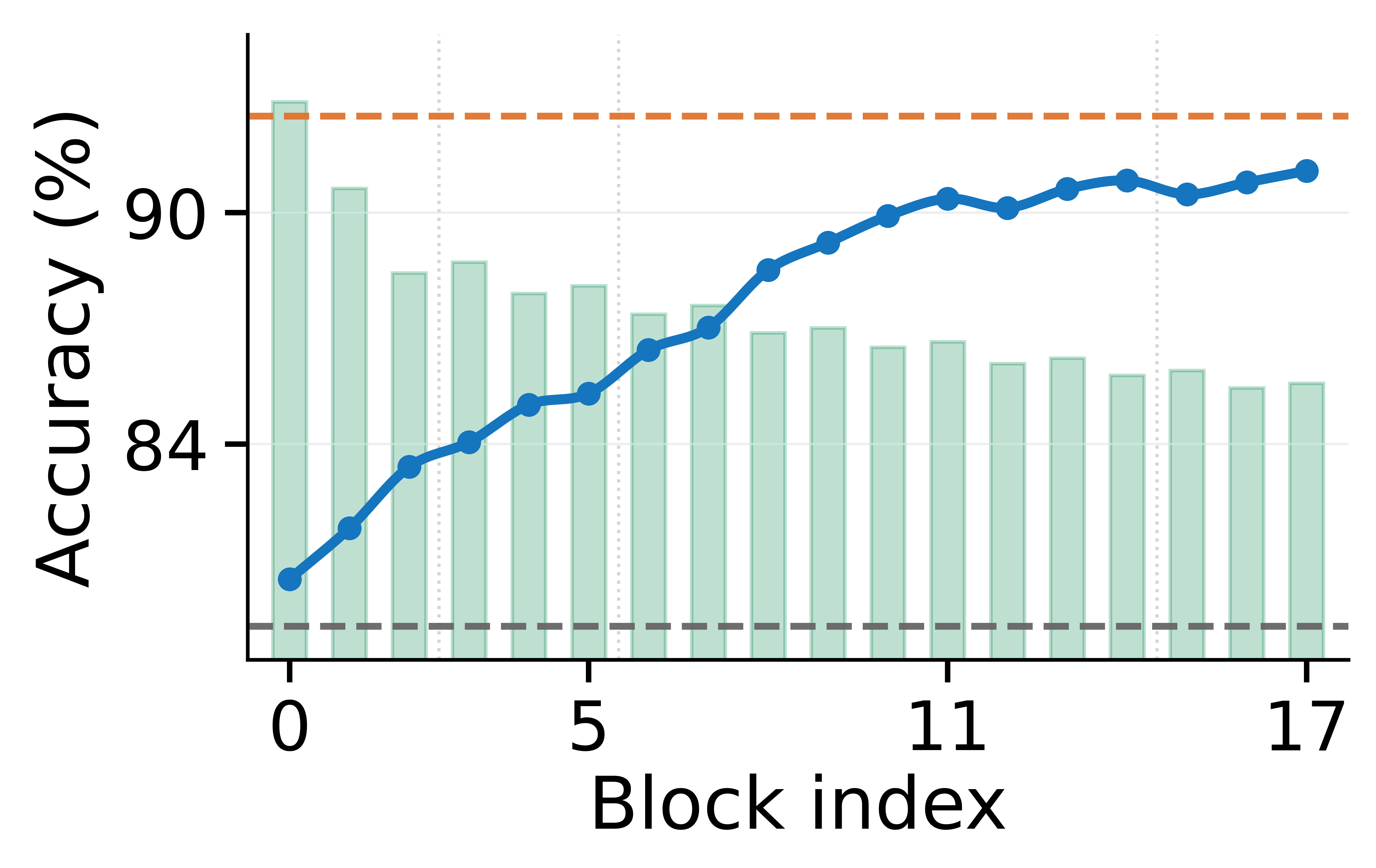} \caption{CNX-T$\to$CNX-N}
            \label{fig:appendix_c10_cnx}
        \end{subfigure}
        \hfill
        \begin{subfigure}[t]{0.45\linewidth}
            \centering
            \includegraphics[width=\linewidth]
            {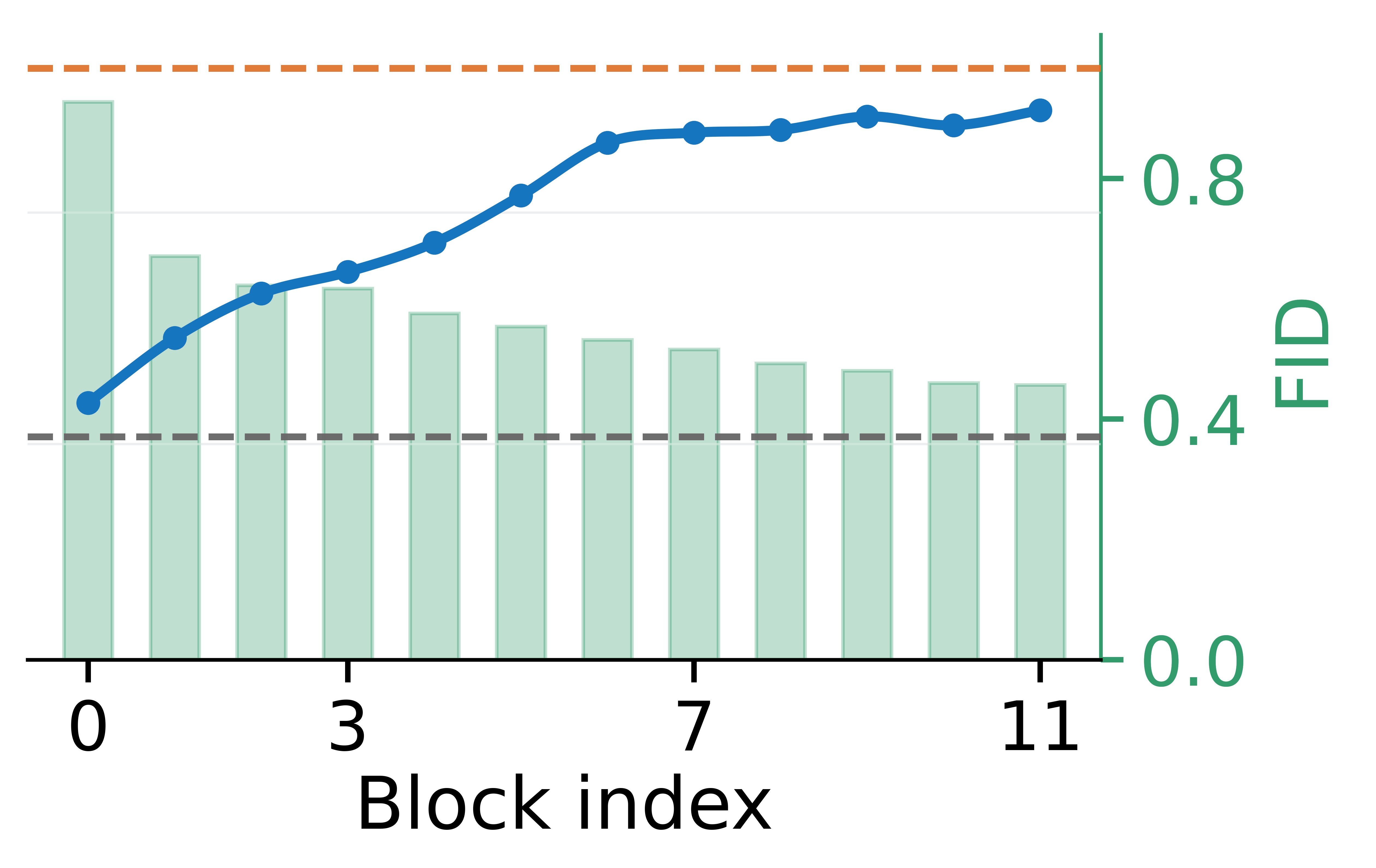} \caption{ViT-S$\to$ViT-T}
            \label{fig:appendix_c10_deit}
        \end{subfigure}

        \caption{ Layer-wise analysis on CIFAR-10.
        We report the single-block accuracy and T-FID obtained using statistics from different teacher blocks, together with the accuracy without statistical guidance and the all-block result.
        }
        \label{fig:appendix_layerwise_c10}
    \end{minipage}
    \hfill
    \begin{minipage}[t]{0.485\linewidth}
        \centering
        \begin{subfigure}{0.78\linewidth}
            \centering
            \includegraphics[width=\linewidth]
            {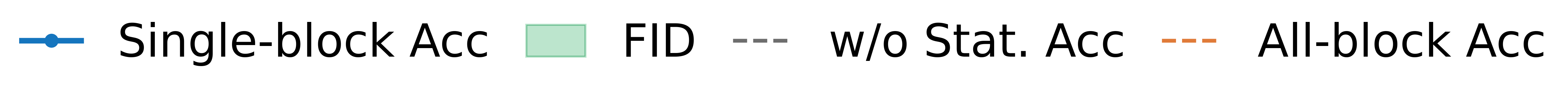}
        \end{subfigure}

        \vspace{0.3em}

        \begin{subfigure}[t]{0.45\linewidth}
            \centering
            \includegraphics[width=\linewidth]
            {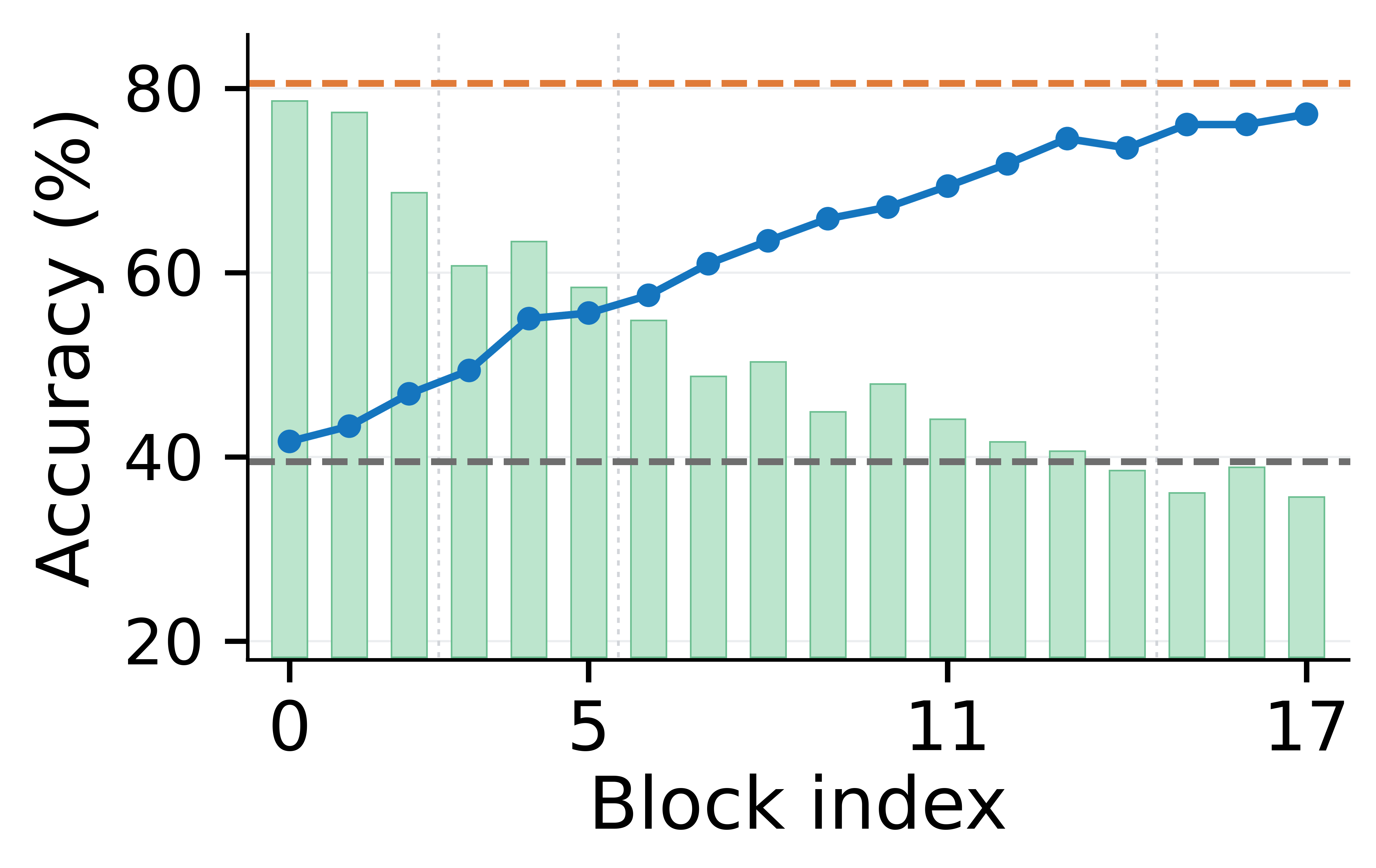} \caption{CNX-T$\to$CNX-N}
            \label{fig:appendix_in100_cnx}
        \end{subfigure}
        \hfill
        \begin{subfigure}[t]{0.45\linewidth}
            \centering
            \includegraphics[width=\linewidth]
            {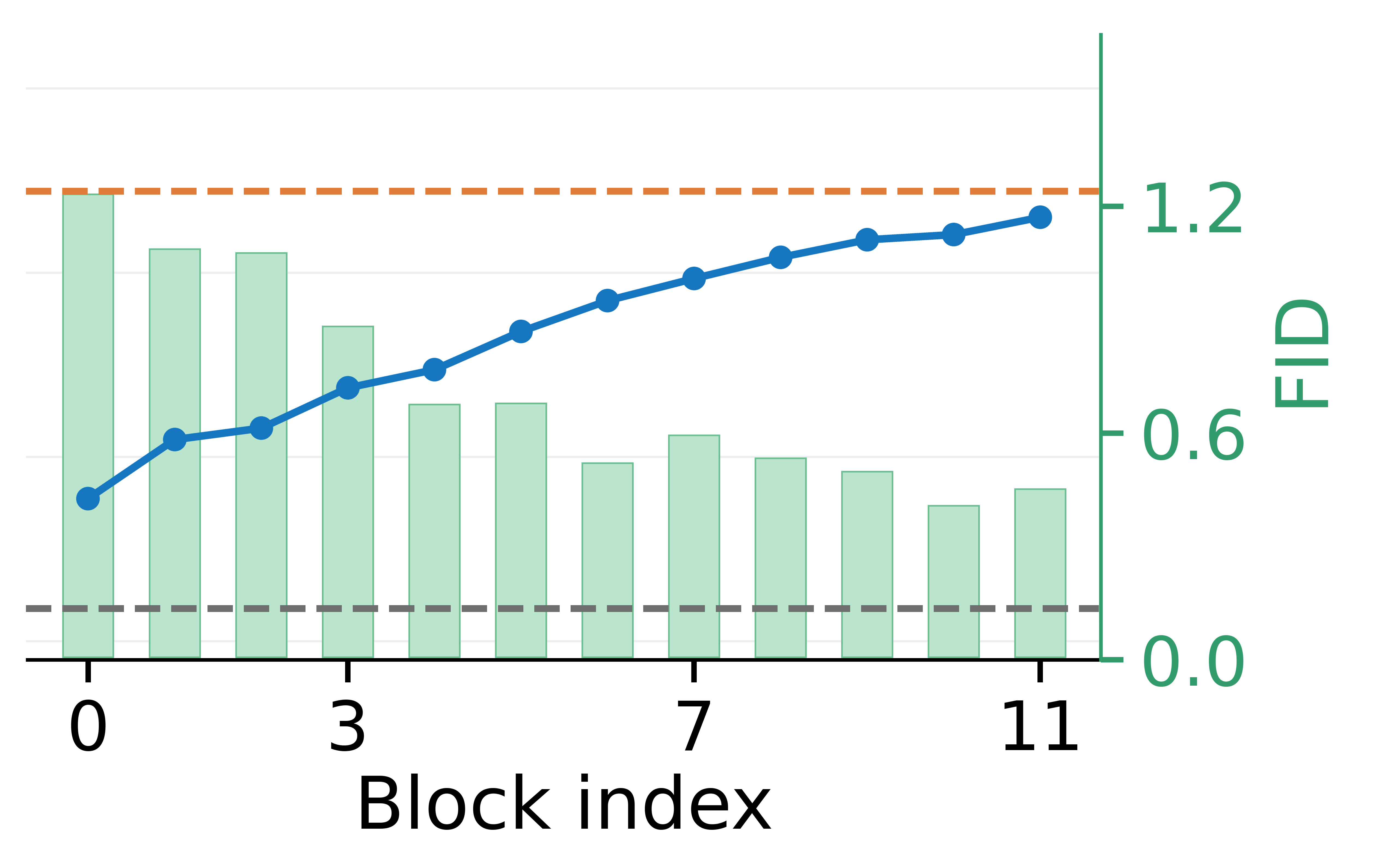} \caption{ViT-S$\to$ViT-T}
            \label{fig:appendix_in100_deit}
        \end{subfigure}

        \caption{ Layer-wise analysis on ImageNet-100.
        We report the single-block accuracy and T-FID obtained using statistics from different teacher blocks, together with the accuracy without statistical guidance and the all-block result.
        }
        \label{fig:appendix_layerwise_in100}
    \end{minipage}
\end{figure*}
To further validate our preliminary observations, Fig.~\ref{fig:appendix_layerwise_c10} and Fig.~\ref{fig:appendix_layerwise_in100} extend the layer-wise analysis to additional modern architectures (ConvNeXt and ViT) across CIFAR-10 and the more challenging ImageNet-100 dataset.
The diagnostic protocol strictly follows the one described in Sec.~\ref{app:layerwise}.

Consistent with the findings in the main paper, the results demonstrate a clear and universal trend across different architectures and datasets.
Statistics from shallow blocks provide only marginal improvements over the unconstrained baseline (w/o Stat. Acc).
As the block index increases, however, the single-block distillation accuracy rises progressively while the T-FID steadily decreases.
Notably, matching the statistics of the deepest layers alone can already recover most of the performance obtained by matching all blocks combined.

This extended analysis provides strong empirical evidence that the phenomenon is not restricted to specific models or scales.
It confirms that the true utility of statistical priors fundamentally lies in their implicit alignment of deep, class-discriminative semantic representations, which further justifies our motivation to construct an explicit, architecture-agnostic semantic prior.

\begin{figure*}[t]
    \centering
    \begin{subfigure}[t]{0.46\linewidth}
        \centering
        \includegraphics[width=\linewidth]{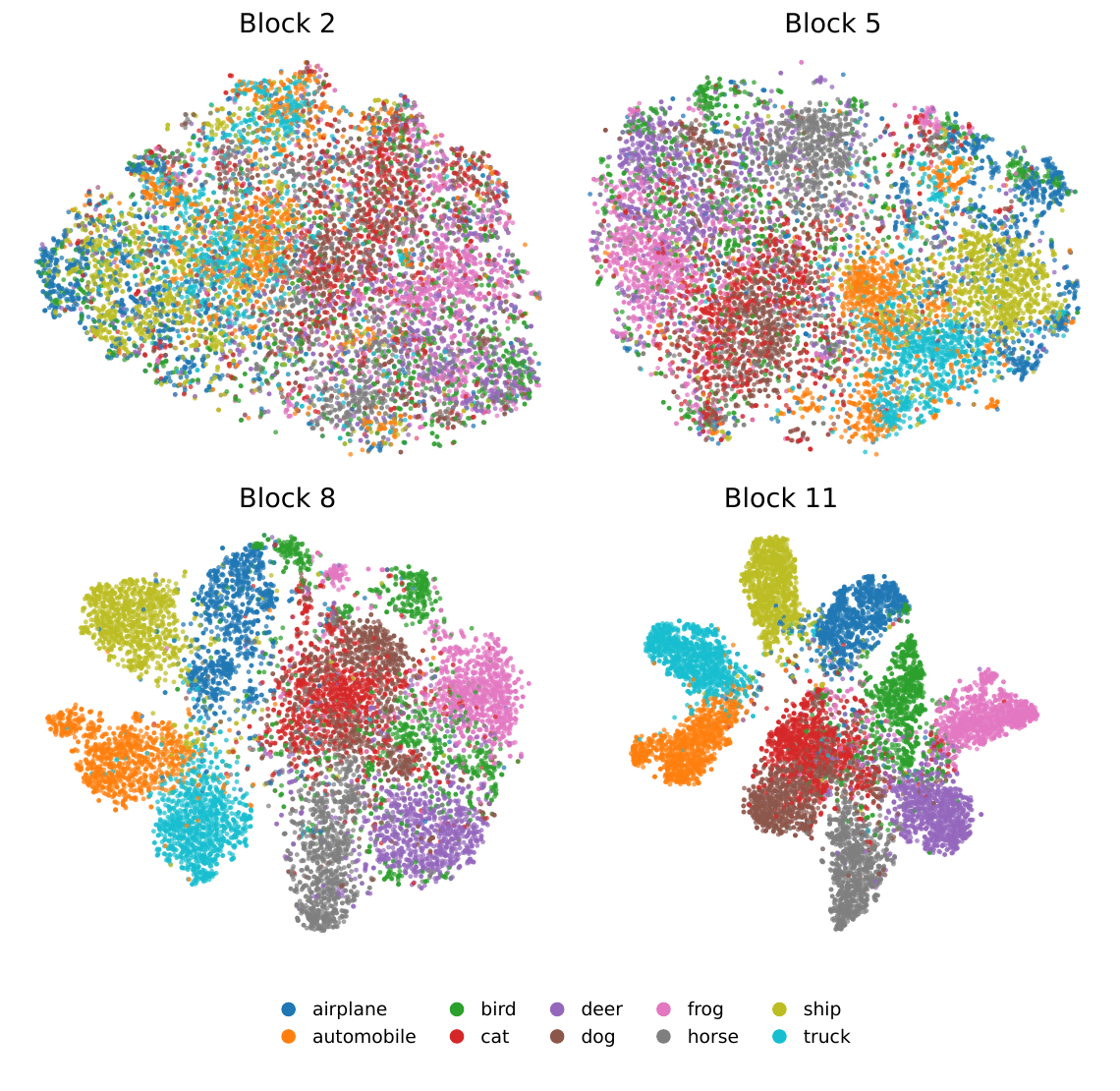}
        \caption{CIFAR-10 (10 classes).}
        \label{fig:appendix_vit_tsne_c10}
    \end{subfigure}\hfill
    \begin{subfigure}[t]{0.46\linewidth}
        \centering
        \includegraphics[width=\linewidth]{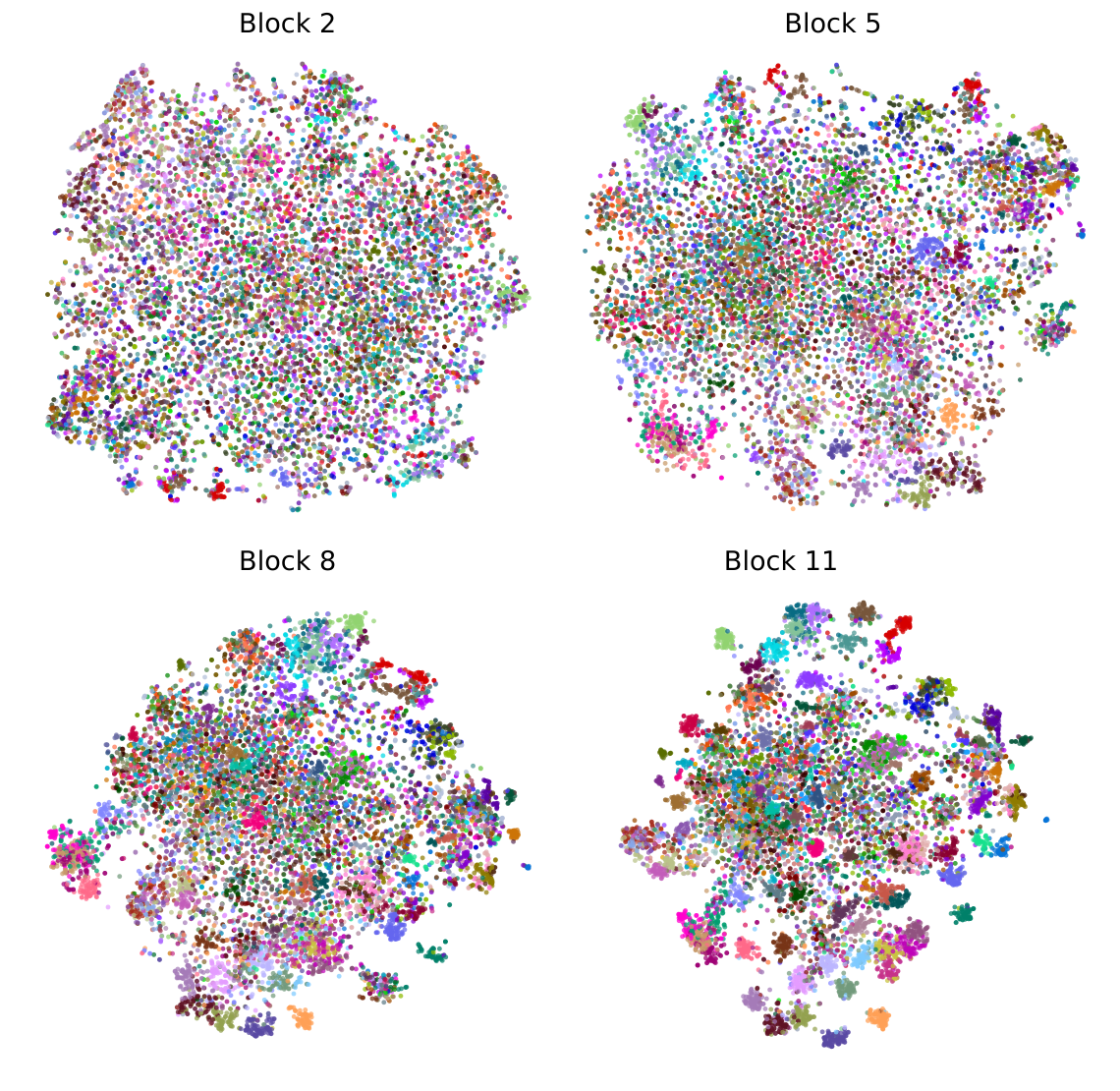}
        \caption{CIFAR-100 (100 classes).}
        \label{fig:appendix_vit_tsne_c100}
    \end{subfigure}
    \caption{t-SNE of ViT-S statistics on real images.
    Each point is one image, represented by the concatenated channel-wise mean and standard deviation of the \texttt{norm1} input at the indicated block and colored by class.
    Class structure emerges progressively with depth and forms compact class clusters at Block 11.
    CIFAR-100 uses one color per class, with the 100 classes drawn from a fixed palette.}
    \label{fig:appendix_vit_tsne}
\end{figure*}

\paragraph{t-SNE of ViT-S statistics.}
Fig.~\ref{fig:appendix_vit_tsne} extends the t-SNE analysis in Fig.~\ref{fig:resnet_bn_statistics_tsne} of the main paper to the ViT-S teacher on real CIFAR images.
Each point is one image, described by the channel-wise mean and standard deviation of the \texttt{norm1} input at Blocks 2, 5, 8, and 11.
On CIFAR-10, the shallow blocks mix all ten classes.
Class groups emerge at Block 5, where ship, automobile, truck, and frog already form compact regions, and become well-defined clusters at Block 8.
At Block 11 the ten classes occupy ten separate clusters.
The full 100-class CIFAR-100 test set follows the same progression: all classes are intermixed at Block 2, class-specific groups appear at Blocks 5 and 8, and at Block 11 many classes form compact, isolated clusters.
The class-discriminative structure of ViT-S statistics is therefore concentrated in the deep blocks, as it is for the BN statistics of RN34.
This matches the single-block accuracy and T-FID trends in Fig.~\ref{fig:appendix_layerwise_c10} and supports building the semantic prior on deep, class-discriminative representations

\begin{figure*}[t]
    \centering
    \begin{subfigure}[t]{0.46\linewidth}
        \centering
        \includegraphics[width=\linewidth]{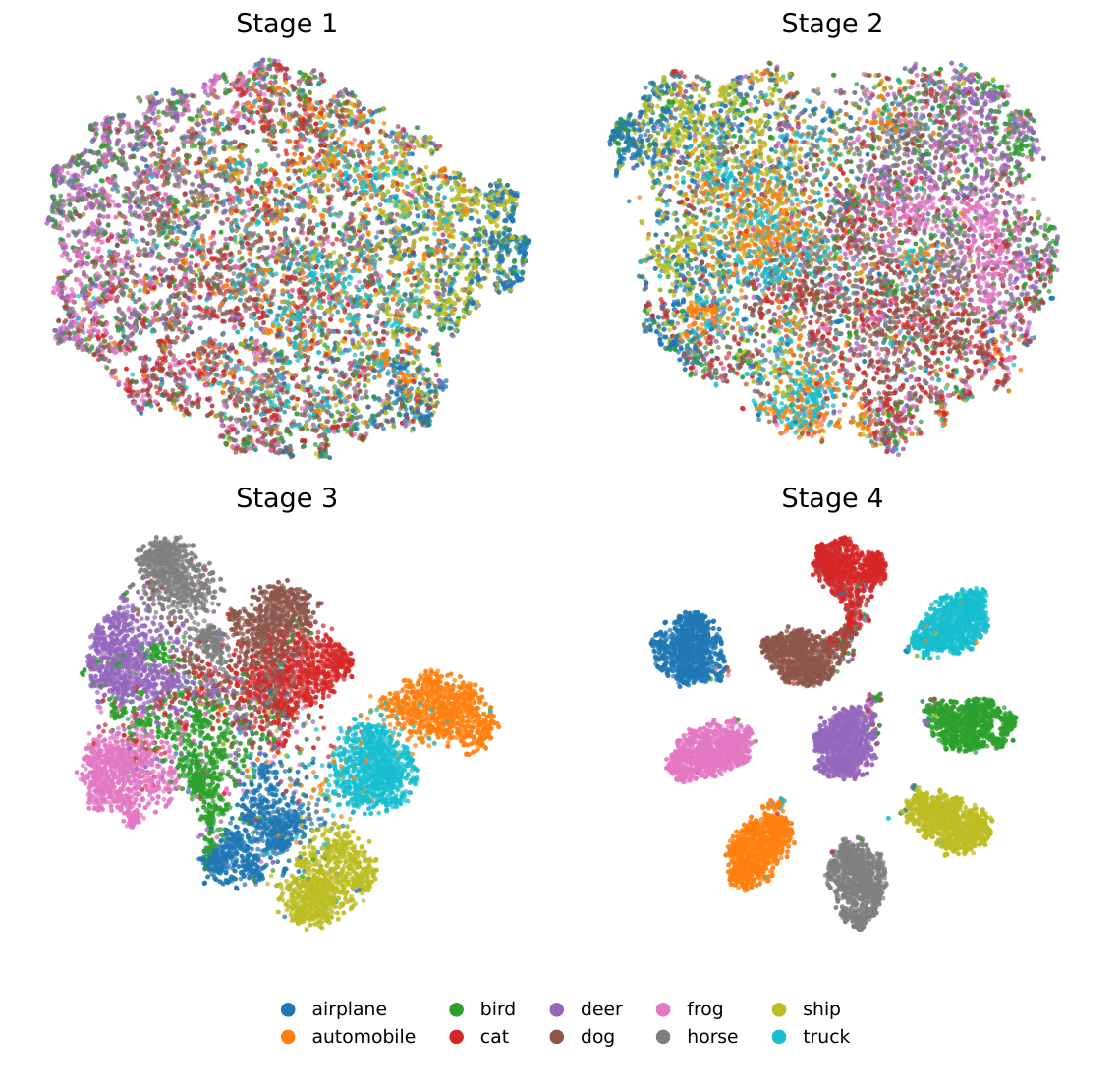}
        \caption{CIFAR-10 (10 classes).}
        \label{fig:appendix_cnx_tsne_c10}
    \end{subfigure}\hfill
    \begin{subfigure}[t]{0.46\linewidth}
        \centering
        \includegraphics[width=\linewidth]{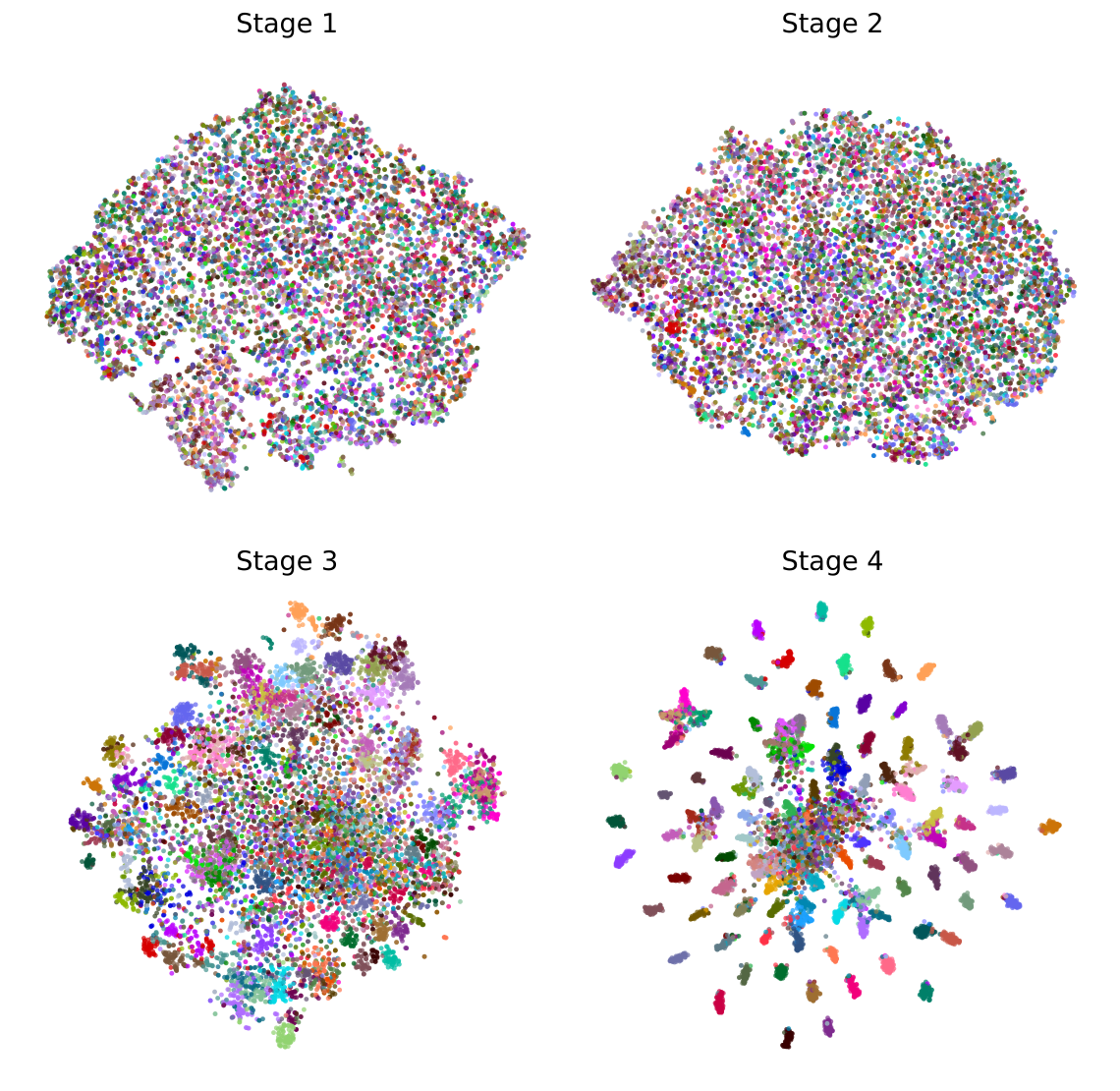}
        \caption{CIFAR-100 (100 classes).}
        \label{fig:appendix_cnx_tsne_c100}
    \end{subfigure}
    \caption{t-SNE of CNX-T statistics on real images.
    Each point is one image, represented by the concatenated channel-wise mean and standard deviation of the \texttt{norm} input at the end of the indicated stage and colored by class.
    Class structure emerges progressively with depth and forms compact class clusters at Stage 4.}
    \label{fig:appendix_cnx_tsne}
\end{figure*}

\paragraph{t-SNE of CNX-T statistics.}
Fig.~\ref{fig:appendix_cnx_tsne} repeats the analysis for the ConvNeXt-Tiny teacher, using the \texttt{norm} input at the end of each of its four stages.
The progression matches ViT-S.
On CIFAR-10, Stages 1 and 2 mix all ten classes, Stage 3 already separates most of them, and Stage 4 places the ten classes in ten isolated clusters.
On CIFAR-100, Stages 1 and 2 are fully intermixed, class-specific groups appear at Stage 3, and at Stage 4 most of the 100 classes form compact, isolated clusters.
ViT-S and CNX-T, two LN-based architectures with different structure, therefore concentrate class-discriminative statistics in their deepest layers.

\subsection{Theoretical Analysis of Statistic Matching}
\label{app:theory}
\input{fig/theory_schematic}
Sec.~\ref{subsec:why-bn} of the main paper reports two observations on the role of BN statistics at different depths (Fig.~\ref{fig:cmi_bn_layerwise_ablation}).
Deep-layer statistics form class-wise clusters, while shallow-layer statistics mix all classes.
Matching the statistics of the deepest block alone approaches matching all blocks, while matching a shallow block gains little over unconstrained synthesis.
From these observations the main paper concludes that BN statistics are effective because of the deep semantic prior they enforce.

This subsection explains why depth makes this difference.
We apply the matching loss of Eq.~\eqref{eq:bn-stat} of the main paper to one layer at a time.
Proposition~\ref{prop:deep-concentration} shows that deep statistics constrain the synthetic features in every direction, including the directions that the classification loss leaves free (Fig.~\ref{fig:theory_schematic_a}).
The intuition follows from standard variance decomposition~\cite{bishop2006pattern}.
The total feature variance equals the sum of the channel variances.
It splits into three non-negative parts: between-class variance, within-class variance, and within-image token variance.
At the deepest layer, real features exhibit Neural Collapse~\cite{papyan2020prevalence}.
Each class clusters tightly around a prototype aligned with its classifier weight, and tokens stay close to their image mean.
The real total variance therefore comes almost entirely from these class prototypes.
For synthetic data, matching channel means and variances fixes the global center and total variance to the real values.
Meanwhile, a low classification loss forces synthetic features to align with the classifier weights as strongly as real features do.
This alignment alone accounts for nearly the entire total variance.
Little variance remains for within-class scatter, token spread, or class means shifted off the real prototypes.
The synthetic features of each class therefore concentrate around the real class prototype.
Proposition~\ref{prop:shallow-cooccurrence} shows that shallow statistics, together with an arbitrarily low classification loss, can be met by synthetic images whose class knowledge does not transfer to real data: a student distilled on them stays at chance accuracy.
These statistics record which low-level features occur, but not which features occur together, and the deeper layers read exactly these combinations.
The closing paragraph, ``Relation to Fig.~\ref{fig:cmi_bn_layerwise_ablation}'', connects both results to the curves of the main paper, and Sec.~\ref{app:layerwise} gives the empirical protocol.
Each result is followed by a short informal explanation, so the proofs can be skipped on a first reading.

\paragraph{Setup.}
Let $P$ be the real distribution tracked by the teacher statistics and $Q$ a synthetic distribution, with samples $(X,Y)\sim P$ and $(\widehat X,Y)\sim Q$ and balanced labels over $C\ge2$ classes.
At layer $l$ the teacher exposes an activation matrix $\mathbf{A}_l(x)\in\mathbb{R}^{N_l\times D_l}$ with token rows $\mathbf{a}_{l,n}(x)$.
For a distribution $R$ the channel statistics are
\begin{equation}
\begin{gathered}
\mu_{l,d}(R)=\mathbb{E}_{R}\Big[\tfrac{1}{N_l}\textstyle\sum_{n}A_{l,n,d}\Big],\\
\sigma^2_{l,d}(R)=\mathbb{E}_{R}\Big[\tfrac{1}{N_l}\textstyle\sum_{n}\big(A_{l,n,d}-\mu_{l,d}(R)\big)^2\Big],
\end{gathered}
\end{equation}%
and the single-layer loss is $\mathcal{L}_{\mathrm{stat}}^{(l)}(Q,P)=\|\boldsymbol{\mu}_l(Q)-\boldsymbol{\mu}_l(P)\|_2+\|\boldsymbol{\sigma}^2_l(Q)-\boldsymbol{\sigma}^2_l(P)\|_2$.
This loss depends only on the $2D_l$ channel means and variances, so it reads only the per-channel marginals of the activations.
To compare full feature distributions, we use the squared 2-Wasserstein distance: for two distributions $R,R'$ on a Euclidean space, $W_2^2(R,R')$ is the smallest value of $\mathbb{E}\|U-U'\|_2^2$ over all joint laws of $U\sim R$ and $U'\sim R'$, and $W_2^2(R,\delta_{\mathbf{a}})=\mathbb{E}_{U\sim R}\|U-\mathbf{a}\|_2^2$ for a point mass $\delta_{\mathbf{a}}$.
Between the channel-independent Gaussians $\mathcal{N}(\boldsymbol{\mu}_l,\mathrm{diag}(\boldsymbol{\sigma}_l^2))$ of $Q$ and $P$, $W_2^2$ equals $\|\boldsymbol{\mu}_l(Q)-\boldsymbol{\mu}_l(P)\|_2^2+\|\boldsymbol{\sigma}_l(Q)-\boldsymbol{\sigma}_l(P)\|_2^2$, where $\boldsymbol{\sigma}_l$ collects the channel standard deviations~\cite{dowson1982frechet,givens1984class}.
Like $\mathcal{L}_{\mathrm{stat}}^{(l)}$, this distance sees only the channel means and spreads.

At the deepest layer $l=L$ (the penultimate representation), let $\mathbf{z}(x)=\frac{1}{N_L}\sum_n\mathbf{a}_{L,n}(x)\in\mathbb{R}^{D_L}$ and $\mathcal{T}(x)=\mathbf{W}^\top\mathbf{z}(x)+\mathbf{b}$.
Softmax ignores a common shift of all logits, so the cross-entropy reads the logits only through the centered logits $\Pi_C\mathcal{T}(x)$, with $\Pi_C=I_C-\frac1C\mathbf{1}_C\mathbf{1}_C^\top$, and the head only through the centered head $\mathring{\mathbf{W}}=\mathbf{W}\Pi_C$.
Let $\mathring{\mathbf{w}}_y$ be the $y$-th column of $\mathring{\mathbf{W}}$, $\bar w^2=\frac1C\sum_y\|\mathring{\mathbf{w}}_y\|_2^2$, and $\tilde{\mathbf{z}}=\mathbf{z}-\boldsymbol{\mu}_L(P)$.
We describe the trained teacher at this layer by two real-data quantities.
For some $\alpha>0$, let
\begin{equation}
\begin{gathered}
\varepsilon_{\mathrm{NC}}=\mathbb{E}_{P}\big\|\tilde{\mathbf{z}}(X)-\alpha\mathring{\mathbf{w}}_Y\big\|_2^2,\\
\varepsilon_{\mathrm{sp}}=\mathbb{E}_{P}\Big[\tfrac{1}{N_L}\textstyle\sum_n\|\mathbf{a}_{L,n}(X)-\mathbf{z}(X)\|_2^2\Big]
\end{gathered}
\end{equation}%
both be small.
A small $\varepsilon_{\mathrm{NC}}$ is Neural Collapse~\cite{papyan2020prevalence,fang2021exploring}: the pooled features of each class concentrate at their class mean, and the centered class means align with the columns $\alpha\mathring{\mathbf{w}}_y$ of the head.
A small $\varepsilon_{\mathrm{sp}}$ means that the tokens of each image stay close to their spatial mean.
Since the columns $\mathring{\mathbf{w}}_y$ sum to zero, the class prototypes $\boldsymbol{\mu}_L(P)+\alpha\mathring{\mathbf{w}}_y$ have between-class variance $\alpha^2\bar w^2$, so a small $\varepsilon_{\mathrm{NC}}$ also means that almost all of the variance of the pooled real feature is between-class variance.
At the prototype $\tilde{\mathbf{z}}=\alpha\mathring{\mathbf{w}}_y$, the centered head assigns class $y$ the score $\langle\mathring{\mathbf{w}}_y,\alpha\mathring{\mathbf{w}}_y\rangle=\alpha\|\mathring{\mathbf{w}}_y\|_2^2$, whose class average is $\alpha\bar w^2$.
The head is arbitrary: it need not be an equiangular simplex, so the inter-class similarities carried by $\mathring{\mathbf{W}}$ are kept.
For a distribution $R$ of input--label pairs, $\mathcal{L}_{\mathrm{CE}}(R)=\mathbb{E}_{(x,y)\sim R}[\ell_{\mathrm{CE}}(\mathcal{T}(x),y)]$ is the cross-entropy of the teacher, with $\ell_{\mathrm{CE}}(\mathbf{u},y)=-\log(e^{u_y}/\sum_ce^{u_c})=\log(1+\sum_{c\ne y}e^{u_c-u_y})$.
Finally, let $\psi(s)=\log\!\big(1+(C-1)e^{-\frac{C}{C-1}s}\big)$.
It is the smallest cross-entropy that a zero-sum logit vector with target entry $s$ can reach, attained when all $C-1$ non-target entries equal $-s/(C-1)$, so that each $u_c-u_y=-\frac{C}{C-1}s$ (Step~2 below).
It is strictly decreasing and convex, and we write $\psi^{-1}$ for its inverse on $(0,\infty)$, so reaching a loss of at most $\mathcal{L}$ requires a target logit of at least $\psi^{-1}(\mathcal{L})$.

\begin{proposition}[deep-layer concentration]\label{prop:deep-concentration}
Let $Q$ be any synthetic distribution with balanced labels, and set $\delta_\mu=\|\boldsymbol{\mu}_L(Q)-\boldsymbol{\mu}_L(P)\|_2$, $\delta_{\sigma}=\|\boldsymbol{\sigma}_L^2(Q)-\boldsymbol{\sigma}_L^2(P)\|_2$.
Then
\begin{equation}
\begin{split}
&\mathbb{E}_{Q}\big\|\tilde{\mathbf{z}}(\widehat X)-\alpha\mathring{\mathbf{w}}_Y\big\|_2^2\\
&\quad+\mathbb{E}_{Q}\Big[\tfrac{1}{N_L}\textstyle\sum_n\|\mathbf{a}_{L,n}(\widehat X)-\mathbf{z}(\widehat X)\|_2^2\Big]\le\Xi_L,
\end{split}
\end{equation}
where
\begin{equation}
\begin{split}
\Xi_L={}&2\alpha\big(\alpha\bar w^2-\psi^{-1}(\mathcal{L}_{\mathrm{CE}}(Q))\big)+2\alpha\bar w\sqrt{\varepsilon_{\mathrm{NC}}}\\
&+\varepsilon_{\mathrm{NC}}+\varepsilon_{\mathrm{sp}}+\sqrt{D_L}\,\delta_\sigma+\delta_\mu^2 ,
\end{split}
\end{equation}
and consequently $\frac1C\sum_{y}W_2^2\big(Q_{\mathbf{z}\mid Y=y},P_{\mathbf{z}\mid Y=y}\big)\le2\Xi_L+2\varepsilon_{\mathrm{NC}}$.
\end{proposition}

\textit{Proof.}
\textbf{Step 1: total variance.}
For any distribution $R$, since $\sum_n(\mathbf{a}_{L,n}-\mathbf{z})=\mathbf{0}$, the sum of the channel variances splits into the variance of the pooled feature and the spatial dispersion of the tokens:
\begin{equation}
\begin{split}
\textstyle\sum_{d}\sigma^2_{L,d}(R)={}&\mathbb{E}_R\|\mathbf{z}-\boldsymbol{\mu}_L(R)\|_2^2\\
&+\mathbb{E}_R\Big[\tfrac{1}{N_L}\textstyle\sum_n\|\mathbf{a}_{L,n}-\mathbf{z}\|_2^2\Big].
\end{split}
\end{equation}%
For $R=P$, the first term is close to the between-class variance $\mathbb{E}_P\|\alpha\mathring{\mathbf{w}}_Y\|_2^2=\alpha^2\bar w^2$, and Minkowski's inequality gives $\mathbb{E}_P\|\tilde{\mathbf{z}}\|_2^2\le(\alpha\bar w+\sqrt{\varepsilon_{\mathrm{NC}}})^2$, hence $\sum_d\sigma^2_{L,d}(P)\le(\alpha\bar w+\sqrt{\varepsilon_{\mathrm{NC}}})^2+\varepsilon_{\mathrm{sp}}$.
For $R=Q$, Cauchy--Schwarz gives $\sum_d\sigma^2_{L,d}(Q)\le\sum_d\sigma^2_{L,d}(P)+\sqrt{D_L}\,\delta_\sigma$, and $\mathbb{E}_Q[\mathbf{z}]=\boldsymbol{\mu}_L(Q)$ gives $\mathbb{E}_Q\|\tilde{\mathbf{z}}\|_2^2=\mathbb{E}_Q\|\mathbf{z}-\boldsymbol{\mu}_L(Q)\|_2^2+\delta_\mu^2$.
Therefore
\begin{equation}
\begin{split}
&\mathbb{E}_{Q}\|\tilde{\mathbf{z}}\|_2^2+\mathbb{E}_{Q}\Big[\tfrac{1}{N_L}\textstyle\sum_n\|\mathbf{a}_{L,n}-\mathbf{z}\|_2^2\Big]\\
&\le\alpha^2\bar w^2+2\alpha\bar w\sqrt{\varepsilon_{\mathrm{NC}}}+\varepsilon_{\mathrm{NC}}+\varepsilon_{\mathrm{sp}}\\
&\quad+\sqrt{D_L}\,\delta_\sigma+\delta_\mu^2.
\end{split}\label{eq:theory-variance-budget}
\end{equation}%
\textbf{Step 2: cross-entropy.}
The centered logits $\mathbf{u}=\Pi_C\mathcal{T}(x)=\mathring{\mathbf{W}}^\top\tilde{\mathbf{z}}+\mathbf{b}_{\mathrm{eff}}$ have zero sum, where $\mathbf{b}_{\mathrm{eff}}=\Pi_C(\mathbf{b}+\mathbf{W}^\top\boldsymbol{\mu}_L(P))$.
Because $\sum_{c\ne y}(u_c-u_y)=-Cu_y$, Jensen's inequality over the $C-1$ non-target logits gives, for every zero-sum $\mathbf{u}$,
\begin{equation}
\ell_{\mathrm{CE}}(\mathbf{u},y)\ge\psi(u_y),
\label{eq:theory-ce-psi}
\end{equation}%
with equality iff $u_c=-u_y/(C-1)$ for all $c\ne y$.
Taking expectation under $Q$ and using convexity and monotonicity of $\psi$, $\mathbb{E}_Q[u_Y]\ge\psi^{-1}(\mathcal{L}_{\mathrm{CE}}(Q))$.
The balanced prior gives $\mathbb{E}_Q[b_{\mathrm{eff},Y}]=\frac1C\sum_yb_{\mathrm{eff},y}=0$, so the bias drops out and
\begin{equation}
\mathbb{E}_Q\langle\mathring{\mathbf{w}}_Y,\tilde{\mathbf{z}}\rangle\ge\psi^{-1}(\mathcal{L}_{\mathrm{CE}}(Q)).\label{eq:theory-ce-margin}
\end{equation}%
\textbf{Step 3: completing the square.}
Since $\mathbb{E}_Q\|\mathring{\mathbf{w}}_Y\|_2^2=\bar w^2$, $\mathbb{E}_Q\|\tilde{\mathbf{z}}-\alpha\mathring{\mathbf{w}}_Y\|_2^2=\mathbb{E}_Q\|\tilde{\mathbf{z}}\|_2^2+\alpha^2\bar w^2-2\alpha\mathbb{E}_Q\langle\mathring{\mathbf{w}}_Y,\tilde{\mathbf{z}}\rangle$.
Adding the spatial dispersion term and applying Eqs.~\eqref{eq:theory-variance-budget} and~\eqref{eq:theory-ce-margin} gives the first claim.

\noindent\textbf{Step 4: $W_2$ bound.}
For class $y$ let $\boldsymbol{\nu}_y=\boldsymbol{\mu}_L(P)+\alpha\mathring{\mathbf{w}}_y$.
The triangle inequality for $W_2$ through $\delta_{\boldsymbol{\nu}_y}$ and $(a+b)^2\le2a^2+2b^2$ give $W_2^2(Q_{\mathbf{z}\mid y},P_{\mathbf{z}\mid y})\le2\mathbb{E}_{Q\mid y}\|\tilde{\mathbf{z}}-\alpha\mathring{\mathbf{w}}_y\|_2^2+2\mathbb{E}_{P\mid y}\|\tilde{\mathbf{z}}-\alpha\mathring{\mathbf{w}}_y\|_2^2$.
Averaging over $y$ yields the second claim.

\paragraph{Reading the bound.}
The left-hand side of Proposition~\ref{prop:deep-concentration} measures all $D_L$ coordinates of the synthetic feature, including the kernel of $\mathring{\mathbf{W}}^\top$, the directions orthogonal to every class direction $\mathring{\mathbf{w}}_y$.
This kernel has dimension at least $D_L-C+1$ and receives no gradient from $\mathcal{L}_{\mathrm{CE}}$, because the gradient of $\mathcal{L}_{\mathrm{CE}}$ with respect to $\mathbf{z}$ lies in the span of the $\mathring{\mathbf{w}}_y$.
Writing $\boldsymbol{\mu}_{Q,y}$ for the mean of $\mathbf{z}(\widehat X)$ given $Y=y$, the first term of the left-hand side splits into $\frac1C\sum_y\|\boldsymbol{\mu}_{Q,y}-\boldsymbol{\mu}_L(P)-\alpha\mathring{\mathbf{w}}_y\|_2^2$, the distance of the synthetic class means from the real prototypes, and $\mathbb{E}_Q\|\mathbf{z}(\widehat X)-\boldsymbol{\mu}_{Q,Y}\|_2^2$, the within-class variance of the synthetic features, so the bound controls both.
The statistic loss supplies a cap on the total variance $\sum_d\sigma^2_{L,d}$.
This quantity is a trace, so it equals the sum of the variances along any orthonormal basis, this kernel included, and it does not depend on cross-channel structure.
On real data it is close to $\alpha^2\bar w^2$, the between-class variance, because $\varepsilon_{\mathrm{NC}}$ and $\varepsilon_{\mathrm{sp}}$ are small (Step~1).
A loss $\mathcal{L}_{\mathrm{CE}}(Q)$ requires an average target score of at least $\psi^{-1}(\mathcal{L}_{\mathrm{CE}}(Q))$, and the real prototypes have the class-average score $\alpha\bar w^2$.
The first term of $\Xi_L$ compares the two, and it is at most zero once $\mathcal{L}_{\mathrm{CE}}(Q)\le\psi(\alpha\bar w^2)$.
The feature then carries the prototype component, which takes up most of this total variance (Step~3), and the variance left for the kernel and for spatial dispersion is bounded by $\Xi_L$.
The limiting case is exact.
Under exact Neural Collapse with a simplex head, that is, $\varepsilon_{\mathrm{NC}}=\varepsilon_{\mathrm{sp}}=0$, $\mathring{\mathbf{W}}^\top\mathring{\mathbf{W}}=\frac{C}{C-1}\bar w^2\,\Pi_C$ and $\mathbf{b}_{\mathrm{eff}}=\mathbf{0}$, the real data reach $\mathcal{L}_{\mathrm{CE}}(P)=\psi(\alpha\bar w^2)$.
Every synthetic distribution with $\delta_\mu=\delta_\sigma=0$ and $\mathcal{L}_{\mathrm{CE}}(Q)\le\mathcal{L}_{\mathrm{CE}}(P)$ then has $\Xi_L\le0$, so $\tilde{\mathbf{z}}(\widehat X)=\alpha\mathring{\mathbf{w}}_Y$ and $\mathbf{a}_{L,n}(\widehat X)=\mathbf{z}(\widehat X)$ almost surely in every coordinate: the synthetic features coincide with the real class prototypes.

\begin{proposition}[shallow-layer matching leaves co-occurrence undetermined]\label{prop:shallow-cooccurrence}
For every $C\ge2$, every even $N\ge2$ and every $\varepsilon>0$, there exist a two-stage spatial ReLU teacher $\mathcal{T}$ with exact Neural Collapse on $P$ ($\varepsilon_{\mathrm{NC}}=0$, accuracy 1) and a synthetic distribution $Q^{\mathrm{sh}}$ such that: (i) $\mathcal{L}_{\mathrm{stat}}^{(1)}(Q^{\mathrm{sh}},P)=0$; (ii) $\mathcal{L}_{\mathrm{CE}}(Q^{\mathrm{sh}})<\min(\varepsilon,\mathcal{L}_{\mathrm{CE}}(P))$; (iii) $\frac1C\sum_yW_2^2(Q^{\mathrm{sh}}_{\mathbf{z}_2\mid y},P_{\mathbf{z}_2\mid y})=\frac14$; (iv) for every temperature $T>0$, the unique minimum-Frobenius-norm minimizer of the synthetic KL loss $\mathbb{E}_{Q^{\mathrm{sh}}}[D_{\mathrm{KL}}(\mathrm{Softmax}(\mathcal{T}(\widehat X)/T)\,\|\,\mathrm{Softmax}(\mathcal{S}(\widehat X)/T))]$ over the linear students $\mathcal{S}=\mathbf{W}_{\mathrm{obj}}^\top\boldsymbol{\phi}^{\mathrm{obj}}+\mathbf{W}_{\mathrm{tex}}^\top\boldsymbol{\phi}^{\mathrm{tex}}$ on the two pooled branches of layer~2 (defined in the construction) outputs $\mathbf{0}$ on $P$, so its accuracy on $P$ is $1/C$; (v) on the same teacher, $\mathcal{L}_{\mathrm{stat}}^{(2)}(Q^{\mathrm{sh}},P)\ge\frac{1}{2\sqrt C}>0$.
\end{proposition}

\textit{Construction.}
Layer~1 is the identity on $\mathbf{A}_1(x)=x\in\mathbb{R}^{N\times4C}$, with four channels per class $c$: two object-part channels $u_{c,1},u_{c,2}$ and two texture channels $v_{c,1},v_{c,2}$.
Layer~2 computes, at every token $n$ and for every class $c$,
\begin{equation}
\begin{split}
a_{2,c}[n]={}&\underbrace{2N\,\mathrm{ReLU}\big(u_{c,1}[n]+u_{c,2}[n]-1.5\big)}_{\phi^{\mathrm{obj}}_c[n]}\\
&+\underbrace{3\,\mathrm{ReLU}\big(v_{c,1}[n]+v_{c,2}[n]-1\big)}_{\phi^{\mathrm{tex}}_c[n]},
\end{split}
\end{equation}%
followed by average pooling over tokens, $\mathbf{z}_2=\boldsymbol{\phi}^{\mathrm{obj}}+\boldsymbol{\phi}^{\mathrm{tex}}$ with $\boldsymbol{\phi}^{\mathrm{obj}}=\frac1N\sum_n\boldsymbol{\phi}^{\mathrm{obj}}[n]$ and $\boldsymbol{\phi}^{\mathrm{tex}}=\frac1N\sum_n\boldsymbol{\phi}^{\mathrm{tex}}[n]$, and the head $\mathbf{W}=\beta\Pi_C$, where $\beta>0$ satisfies $\log(1+(C-1)e^{-\frac32\beta})<\varepsilon$.
The object branch fires only when two part channels are active at the same token, and the texture branch fires only when its two channels have the same sign.

On $P$, draw $Y$, a token $U$ and token subsets $S_c$ with $|S_c|=N/2$ ($c=1,\dots,C$), independently and uniformly.
Both part channels of class $Y$ equal $1$ at $U$ and $0$ elsewhere, all other part channels are $0$, and every texture pair satisfies $v_{c,1}[n]=-v_{c,2}[n]=R_{c,n}$, where $R_{c,n}=+1$ on $S_c$ and $R_{c,n}=-1$ off $S_c$.
Then $\mathbf{z}_2(X)=\mathbf{e}_Y$, so $\tilde{\mathbf{z}}_2=\mathring{\mathbf{w}}_Y/\beta$ and $\varepsilon_{\mathrm{NC}}=0$.
On $Q^{\mathrm{sh}}$, draw $Y$, an ordered pair of distinct tokens $U_1\ne U_2$ and a token subset $S$ with $|S|=N/2$, independently and uniformly.
Part channel $u_{Y,1}$ is $1$ at $U_1$, part channel $u_{Y,2}$ is $1$ at $U_2$, and both are $0$ elsewhere.
The two texture channels of class $Y$ both equal $+1$ on $S$ and $-1$ off $S$.
All other channels follow $P$.

\textit{Verification.}
(i) Every part channel has a single $1$ at a uniformly random token when $Y=c$ under both distributions, and every texture channel has per-image mean $0$ and second moment $1$ under both distributions.
Hence $\boldsymbol{\mu}_1$ and $\boldsymbol{\sigma}_1^2$ coincide.
(ii) On $Q^{\mathrm{sh}}$, $u_{Y,1}[n]+u_{Y,2}[n]\le1<1.5$, so $\boldsymbol{\phi}^{\mathrm{obj}}\equiv\mathbf{0}$, while the texture branch equals $3$ on $S$, which gives $\mathbf{z}_2(\widehat X)=\frac32\mathbf{e}_Y$.
The logit margin is $\frac32\beta$ on $Q^{\mathrm{sh}}$ and $\beta$ on $P$, so $\mathcal{L}_{\mathrm{CE}}(Q^{\mathrm{sh}})=\log(1+(C-1)e^{-\frac32\beta})<\min(\varepsilon,\log(1+(C-1)e^{-\beta}))$.
(iii) Given $Y=y$ the two layer-2 laws are the point masses $\mathbf{e}_y$ and $\frac32\mathbf{e}_y$, so $W_2^2=(\frac32-1)^2=\frac14$.
(iv) On $Q^{\mathrm{sh}}$, $\boldsymbol{\phi}^{\mathrm{obj}}\equiv\mathbf{0}$ and $\boldsymbol{\phi}^{\mathrm{tex}}=\frac32\mathbf{e}_Y$.
The loss does not depend on $\mathbf{W}_{\mathrm{obj}}$, and it vanishes exactly for $\mathbf{W}_{\mathrm{tex}}=\beta\Pi_C+\mathbf{c}\mathbf{1}_C^\top$ with $\mathbf{c}\in\mathbb{R}^C$, which reproduce the teacher logits up to a common shift.
Since $\langle\Pi_C,\mathbf{c}\mathbf{1}_C^\top\rangle_F=0$, the minimum-norm minimizer is $(\mathbf{W}_{\mathrm{obj}},\mathbf{W}_{\mathrm{tex}})=(\mathbf{0},\beta\Pi_C)$ for every $T>0$.
On $P$ every texture pair has opposite signs, so $\boldsymbol{\phi}^{\mathrm{tex}}\equiv\mathbf{0}$ and the student outputs $\mathbf{0}$.
(v) At layer~2, $\mu_{2,c}(P)=\frac1C$ and $\mu_{2,c}(Q^{\mathrm{sh}})=\frac{3}{2C}$ for every $c$, so $\|\boldsymbol{\mu}_2(Q^{\mathrm{sh}})-\boldsymbol{\mu}_2(P)\|_2=\frac{1}{2\sqrt C}$. $\square$

\paragraph{Reading the construction.}
The two distributions share every layer-1 channel marginal and differ in which channels are active together.
On $P$ the two part channels of the class fire at the same token and every texture pair has opposite signs; on $Q^{\mathrm{sh}}$ they fire at different tokens and the texture pair of the class shares a sign.
The ReLU conjunctions of layer~2 read exactly this co-occurrence, so a layer-1 statistic cannot determine the pooled layer-2 feature.
On $Q^{\mathrm{sh}}$ the object branch stays silent and the texture branch carries the class, so a student distilled on $Q^{\mathrm{sh}}$ learns only the texture branch.
On $P$ the texture branch is silent, and this student outputs $\mathbf{0}$ (claim (iv)).
At layer~2 the conjunction outcome appears in the channel means and variances, and $\mathcal{L}_{\mathrm{stat}}^{(2)}$ separates the two distributions on the same teacher (claim (v)).

\paragraph{Relation to \texorpdfstring{Fig.~\ref{fig:cmi_bn_layerwise_ablation}}{Fig. 2}.}
The two real-data quantities of the Setup also explain the deep clusters in Fig.~\ref{fig:resnet_bn_statistics_tsne} and in Figs.~\ref{fig:appendix_vit_tsne} and~\ref{fig:appendix_cnx_tsne}, where each point joins the per-image channel mean $\mathbf{z}(x)$ and the per-image channel standard deviation $\mathbf{s}(x)$.
Since $\|\mathbf{s}(x)\|_2^2=\frac{1}{N_L}\sum_n\|\mathbf{a}_{L,n}(x)-\mathbf{z}(x)\|_2^2$, we have $\varepsilon_{\mathrm{sp}}=\mathbb{E}_P\|\mathbf{s}(X)\|_2^2$.
A small $\varepsilon_{\mathrm{NC}}$ places $\mathbf{z}(x)$ near its class prototype, and a small $\varepsilon_{\mathrm{sp}}$ keeps $\mathbf{s}(x)$ near $\mathbf{0}$, so the deepest-layer statistics form one cluster per class.
During synthesis, Proposition~\ref{prop:deep-concentration} shows that at the deepest layer, where the classifier reads $\mathbf{z}$ linearly, the channel statistics and the classification loss together determine the class-conditional representation up to $\Xi_L$.
Among all laws with given means and covariances, Gaussians are the closest in $W_2$~\cite{dowson1982frechet}.
Hence the squared Fr\'echet distance between the Gaussian fits of $Q_{\mathbf{z}}$ and $P_{\mathbf{z}}$, which T-FID reports on teacher features, is at most $W_2^2(Q_{\mathbf{z}},P_{\mathbf{z}})\le\frac1C\sum_yW_2^2(Q_{\mathbf{z}\mid Y=y},P_{\mathbf{z}\mid Y=y})$, so the same bound $2\Xi_L+2\varepsilon_{\mathrm{NC}}$ controls it.
Proposition~\ref{prop:shallow-cooccurrence} shows that a shallow layer followed by nonlinear conjunctions does not have this property.
Together, the two results match the near-zero gain of shallow single-block matching and the low T-FID of the deepest blocks, which recover most of the all-block accuracy (Figs.~\ref{fig:cmi_bn_layerwise_ablation_cnn}--\ref{fig:cmi_bn_layerwise_ablation_vit}, and Figs.~\ref{fig:appendix_layerwise_c10} and~\ref{fig:appendix_layerwise_in100} for further architectures and datasets).
For intermediate depths, the separation of real-data features across classes grows progressively with depth~\cite{he2023law,raghu2021vision}, which is consistent with the smooth rise of single-block accuracy between these two ends.

\section{Methodology Theoretical Foundations}
\label{app:instantiation}

\subsection{Cross-Architecture CDSA Instantiation}
\label{app:cdsa_instantiation}
The interface in Eq.~\eqref{eq:attribution-interface} requires only the spatial response $\mathbf{R}^{f}\in\mathbb{R}^{H_p\times W_p\times D_f}$; backbones differ solely in how this response is recovered, depending on whether an explicit spatial representation survives until classification.
We describe the two cases below and omit the superscript $f$ for brevity.

\paragraph{Feature-map-based architectures.}
For ResNet, ConvNeXt, and hierarchical transformers such as Swin, the response is the final classifier-compatible feature map taken before global pooling,
\begin{equation}
\mathbf{R}(\hat{x})
=
\Phi\!\left(\mathbf{F}_{\mathrm{final}}(\hat{x})\right)
\in\mathbb{R}^{H_p\times W_p\times D_f},
\label{eq:featuremap-response}
\end{equation}
$\Phi(\cdot)$ is the terminal normalization of the backbone.
ResNet has none, so $\Phi$ is the identity and Eq.~\eqref{eq:attribution-interface} is standard class-activation mapping.
Swin normalizes its final tokens before pooling, so $\Phi$ is this LayerNorm and the normalized tokens are restored to the spatial grid of the last stage.
ConvNeXt normalizes the pooled feature $\bar{\mathbf{f}}=\frac{1}{H_pW_p}\sum_{i,j}\mathbf{F}_{\mathrm{final}}[i,j]$, so $\Phi$ applies the same LayerNorm at every location with the standard deviation of $\bar{\mathbf{f}}$:
\begin{equation}
\Phi\!\left(\mathbf{F}_{\mathrm{final}}\right)[i,j]
=
\mathbf{g}\odot
\frac{\mathbf{F}_{\mathrm{final}}[i,j]-\mu\!\left(\mathbf{F}_{\mathrm{final}}[i,j]\right)}
{\sigma\!\left(\bar{\mathbf{f}}\right)},
\label{eq:convnext-response}
\end{equation}
where $\mathbf{g}$ is the LayerNorm gain, and $\mu(\cdot)$ and $\sigma(\cdot)$ are the mean and standard deviation over channels.
The spatial average of $\Phi(\mathbf{F}_{\mathrm{final}})$ equals the LayerNorm output of $\bar{\mathbf{f}}$ up to its bias, so Eq.~\eqref{eq:attribution-interface} is class-activation mapping for all three backbones.

\paragraph{CLS-token-based architectures.}
Plain ViTs retain no such feature map, and their raw attention indicates spatial saliency without category-specific evidence.
We therefore intercept the attention-weighted values before token aggregation.
Let index $0$ denote the CLS token and indices $1{:}N$ the $N=H_pW_p$ patch tokens.
In the final multi-head attention block, the CLS-to-patch attention of head $h\in\{1,\ldots,H\}$ is
\begin{equation}
\mathbf{a}^{(h)}
=
\operatorname{Softmax}\!\left(
\frac{\mathbf{q}^{(h)}_{0}\bigl(\mathbf{K}^{(h)}_{1:N}\bigr)^{\top}}{\sqrt{d_h}}
\right)\in\mathbb{R}^{N},
\label{eq:vit-attention}
\end{equation}
with $d_h$ the per-head dimension.
Rather than summing over patches, we keep the contribution of each patch $n$ across all heads and merge them through the output projection $\mathbf{W}^{O}$:
\begin{equation}
\mathbf{r}_n
=
\operatorname{Concat}\!\left(
a_n^{(1)}\mathbf{V}^{(1)}_{n},\ldots,a_n^{(H)}\mathbf{V}^{(H)}_{n}
\right)\mathbf{W}^{O}\in\mathbb{R}^{D_f},
\label{eq:patch-contribution}
\end{equation}
and rearranging $\{\mathbf{r}_n\}_{n=1}^{N}$ onto the patch grid gives $\mathbf{R}(\hat{x})\in\mathbb{R}^{H_p\times W_p\times D_f}$.

\subsection{Orthogonal ``What''-``Where'' Prior}
\label{app:theory_what_where}
BN-free teachers store no channel statistics (Sec.~\ref{app:theory}), so synthesis without further priors relies on the classification loss $\mathcal{L}_{\mathrm{CE}}$ alone.
Sec.~\ref{subsec:overview} of the main paper splits the semantic prior into \emph{what} the semantics span and \emph{where} they localize.
CSC (Sec.~\ref{subsec:csc}) conditions the generator on language-derived embeddings, and SSA (Sec.~\ref{subsec:ssa}) adds an attribution loss to $\mathcal{L}_{\mathrm{CE}}$ in the generator objective of Eq.~\eqref{eq:generator-objective}.
This subsection explains why $\mathcal{L}_{\mathrm{CE}}$ needs both priors.
The Setup shows that the split is built into CDSA (Eq.~\eqref{eq:attribution-interface}).
Each attribution map separates into its spatial mean, the ``What'' component, which is what the classifier reads after pooling, and the remaining spatial pattern, the ``Where'' component, which records where the evidence lies (Fig.~\ref{fig:theory_schematic_b}).
Proposition~\ref{prop:ce-degeneracy} shows that the classification loss falls short on both components: it is blind to where the evidence lies, and it pushes all images of a class toward the same logits.
Proposition~\ref{prop:ssa-where} shows that SSA (Eqs.~\eqref{eq:structural-prior}--\eqref{eq:spatial-alignment}) closes the first gap, as it gives the evidence of each class a compact, center-biased layout.
Proposition~\ref{prop:csc-what} shows how CSC (Eqs.~\eqref{eq:semantic-filtering}--\eqref{eq:persistent-categorical-modulation}) acts on the second gap: a single point of logits cannot come closer than a fixed distance to a class with several semantic modes, while $K$ semantic clusters let the class be matched one mode at a time.
Together, these results give the mechanism behind the gains of CSC and SSA in Tabs.~\ref{tab:ablation}--\ref{tab:csc} and the broader coverage of the synthetic distribution in Figs.~\ref{fig:tsne}--\ref{fig:dist-quality} of the main paper.
\paragraph{Setup.}
Let $N=H_pW_p\ge2$ be the number of patches and $C\ge2$ the number of classes.
Reshaping the $H_p\times W_p\times C$ maps of Eq.~\eqref{eq:attribution-interface} over the $N$ patches, equip the CDSA tensor space $\mathcal{M}=\mathbb{R}^{N\times C}$ with the normalized norm $\|\mathbf{M}\|_{\mathcal{M}}=\frac{1}{\sqrt{N}}\|\mathbf{M}\|_F$.
Each row $\mathbf{M}[n]\in\mathbb{R}^C$ is the class evidence that patch $n$ contributes to the prediction, and each column $\mathbf{M}_c\in\mathbb{R}^N$ is the spatial map of class $c$.
The spatial average $\mathcal{A}(\mathbf{M})=\frac{1}{N}\sum_n\mathbf{M}[n]\in\mathbb{R}^C$ is the ``What'' component of $\mathbf{M}$, and the residual $\mathcal{P}_{\mathrm{where}}\mathbf{M}$ with rows $\mathbf{M}[n]-\mathcal{A}(\mathbf{M})$ is its ``Where'' component, which lies in the space $\ker\mathcal{A}$ of spatially zero-mean tensors.
Like a mean and the deviations from it, the spatially constant tensor with rows $\mathcal{A}(\mathbf{M})$ and the residual $\mathcal{P}_{\mathrm{where}}\mathbf{M}$ are orthogonal in $\mathcal{M}$, and
\begin{equation}
\|\mathbf{M}\|_{\mathcal{M}}^2=\|\mathcal{A}(\mathbf{M})\|_2^2+\|\mathcal{P}_{\mathrm{where}}\mathbf{M}\|_{\mathcal{M}}^2.
\label{eq:theory-what-where}
\end{equation}%
From here on, $\mathbf{z}(x)\in\mathbb{R}^C$ denotes the logits, as $\mathbf{z}^{\mathcal{T}}$ in Fig.~\ref{fig:framework} of the main paper (Sec.~\ref{app:theory} used $\mathbf{z}$ for the pooled feature).
For ResNet, Swin and ConvNeXt they are the spatial average of CDSA up to a constant, $\mathbf{z}(x)=\mathcal{A}(\mathbf{M}(x))+\tilde{\mathbf{b}}$ (Sec.~\ref{app:cdsa_instantiation}).
In plain ViT, the attention of the CLS token normalizes over itself and the $N$ patches, so each head $h$ of the final block passes the patch tokens to the CLS token through the sum $\sum_na_n^{(h)}\mathbf{V}^{(h)}_n$ of its weighted values in Eq.~\eqref{eq:patch-contribution}, scaled by the total attention of the head on the patches.
With the attention weights fixed, any change of these weighted values with zero spatial mean therefore leaves the logits unchanged and changes $\mathbf{M}$ only in $\ker\mathcal{A}$.
By Eq.~\eqref{eq:theory-what-where}, synthetic attributions come close to real ones only when both their ``What'' and their ``Where'' components do.
For any teacher, Softmax reads the logits only through the centered logits $\mathbf{u}=\Pi_C\mathbf{z}$, with $\Pi_C=I_C-\frac1C\mathbf{1}_C\mathbf{1}_C^\top$ as in Sec.~\ref{app:theory}.
Propositions~\ref{prop:ce-degeneracy}(ii) and~\ref{prop:csc-what} work with the laws of $\mathbf{u}$, and a frozen teacher maps bounded images to bounded $\mathbf{u}$, which gives the radius $B_u$ of Proposition~\ref{prop:ce-degeneracy}(ii).

\begin{proposition}[double degeneracy of the classification loss]\label{prop:ce-degeneracy}
Fix a class $y$ and a radius $B_u>0$.
(i) If $\mathbf{z}(\mathbf{M})=\mathcal{A}(\mathbf{M})+\tilde{\mathbf{b}}$, then $\ell_{\mathrm{CE}}(\mathbf{z}(\mathbf{M}+\boldsymbol{\Delta}),y)=\ell_{\mathrm{CE}}(\mathbf{z}(\mathbf{M}),y)$ for every $\mathbf{M}\in\mathcal{M}$ and $\boldsymbol{\Delta}\in\ker\mathcal{A}$, and $\mathcal{P}_{\mathrm{where}}\nabla_{\mathbf{M}}\ell_{\mathrm{CE}}=\mathbf{0}$.
(ii) Among distributions of centered logits supported on $\mathcal{B}_{B_u}=\{\mathbf{u}\in\mathbf{1}_C^\perp:\|\mathbf{u}\|_2\le B_u\}$, the expected loss $\mathbb{E}[\ell_{\mathrm{CE}}(\mathbf{u},y)]$ has the unique minimizer $\delta_{\mathbf{u}^\star_y}$, where $\mathbf{u}^\star_y$ is the point of $\mathcal{B}_{B_u}$ with the largest target logit.
The centered logits of class $y$ therefore collapse to a single point, at which all $C-1$ non-target logits are equal.
\end{proposition}

\textit{Proof.}
(i) $\mathcal{A}(\boldsymbol{\Delta})=\mathbf{0}$ leaves $\mathbf{z}$ unchanged, and since $\mathbf{z}$ depends on $\mathbf{M}$ only through $\mathcal{A}(\mathbf{M})$, every patch row receives the same gradient, which has no ``Where'' component.
(ii) For $\mathbf{u}\in\mathcal{B}_{B_u}$, $\sum_{c\ne y}u_c=-u_y$ and Cauchy--Schwarz give $\sum_{c\ne y}u_c^2\ge u_y^2/(C-1)$, with equality iff the non-target logits are equal, hence $u_y\le B_u\sqrt{(C-1)/C}$ with equality iff $\mathbf{u}=\mathbf{u}^\star_y$, the point of norm $B_u$ with $u_y>0$ and equal non-target logits.
By Eq.~\eqref{eq:theory-ce-psi}, an equality at equal non-target logits, and the strict monotonicity of the function $\psi$ of Sec.~\ref{app:theory}, $\ell_{\mathrm{CE}}(\mathbf{u},y)\ge\psi(u_y)\ge\psi(B_u\sqrt{(C-1)/C})=\ell_{\mathrm{CE}}(\mathbf{u}^\star_y,y)$, and the second inequality is strict for $\mathbf{u}\ne\mathbf{u}^\star_y$. $\square$

SSA acts on the first degeneracy and CSC on the second.

\paragraph{SSA template.}
Let $A>0$ and $B\ge0$ in Eq.~\eqref{eq:structural-prior} of the main paper.

Below, $\mathcal{A}$ and $\mathcal{P}_{\mathrm{where}}$ act on single-channel maps $\mathbf{m}\in\mathbb{R}^N$, with the normalized norm $\|\mathbf{m}\|_{1/N}=\frac{1}{\sqrt N}\|\mathbf{m}\|_2$.

\begin{proposition}[SSA locks the ``Where'' component]\label{prop:ssa-where}
Let the patch grid have $\max(H_p,W_p)\ge3$, and let $\bar{\mathbf{M}}_c=\mathbb{E}_Q[\mathrm{ReLU}(\mathbf{M}_c(\widehat X))]\in\mathbb{R}^N_{\ge0}$, $\bar{\mathbf{M}}_c\ne\mathbf{0}$, be the batch-averaged positive attribution of class $c$ (Sec.~\ref{subsec:ssa} of the main paper), with $\widehat X\sim Q$ as in Sec.~\ref{app:theory}.
The SSA term $1-\cos(\bar{\mathbf{M}}_c,\boldsymbol{\pi})$ of Eq.~\eqref{eq:spatial-alignment} in the main paper vanishes iff $\bar{\mathbf{M}}_c=\lambda_c\boldsymbol{\pi}$ for some $\lambda_c>0$.
At this optimum the ``Where'' component $\mathcal{P}_{\mathrm{where}}\bar{\mathbf{M}}_c=\lambda_c\mathcal{P}_{\mathrm{where}}\boldsymbol{\pi}$ is nonzero.
A spatially flat attribution, $\mathcal{P}_{\mathrm{where}}\bar{\mathbf{M}}_c=\mathbf{0}$, incurs the strictly positive loss $1-\mathcal{A}(\boldsymbol{\pi})/\|\boldsymbol{\pi}\|_{1/N}$.
\end{proposition}

\textit{Proof.}
Cauchy--Schwarz gives $\cos(\bar{\mathbf{M}}_c,\boldsymbol{\pi})=1$ iff $\bar{\mathbf{M}}_c$ is a positive multiple of $\boldsymbol{\pi}$.
Then $\mathcal{P}_{\mathrm{where}}\bar{\mathbf{M}}_c=\lambda_c\mathcal{P}_{\mathrm{where}}\boldsymbol{\pi}\ne\mathbf{0}$ because $\boldsymbol{\pi}$ is non-constant: along a grid axis with at least three positions, the squared distance to the peak takes at least two distinct values, and $A>0$ turns them into distinct template values.
For $\bar{\mathbf{M}}_c=a\mathbf{1}_N$ with $a>0$, $\cos(\mathbf{1}_N,\boldsymbol{\pi})=\mathcal{A}(\boldsymbol{\pi})/\|\boldsymbol{\pi}\|_{1/N}$, which is below $1$ for the same reason. $\square$

\begin{proposition}[point-mass floor and per-mode matching under CSC]\label{prop:csc-what}
Fix a class $y$, and let the centered logits of its real images follow $P_y=\frac{1}{K}\sum_{k=1}^K P_{y,k}$ with $K\ge2$ modes whose means are pairwise at least $\Delta$ apart.
(i) Every point mass satisfies $W_2^2(\delta_{\mathbf{a}},P_y)\ge\frac{K-1}{2K}\Delta^2$.
(ii) If CSC draws the semantic embedding of class $y$ from $K$ clusters with equal probability (Eq.~\eqref{eq:semantic-filtering} in the main paper keeps $r$ embeddings in each cluster) and $Q_{y,k}$ is the law of the centered teacher logits of the images synthesized from cluster $k$, so that class $y$ follows $Q^{\mathrm{CSC}}_y=\frac1K\sum_kQ_{y,k}$, then
\begin{equation}
W_2^2\big(Q^{\mathrm{CSC}}_y,P_y\big)\le\frac{1}{K}\textstyle\sum_{k=1}^K W_2^2(Q_{y,k},P_{y,k}).
\end{equation}
\end{proposition}

\textit{Proof.}
(i) $W_2^2(\delta_{\mathbf{a}},P_y)=\mathbb{E}_{P_y}\|\mathbf{u}-\mathbf{a}\|_2^2\ge\mathrm{Tr}\,\mathrm{Cov}(P_y)$, which is at least the between-mode variance $\frac{1}{K}\sum_k\|\mathbf{m}_k-\bar{\mathbf{m}}\|_2^2=\frac{1}{2K^2}\sum_{k,k'}\|\mathbf{m}_k-\mathbf{m}_{k'}\|_2^2\ge\frac{K-1}{2K}\Delta^2$, where $\mathbf{m}_k$ is the mean of $P_{y,k}$ and $\bar{\mathbf{m}}$ their average.
(ii) The equal-weight mixture of optimal couplings between $Q_{y,k}$ and $P_{y,k}$ is a coupling between $Q^{\mathrm{CSC}}_y$ and $P_y$. $\square$

\paragraph{Reading Propositions~\ref{prop:ce-degeneracy}--\ref{prop:csc-what}.}
When synthesis concentrates the logits of class $y$ at one point, as the one-hot loss favors (Proposition~\ref{prop:ce-degeneracy}(ii)), the distance to the real class distribution is at least $\sqrt{(K-1)/(2K)}\,\Delta$ (Proposition~\ref{prop:csc-what}(i)).
With $K$ semantic clusters (Proposition~\ref{prop:csc-what}(ii)), the squared distance is at most the average of the per-mode errors $W_2^2(Q_{y,k},P_{y,k})$.
Each per-mode error splits into the squared distance between the means of $Q_{y,k}$ and $P_{y,k}$ and the squared distance between the two laws centered at their means~\cite{peyre2019computational}.
Once the images of each cluster center on their own mode, the gap between modes therefore drops out, and only the errors within the modes remain.
This is the step from the single-embedding ``Prototype'' row to the multi-embedding ``Diverse'' and ``Filtered'' rows in Tab.~\ref{tab:csc_input} of the main paper, and it matches the broader coverage of the real distribution in Figs.~\ref{fig:tsne} and~\ref{fig:PRDC} and the lower intra-class Vendi MAE in Fig.~\ref{fig:FID} of the main paper.

The between-mode gap and the per-mode errors correspond to the two parameters of the semantic bank in Fig.~\ref{fig:hyper-cluster} of the main paper, the number $K$ of clusters and the number $r$ of in-distribution embeddings that the filtering of Eq.~\eqref{eq:semantic-filtering} keeps in each cluster, respectively.

SSA gives the ``Where'' component a target with a nonzero spatial layout (Proposition~\ref{prop:ssa-where}) in the subspace on which $\mathcal{L}_{\mathrm{CE}}$ has no gradient (Proposition~\ref{prop:ce-degeneracy}(i)).
The constraint acts on the batch average $\bar{\mathbf{M}}_c$, and it also gives the individual synthetic images a nonzero ``Where'' component on average.
The ``Where'' energy $\|\mathcal{P}_{\mathrm{where}}\mathbf{m}\|_{1/N}^2=\frac{1}{2N^2}\sum_{n,n'}(m_n-m_{n'})^2$ is half the mean squared difference between patches, which the $1$-Lipschitz ReLU can only shrink.
Jensen's inequality for the convex function $\|\mathcal{P}_{\mathrm{where}}(\cdot)\|_{1/N}^2$ then gives $\mathbb{E}_Q\|\mathcal{P}_{\mathrm{where}}\mathbf{M}_c(\widehat X)\|_{1/N}^2\ge\mathbb{E}_Q\|\mathcal{P}_{\mathrm{where}}\mathrm{ReLU}(\mathbf{M}_c(\widehat X))\|_{1/N}^2\ge\|\mathcal{P}_{\mathrm{where}}\bar{\mathbf{M}}_c\|_{1/N}^2$.
The right-hand side is positive as soon as the SSA term of class $c$ is below the loss $1-\mathcal{A}(\boldsymbol{\pi})/\|\boldsymbol{\pi}\|_{1/N}$ of a flat map (Proposition~\ref{prop:ssa-where}).
Because SSA sets the layout of the ``Where'' component, the shape of the template matters.
The center-biased template, which follows the layout of natural objects (Sec.~\ref{subsec:ssa} of the main paper), gives the highest accuracy in Tab.~\ref{tab:ssa-priors}, and the inverted template gives the lowest.
The two priors therefore act on the two orthogonal components of Eq.~\eqref{eq:theory-what-where}, which matches the gain from combining them in Tab.~\ref{tab:ablation} of the main paper (90.51, against 89.77 for CSC and 87.51 for SSA alone).

\subsection{Synthetic-to-Real Generalization Theory under SSD}
\label{app:theory_ssd}
Sec.~\ref{subsec:SSD} of the main paper adds SSD (Eq.~\eqref{eq:student-objective}) to the KL loss in the distillation objective of Eq.~\eqref{eq:distillation-objective}, because the KL loss passes on what the teacher predicts but discards where the teacher finds its evidence.
This subsection shows why this spatial evidence matters for a student that learns from synthetic images and is tested on real ones.
The KL loss fixes the centered logits, and with them the spatial mean of each attribution map, while SSD fixes the direction of the map, and Lemma~\ref{lem:scale-direction} shows that the two together fix the whole map.
Theorem~\ref{thm:ssd-bound} turns this into a bound on the error on real data.
With KL alone, the student sees each synthetic image through its pooled feature, and the transfer to real data is governed by a factor $\kappa_{\mathrm{pool}}$.
SSD lets the student see every patch, and the factor becomes $\kappa_{\mathrm{patch}}\le\kappa_{\mathrm{pool}}$, because the patches of an image span directions that their average hides.
Theorem~\ref{thm:ssd-necessity} gives a concrete example (Fig.~\ref{fig:theory_schematic_c}).
The synthetic images match the real deep-layer statistics, yet their background always agrees with the class.
With KL alone, the student reproduces the predictions of the teacher on these images exactly, yet it cannot tell the background from the object and reaches chance accuracy $1/C$ on real images.
With SSD, it looks where the teacher looks and classifies every real image correctly.
These results give the mechanism behind the SSD gain in Tab.~\ref{tab:ablation} of the main paper.

\paragraph{Setup.}
Fix a snapshot of the surrogate pool $\widehat{\mathcal{D}}$ over which Eq.~\eqref{eq:student-objective} of the main paper averages, let $Q_X=\mathrm{Unif}(\widehat{\mathcal{D}})$, and let $P_X$ be the law of the real test images.
Subscripts $Q$ and $P$ on risks and losses refer to these two laws.
SSD compares the maps on a common grid of $N\ge2$ patches, with the inner product $\langle\mathbf{m},\mathbf{m}'\rangle_{1/N}=\frac1N\langle\mathbf{m},\mathbf{m}'\rangle$ and the norm $\|\mathbf{m}\|_{1/N}$ of Sec.~\ref{app:theory_what_where}, so that the spatial mean is $\bar m=\langle\mathbf{m},\mathbf{1}_N\rangle_{1/N}$ and $\|\mathbf{1}_N\|_{1/N}=1$.
For $T>0$ and $p_T(\mathbf{z})=\mathrm{Softmax}(\mathbf{z}/T)$, the $T^2$-scaled distillation loss~\cite{hinton2015distilling} is $\ell_{\mathrm{KL}}=T^2D_{\mathrm{KL}}(p_T(\mathbf{z}^{\mathcal{T}})\,\|\,p_T(\mathbf{z}^{\mathcal{S}}))$, with mean $\mathcal{R}_R^{\mathrm{KL}}=\mathbb{E}_R[\ell_{\mathrm{KL}}]$ under a law $R$.
With $\cos(\mathbf{u},\mathbf{v})=0$ if $\mathbf{u}=\mathbf{0}$ or $\mathbf{v}=\mathbf{0}$, the SSD loss of Eq.~\eqref{eq:student-objective} on the pool is $\mathcal{L}_{\mathrm{SSD},Q}=\mathbb{E}_{Q_X}[\frac1C\sum_c(1-\cos(\mathbf{M}_c^{\mathcal{T}},\mathbf{M}_c^{\mathcal{S}}))]$, and $\mathcal{R}_Q^{\mathrm{KL}}+\lambda_{\mathrm{SSD}}\mathcal{L}_{\mathrm{SSD},Q}$ is the objective of Eq.~\eqref{eq:distillation-objective}.

\paragraph{KL loss and centered logits.}
The Softmax ignores a common shift of the logits, so the KL loss sees only the centered logits $\Pi_C\mathbf{z}$, and at a high temperature it reduces to matching them~\cite{hinton2015distilling}.
As a function of $\mathbf{u}=\Pi_C\mathbf{z}^{\mathcal{S}}$, $\ell_{\mathrm{KL}}$ differs from $T^2\log\sum_ce^{u_c/T}$ by an affine term, vanishes with its gradient at $\Pi_C\mathbf{z}^{\mathcal{T}}$, and has the Hessian $\mathrm{diag}(\mathbf{p})-\mathbf{p}\mathbf{p}^\top$ with $\mathbf{p}=p_T(\mathbf{u})$~\cite{boyd2004convex}.
For $\mathbf{v}\in\mathbf{1}_C^\perp$, its quadratic form is at most $\frac12\|\mathbf{v}\|_2^2$~\cite{bohning1992multinomial} and at least $\min_cp_c\|\mathbf{v}\|_2^2$, since it is $\mathrm{Var}_{c\sim\mathbf{p}}(v_c)$ and $\sum_cv_c=0$.
If both centered logits have entries in $[-B_z,B_z]$, every class probability on the segment between them is at least $p_{\min}(B_z,T)=(1+(C-1)e^{2B_z/T})^{-1}$, and a second-order Taylor expansion gives
\begin{equation}
\tfrac{p_{\min}(B_z,T)}{2}\|\Pi_C(\mathbf{z}^{\mathcal{S}}-\mathbf{z}^{\mathcal{T}})\|_2^2\le\ell_{\mathrm{KL}}\le\tfrac{1}{4}\|\Pi_C(\mathbf{z}^{\mathcal{S}}-\mathbf{z}^{\mathcal{T}})\|_2^2.
\label{eq:kl-sandwich}
\end{equation}%
A high temperature keeps $p_{\min}$ of the order of $1/C$ ($T=20$ in all our experiments, Sec.~\ref{app:implementation}), so matching soft labels pins down every centered logit, the non-target ones included.

\paragraph{Direction and norm of a map.}
SSD compares the attribution maps of the student and the teacher through a cosine, so it fixes the direction of each map and leaves its norm free.
The spatial mean supplies the norm when the teacher map overlaps with $\mathbf{1}_N$.
The coherence $\eta(\mathbf{u})=|\cos(\mathbf{u},\mathbf{1}_N)|$, which equals $|\bar u|/\|\mathbf{u}\|_{1/N}$ for $\mathbf{u}\ne\mathbf{0}$, measures this overlap, and by Eq.~\eqref{eq:theory-what-where} its square is the share of the ``What'' component in the squared norm of the map.
The maps in Eq.~\eqref{eq:student-objective} are signed, without the ReLU of SSA, so the non-target classes also have nonzero spatial means in general.

\begin{lemma}[spatial mean and direction determine a map]\label{lem:scale-direction}
Let $\mathbf{u},\mathbf{v}\in\mathbb{R}^N$ with $\eta(\mathbf{u})\ge\eta_0>0$ and $\|\mathbf{v}\|_{1/N}\le B_{\mathcal{S}}$.
Then
\begin{equation*}
\|\mathbf{v}-\mathbf{u}\|_{1/N}^2\le\tfrac{2}{\eta_0^2}(\bar v-\bar u)^2+2\bigl(1+\tfrac{2}{\eta_0^2}\bigr)B_{\mathcal{S}}^2\bigl(1-\cos(\mathbf{u},\mathbf{v})\bigr).
\end{equation*}
\end{lemma}

\textit{Proof.}
Write $\mathbf{v}=\lambda\hat{\mathbf{u}}+\mathbf{w}$ with $\hat{\mathbf{u}}=\mathbf{u}/\|\mathbf{u}\|_{1/N}$ and $\mathbf{w}\perp\hat{\mathbf{u}}$.
Since $\bar v-\bar u=(\lambda-\|\mathbf{u}\|_{1/N})\langle\hat{\mathbf{u}},\mathbf{1}_N\rangle_{1/N}+\langle\mathbf{w},\mathbf{1}_N\rangle_{1/N}$ and $|\langle\hat{\mathbf{u}},\mathbf{1}_N\rangle_{1/N}|\ge\eta_0$, we get $|\lambda-\|\mathbf{u}\|_{1/N}|\le(|\bar v-\bar u|+\|\mathbf{w}\|_{1/N})/\eta_0$.
The bound follows from $\|\mathbf{v}-\mathbf{u}\|_{1/N}^2=(\lambda-\|\mathbf{u}\|_{1/N})^2+\|\mathbf{w}\|_{1/N}^2$ and $\|\mathbf{w}\|_{1/N}^2=\|\mathbf{v}\|_{1/N}^2(1-\cos^2(\mathbf{u},\mathbf{v}))\le2B_{\mathcal{S}}^2(1-\cos(\mathbf{u},\mathbf{v}))$. $\square$

\paragraph{From logits to spatial means.}
For a scale $s>0$, the readout gap $\boldsymbol{\rho}_{\mathcal{S},s}=(\mathcal{A}(\mathbf{M}^{\mathcal{S}})-s\mathcal{A}(\mathbf{M}^{\mathcal{T}}))-\Pi_C(\mathbf{z}^{\mathcal{S}}-\mathbf{z}^{\mathcal{T}})$ measures how far the difference of the spatial means departs from the difference of the centered logits.
For linear readouts $\mathbf{z}=\mathcal{A}(\mathbf{M})+\tilde{\mathbf{b}}$, as in ResNet, Swin and ConvNeXt (Sec.~\ref{app:cdsa_instantiation}), with zero-row-sum heads and equal centered biases, $s=1$ gives $\boldsymbol{\rho}_{\mathcal{S},1}\equiv\mathbf{0}$.
In a CLS-readout ViT teacher, $\mathcal{A}(\mathbf{M}^{\mathcal{T}})$ averages the attention-weighted patch values projected directly through the head, whereas the logits pass their sum and the CLS stream through the final MLP, LayerNorm and head, so the two agree up to a scale $s$ and a remainder, which $\boldsymbol{\rho}_{\mathcal{S},s}$ collects.
The scale $s$ enters only the analysis, since the cosine in Eq.~\eqref{eq:student-objective} is invariant to rescaling the teacher map.

\paragraph{Students and shift factors.}
The student reads a patch representation $\boldsymbol{\psi}(x)\in\mathbb{R}^{N\times D}$ with bounded rows $\boldsymbol{\psi}_n(x)$ and pooled feature $\bar{\boldsymbol{\psi}}=\frac1N\sum_n\boldsymbol{\psi}_n$.
The linear-readout student $\mathcal{S}_{\mathbf{W}}$ with head $\mathbf{W}\in\mathbb{R}^{D\times C}$ has maps $\boldsymbol{\psi}\mathbf{W}$ and logits $\mathbf{W}^\top\bar{\boldsymbol{\psi}}+\tilde{\mathbf{b}}^{\mathcal{S}}$ for a fixed bias $\tilde{\mathbf{b}}^{\mathcal{S}}$, and we write $\boldsymbol{\rho}_{\mathbf{W},s}$ for its readout gap.
The teacher $\mathcal{T}$ is any network with $\|\mathbf{M}_c^{\mathcal{T}}\|_{1/N}\le B_{\mathcal{T}}$, we consider the students with $\|\mathbf{M}_c^{\mathcal{S}_{\mathbf{W}}}\|_{1/N}\le B_{\mathcal{S}}$, and the centered logits of $\mathcal{T}$ and of these students have entries in $[-B_z,B_z]$.
The second-moment matrices $\boldsymbol{\Sigma}_{\mathrm{pool}}(R)=\mathbb{E}_R[\bar{\boldsymbol{\psi}}\bar{\boldsymbol{\psi}}^\top]$ and $\boldsymbol{\Sigma}_{\mathrm{patch}}(R)=\mathbb{E}_R[\frac1N\sum_n\boldsymbol{\psi}_n\boldsymbol{\psi}_n^\top]$ record which feature directions the pooled features and the individual patches of a law $R$ reach.
Their difference $\boldsymbol{\Sigma}_{\mathrm{patch}}(R)-\boldsymbol{\Sigma}_{\mathrm{pool}}(R)=\mathbb{E}_R[\frac1N\sum_n(\boldsymbol{\psi}_n-\bar{\boldsymbol{\psi}})(\boldsymbol{\psi}_n-\bar{\boldsymbol{\psi}})^\top]\succeq\mathbf{0}$ is the within-image covariance of the patches, as in Step~1 of the proof of Proposition~\ref{prop:deep-concentration}.
They also measure the gaps between two students: $\mathbb{E}_R\|\mathbf{z}^{\mathcal{S}_{\mathbf{W}}}-\mathbf{z}^{\mathcal{S}_{\mathbf{W}'}}\|_2^2=\|\boldsymbol{\Sigma}_{\mathrm{pool}}(R)^{1/2}(\mathbf{W}-\mathbf{W}')\|_F^2$ for the logits and $\mathbb{E}_R\|\mathbf{M}^{\mathcal{S}_{\mathbf{W}}}-\mathbf{M}^{\mathcal{S}_{\mathbf{W}'}}\|_{\mathcal{M}}^2=\|\boldsymbol{\Sigma}_{\mathrm{patch}}(R)^{1/2}(\mathbf{W}-\mathbf{W}')\|_F^2$ for the maps.
Along a contrast direction, such as foreground minus background, patches with opposite projections cancel in $\bar{\boldsymbol{\psi}}$ but keep their energy in $\boldsymbol{\Sigma}_{\mathrm{patch}}$.
The shift factors compare the pooled real features, which determine the real-data logits, with the pooled and the patch synthetic features:
\begin{equation}
\begin{gathered}
\kappa_{\mathrm{pool}}=\sup_{\mathbf{u}\ne\mathbf{0}}\frac{\mathbf{u}^\top\boldsymbol{\Sigma}_{\mathrm{pool}}(P_X)\mathbf{u}}{\mathbf{u}^\top\boldsymbol{\Sigma}_{\mathrm{pool}}(Q_X)\mathbf{u}},\\
\kappa_{\mathrm{patch}}=\sup_{\mathbf{u}\ne\mathbf{0}}\frac{\mathbf{u}^\top\boldsymbol{\Sigma}_{\mathrm{pool}}(P_X)\mathbf{u}}{\mathbf{u}^\top\boldsymbol{\Sigma}_{\mathrm{patch}}(Q_X)\mathbf{u}},
\end{gathered}
\end{equation}%
where a ratio $a/0$ counts as $+\infty$ for $a>0$ and as $0$ for $a=0$.
Since $\boldsymbol{\Sigma}_{\mathrm{patch}}(Q_X)\succeq\boldsymbol{\Sigma}_{\mathrm{pool}}(Q_X)$, the patch factor never exceeds the pooled one, $\kappa_{\mathrm{patch}}\le\kappa_{\mathrm{pool}}$.

\begin{theorem}[synthetic-to-real risk bound for linear-readout students]\label{thm:ssd-bound}
Let $\kappa_{\mathrm{patch}}<\infty$, and fix a reference head $\mathbf{W}^\star\in\mathbb{R}^{D\times C}$, a scale $s>0$ and a level $\eta_0\in(0,1]$.
Let $\varepsilon_{Q,s}^2=\mathbb{E}_{Q_X}\|\mathbf{M}^{\mathcal{S}_{\mathbf{W}^\star}}-s\mathbf{M}^{\mathcal{T}}\|_{\mathcal{M}}^2$ and $\varepsilon_P^2=\mathbb{E}_{P_X}\|\Pi_C(\mathbf{z}^{\mathcal{S}_{\mathbf{W}^\star}}-\mathbf{z}^{\mathcal{T}})\|_2^2$ be the approximation errors of the reference student, and let $q_{\mathrm{incoh}}(\eta_0)=\frac{1}{C}\sum_c\mathbb{P}_{Q_X}(\eta(\mathbf{M}_c^{\mathcal{T}})<\eta_0)$ be the fraction of synthetic teacher maps with coherence below $\eta_0$.
Then every student $\mathcal{S}_{\mathbf{W}}$ as above satisfies
\begin{equation}
\mathcal{R}_P^{\mathrm{KL}}(\mathcal{S}_{\mathbf{W}})\le\kappa_{\mathrm{patch}}\bigl(\mathcal{B}_{\mathrm{KD}}(\mathcal{S}_{\mathbf{W}};s)+\varepsilon_{Q,s}^2\bigr)+\tfrac{1}{2}\varepsilon_P^2,
\end{equation}
where
\begin{equation}
\begin{split}
\mathcal{B}_{\mathrm{KD}}(\mathcal{S}_{\mathbf{W}};s)={}&\tfrac{8}{\eta_0^2p_{\min}(B_z,T)}\mathcal{R}_Q^{\mathrm{KL}}(\mathcal{S}_{\mathbf{W}})\\
&+2C\bigl(1+\tfrac{2}{\eta_0^2}\bigr)B_{\mathcal{S}}^2\mathcal{L}_{\mathrm{SSD},Q}(\mathcal{S}_{\mathbf{W}})\\
&+\tfrac{4}{\eta_0^2}\mathbb{E}_{Q_X}\|\boldsymbol{\rho}_{\mathbf{W},s}\|_2^2\\
&+C(B_{\mathcal{S}}+sB_{\mathcal{T}})^2q_{\mathrm{incoh}}(\eta_0).
\end{split}
\end{equation}
\end{theorem}

\textit{Proof.}
By Eq.~\eqref{eq:kl-sandwich} and $(a+b)^2\le2a^2+2b^2$, $\|\mathcal{A}(\mathbf{M}^{\mathcal{S}_{\mathbf{W}}})-s\mathcal{A}(\mathbf{M}^{\mathcal{T}})\|_2^2\le\frac{4}{p_{\min}(B_z,T)}\ell_{\mathrm{KL}}+2\|\boldsymbol{\rho}_{\mathbf{W},s}\|_2^2$.
Lemma~\ref{lem:scale-direction} with $\mathbf{u}=s\mathbf{M}_c^{\mathcal{T}}$ for the classes with $\eta(\mathbf{M}_c^{\mathcal{T}})\ge\eta_0$, and the bound $(B_{\mathcal{S}}+sB_{\mathcal{T}})^2$ for the other classes, summed over $c$ and averaged over $Q_X$, give $\mathbb{E}_{Q_X}\|\mathbf{M}^{\mathcal{S}_{\mathbf{W}}}-s\mathbf{M}^{\mathcal{T}}\|_{\mathcal{M}}^2\le\mathcal{B}_{\mathrm{KD}}(\mathcal{S}_{\mathbf{W}};s)$.
A split through $s\mathbf{M}^{\mathcal{T}}$ gives $\|\boldsymbol{\Sigma}_{\mathrm{patch}}(Q_X)^{1/2}(\mathbf{W}-\mathbf{W}^\star)\|_F^2\le2\mathcal{B}_{\mathrm{KD}}(\mathcal{S}_{\mathbf{W}};s)+2\varepsilon_{Q,s}^2$.
On $P_X$, Eq.~\eqref{eq:kl-sandwich} and a split through $\mathbf{z}^{\mathcal{S}_{\mathbf{W}^\star}}$ with $\|\Pi_C\|\le1$ give $\mathcal{R}_P^{\mathrm{KL}}(\mathcal{S}_{\mathbf{W}})\le\frac12\|\boldsymbol{\Sigma}_{\mathrm{pool}}(P_X)^{1/2}(\mathbf{W}-\mathbf{W}^\star)\|_F^2+\frac12\varepsilon_P^2$, and the definition of $\kappa_{\mathrm{patch}}$ completes the proof. $\square$

\paragraph{Reading Theorem~\ref{thm:ssd-bound}.}
The theorem carries what the student achieves on the synthetic pool to real data in three steps.
(1)~Eq.~\eqref{eq:kl-sandwich} and Lemma~\ref{lem:scale-direction} turn the KL and SSD losses into an error of the student maps at every synthetic patch.
(2)~The reference student $\mathcal{S}_{\mathbf{W}^\star}$ turns this error into the distance $\|\boldsymbol{\Sigma}_{\mathrm{patch}}(Q_X)^{1/2}(\mathbf{W}-\mathbf{W}^\star)\|_F$.
(3)~The factor $\kappa_{\mathrm{patch}}$ carries this distance to the real-data risk.
The reference $\mathbf{W}^\star$, the scale $s$ and the level $\eta_0$ are analysis parameters, and every choice gives a valid bound.
With the KL loss alone, the student is fixed only through the pooled synthetic features, and $\kappa_{\mathrm{pool}}$ takes the place of $\kappa_{\mathrm{patch}}$.
A direction that the pooled synthetic features miss makes $\kappa_{\mathrm{pool}}$ infinite, while the patches can still reach it and keep $\kappa_{\mathrm{patch}}$ finite, as in Theorem~\ref{thm:ssd-necessity}.
Synthetic images whose patches all coincide give $\kappa_{\mathrm{patch}}=\kappa_{\mathrm{pool}}$, so the gain of SSD comes from the variation across the patches of each synthetic image.

\begin{theorem}[constructive separation: strict necessity of SSD]\label{thm:ssd-necessity}
Fix $C\ge 2$, $N\ge 2$, $T>0$ and a background leverage $\alpha_{\mathrm{bg}}>1$.
There exist a patch representation $\boldsymbol{\psi}=(\mathbf{f},\mathbf{g}):\mathcal{X}\to\mathbb{R}^{N\times 2C}$, with a foreground block $\mathbf{f}$ and a background block $\mathbf{g}$ of $C$ channels each, and real and synthetic laws $P$ and $Q$ with the following properties.
The teacher $\mathcal{T}$ with head $\mathbf{W}^{\mathcal{T}}=[\Pi_C;\mathbf{0}]$, logits $(\mathbf{W}^{\mathcal{T}})^\top\bar{\boldsymbol{\psi}}$ and maps $\boldsymbol{\psi}\mathbf{W}^{\mathcal{T}}$ has accuracy $1$ on $P$.
The synthetic law $Q$ has the same deepest-layer channel moments $(\boldsymbol{\mu},\boldsymbol{\sigma}^2)$ of $\boldsymbol{\psi}$ as $P$, so $\mathcal{L}_{\mathrm{stat}}^{(L)}(Q,P)=0$ in the notation of Sec.~\ref{app:theory}, and every teacher map on $Q$ has coherence $\frac{1}{\sqrt{N}}$.
For the bias-free linear-readout students $\mathcal{S}_{\mathbf{W}}$ with unconstrained heads $\mathbf{W}=[\mathbf{W}_{\mathrm{fg}};\mathbf{W}_{\mathrm{bg}}]$, $\mathbf{W}_{\mathrm{fg}},\mathbf{W}_{\mathrm{bg}}\in\mathbb{R}^{C\times C}$:
(1)~$\kappa_{\mathrm{pool}}=+\infty$, whereas $\kappa_{\mathrm{patch}}=1$.
(2)~Under KL-only distillation ($\lambda_{\mathrm{SSD}}=0$), the minimum-Frobenius-norm global minimizer $\mathbf{W}_{\mathrm{KL}}^\star$, which is the limit of gradient descent from $\mathbf{0}$ with a small enough step size, has $\mathcal{R}_Q^{\mathrm{KL}}=0$, yet it has accuracy $\frac{1}{C}$ on $P$ and $\mathcal{R}_P^{\mathrm{KL}}>0$.
(3)~For every $\lambda_{\mathrm{SSD}}>0$, every global minimizer $\mathbf{W}_{\mathrm{SSD}}^\star$ of $\mathcal{R}_Q^{\mathrm{KL}}+\lambda_{\mathrm{SSD}}\mathcal{L}_{\mathrm{SSD},Q}$ has $\mathbf{W}_{\mathrm{bg}}=\mathbf{0}$, $\mathcal{R}_P^{\mathrm{KL}}=0$ and accuracy $1$ on $P$.
\end{theorem}

\textit{Construction.}
Draw $Y\sim\mathrm{Unif}(\{1,\dots,C\})$ and an independent object patch $U\sim\mathrm{Unif}(\{1,\dots,N\})$, and set $\mathbf{f}_U=N\mathbf{e}_Y$, $\mathbf{g}_U=\mathbf{0}$ and $\mathbf{f}_n=\mathbf{0}$ for $n\ne U$.
For $n\ne U$, set $\mathbf{g}_n=\frac{\alpha_{\mathrm{bg}}N}{N-1}\mathbf{e}_Y$ on $Q$, and $\mathbf{g}_n=\frac{\alpha_{\mathrm{bg}}N}{N-1}\mathbf{e}_{B}$ on $P$ with an independent background class $B\sim\mathrm{Unif}(\{1,\dots,C\})$.
Every channel marginal coincides between $Q$ and $P$.
The pooled foreground is $\bar{\mathbf{f}}=\mathbf{e}_Y$ on both laws, while the pooled background $\bar{\mathbf{g}}$ is $\alpha_{\mathrm{bg}}\mathbf{e}_Y$ on $Q$ and $\alpha_{\mathrm{bg}}\mathbf{e}_B$ on $P$.
The teacher logits $\Pi_C\mathbf{e}_Y$ peak at $Y$, and the teacher map $\mathbf{M}_c^{\mathcal{T}}=N(\mathbf{1}\{c=Y\}-\frac1C)\mathbf{e}_U$ is a nonzero multiple of $\mathbf{e}_U$, whose coherence is $\frac{1}{\sqrt N}$.

\textit{Proof.}
(1)~The direction $\mathbf{u}=(\alpha_{\mathrm{bg}}\mathbf{v},-\mathbf{v})$ with $\mathbf{v}\in\mathbf{1}_C^\perp\setminus\{\mathbf{0}\}$ gives $\mathbf{u}^\top\bar{\boldsymbol{\psi}}=0$ on $Q$ but $\mathbb{E}_{P}[(\mathbf{u}^\top\bar{\boldsymbol{\psi}})^2]>0$, so $\kappa_{\mathrm{pool}}=+\infty$.
For $\mathbf{u}=(\mathbf{a},\mathbf{b})$ and independent uniform $Y,B$, the construction gives $\mathbf{u}^\top\boldsymbol{\Sigma}_{\mathrm{patch}}(Q_X)\mathbf{u}=\mathbb{E}[Na_Y^2+\frac{N}{N-1}\alpha_{\mathrm{bg}}^2b_Y^2]=\mathbb{E}[Na_Y^2+\frac{N}{N-1}\alpha_{\mathrm{bg}}^2b_B^2]$ and $\mathbf{u}^\top\boldsymbol{\Sigma}_{\mathrm{pool}}(P_X)\mathbf{u}=\mathbb{E}[(a_Y+\alpha_{\mathrm{bg}}b_B)^2]$.
By Cauchy--Schwarz with the weights $\frac1N+\frac{N-1}{N}=1$, $(p+q)^2\le Np^2+\frac{N}{N-1}q^2$, with equality for every $(Y,B)$ at $\mathbf{a}=\mathbf{1}_C$ and $\alpha_{\mathrm{bg}}\mathbf{b}=(N-1)\mathbf{1}_C$, so $\kappa_{\mathrm{patch}}=1$.
(2)~On $Q$, the student logits are $(\mathbf{W}_{\mathrm{fg}}+\alpha_{\mathrm{bg}}\mathbf{W}_{\mathrm{bg}})^\top\mathbf{e}_Y$, so $\mathcal{R}_Q^{\mathrm{KL}}$ depends on $\mathbf{W}$ only through $(\mathbf{W}_{\mathrm{fg}}+\alpha_{\mathrm{bg}}\mathbf{W}_{\mathrm{bg}})\Pi_C$ and vanishes iff this matrix equals $\Pi_C$.
Gradient descent from $\mathbf{0}$ therefore stays in $\{\mathbf{W}_{\mathrm{bg}}=\alpha_{\mathrm{bg}}\mathbf{W}_{\mathrm{fg}},\ \mathbf{W}_{\mathrm{fg}}\mathbf{1}_C=\mathbf{0}\}$, where $\mathcal{R}_Q^{\mathrm{KL}}$ is smooth, strictly convex and coercive, and it converges to the minimum-norm minimizer $\mathbf{W}_{\mathrm{KL}}^\star=\frac{1}{1+\alpha_{\mathrm{bg}}^2}[\Pi_C;\alpha_{\mathrm{bg}}\Pi_C]$.
On $P$, its logits $\frac{1}{1+\alpha_{\mathrm{bg}}^2}\Pi_C(\mathbf{e}_Y+\alpha_{\mathrm{bg}}^2\mathbf{e}_B)$ peak at $B$, so it is correct only when $B=Y$, which gives accuracy $\frac1C$ and $\mathcal{R}_P^{\mathrm{KL}}>0$.
(3)~The teacher head attains $\mathcal{R}_Q^{\mathrm{KL}}=\mathcal{L}_{\mathrm{SSD},Q}=0$, so the global minimizers are the heads at which both losses vanish.
On $Q$, $\mathbf{M}_c^{\mathcal{S}_{\mathbf{W}}}=N(\mathbf{W}_{\mathrm{fg}})_{Y,c}\mathbf{e}_U+\frac{\alpha_{\mathrm{bg}}N}{N-1}(\mathbf{W}_{\mathrm{bg}})_{Y,c}\sum_{n\ne U}\mathbf{e}_n$, and $\mathbf{M}_c^{\mathcal{T}}$ is a nonzero multiple of $\mathbf{e}_U\perp\sum_{n\ne U}\mathbf{e}_n$, so $\mathcal{L}_{\mathrm{SSD},Q}=0$ forces $\mathbf{W}_{\mathrm{bg}}=\mathbf{0}$.
Then $\mathcal{R}_Q^{\mathrm{KL}}=0$ gives $\mathbf{W}_{\mathrm{fg}}=\Pi_C+\boldsymbol{\tau}\mathbf{1}_C^\top$ for some $\boldsymbol{\tau}\in\mathbb{R}^C$, whose logits on $P$ are $\mathbf{z}^{\mathcal{T}}+\tau_Y\mathbf{1}_C$, hence $\mathcal{R}_P^{\mathrm{KL}}=0$ and accuracy $1$. $\square$

\paragraph{Reading Theorem~\ref{thm:ssd-necessity}.}
On $Q$, the background of every image shows the class of its object, so the pooled foreground and background are collinear ($\kappa_{\mathrm{pool}}=+\infty$).
Gradient descent on the KL loss then moves the two weight blocks together and relies mostly on the stronger background cue, which on real images is independent of the object.
This pool is consistent with Proposition~\ref{prop:deep-concentration}: the real backgrounds vary independently of the object along the kernel of the teacher head, so $\varepsilon_{\mathrm{NC}}$ is large.
Within each image, the object and the background occupy different patches ($\kappa_{\mathrm{patch}}=1$), so SSD, which matches the maps patch by patch, sets $\mathbf{W}_{\mathrm{bg}}=\mathbf{0}$.
In Theorem~\ref{thm:ssd-bound} with $\eta_0=1/\sqrt N$, $\mathbf{W}^\star=\mathbf{W}^{\mathcal{T}}$ and $s=1$, the incoherent fraction, both approximation errors, and the KL loss and readout gap of $\mathbf{W}_{\mathrm{KL}}^\star$ all vanish, which leaves only its SSD term, positive because its background block is nonzero.
This is the mechanism of Sec.~\ref{subsec:SSD} of the main paper, and adding SSD to CSC and SSA in Tab.~\ref{tab:ablation} raises the accuracy from 90.51 to 93.91.

\section{Extended Implementation Details}
\label{app:implementation}
\subsection{Detailed Experimental Settings}
\label{app:exp_settings}
\paragraph{Datasets and Architectures.}

We evaluate UniDFKD on three widely adopted benchmarks: CIFAR-10, CIFAR-100~\cite{krizhevsky2009learning}, and ImageNet-100.
Specifically, CIFAR-10 and CIFAR-100 consist of 50,000 training and 10,000 validation images, categorized into 10 and 100 classes, respectively.
To ensure a fair comparison with prior data-free knowledge distillation methods, we maintain the native $32 \times 32$ spatial resolution exclusively for the classical ResNet-34 to ResNet-18 (RN34$\to$RN18) setting.
For all other distillation scenarios, the input images are upsampled to $224 \times 224$ via bilinear interpolation to strictly align with the default input configurations of modern architectures.
ImageNet-100 is a 100-class subset of the large-scale ILSVRC-2012 dataset~\cite{deng2009imagenet}.
We adopt the rigorously established CMC split~\cite{tian2020contrastive}, which comprises 126,689 training images and 5,000 validation images, providing a more challenging, high-resolution testbed.

To comprehensively evaluate the architectural generalizability of our framework, our experiments span four representative architecture families that encompass distinct structural paradigms (convolution- vs. attention-based) and normalization types (BN vs. LN): (1) the classic Batch Normalization (BN)-based \textbf{ResNet} (RN)~\cite{he2016deep}, serving as the foundational baseline for traditional CNNs; (2) the Layer Normalization (LN)-based \textbf{ConvNeXt} (CNX)~\cite{liu2022convnet}, representing modernized CNNs that effectively bridge the performance gap with Transformers; (3) the quintessential \textbf{Vision Transformer} (ViT)~\cite{dosovitskiy2021image}, which pioneers the application of patch-based global self-attention in vision tasks; and (4) the \textbf{Swin Transformer} (Swin)~\cite{liu2021swin}, a state-of-the-art hierarchical Transformer distinguished by its efficient shifted window-based local attention.
This diverse selection rigorously validates the robustness of UniDFKD across highly heterogeneous knowledge transfer scenarios.

Table~\ref{tab:arch_configs} lists the configuration of every LN-based teacher and student.
All models take $224\times224$ inputs.
ViT-S and ViT-T are the DeiT-Small and DeiT-Tiny architectures with $16\times16$ patches, 12 blocks, and 197 tokens per image.
Swin-T is the standard Swin-Tiny with $7\times7$ windows, and Swin-N keeps its stage layout and head counts with half the embedding width.
CNX-T is ConvNeXt-Tiny and CNX-N is ConvNeXt-Nano, which keeps the four-stage design with narrower widths and one fewer block per stage
Each student therefore retains the architecture family of its teacher and holds between one quarter and around one half of its parameters.

\begin{table*}[t]
\centering
\caption{Configuration of the LN-based teachers and students.
Block depths and widths are listed per stage for the hierarchical models and as a single value for the plain ViT.
Parameters are counted with a 10-class head, and FLOPs denote multiply--accumulate operations for one $224\times224$ image.}
\label{tab:arch_configs}
\vspace{-4pt}
\small
\setlength{\tabcolsep}{5pt}
\renewcommand{\arraystretch}{1.1}
\begin{tabular}{llcccccc}
\toprule
\textbf{Model} & \textbf{Role} & \textbf{Block depths} & \textbf{Widths} & \textbf{Heads} & \textbf{Norm.} & \textbf{Params (M)} & \textbf{FLOPs (G)} \\
\midrule
ViT-S  & Teacher & 12        & 384                & 6         & LN & 21.7 & 4.25 \\
ViT-T  & Student & 12        & 192                & 3         & LN & 5.5  & 1.08 \\
\midrule
Swin-T & Teacher & 2-2-6-2   & 96-192-384-768     & 3-6-12-24 & LN & 27.5 & 4.51 \\
Swin-N & Student & 2-2-6-2   & 48-96-192-384      & 3-6-12-24 & LN & 6.9  & 1.17 \\
\midrule
CNX-T  & Teacher & 3-3-9-3   & 96-192-384-768     & --        & LN & 27.8 & 4.47 \\
CNX-N  & Student & 2-2-8-2   & 80-160-320-640     & --        & LN & 15.0 & 2.46 \\
\bottomrule
\end{tabular}
\end{table*}

\paragraph{Comparison Methods.}
We benchmark our proposed UniDFKD against a carefully selected set of representative Data-Free Knowledge Distillation (DFKD) methods that span various technical paradigms: \textbf{DeepInversion}~\cite{yin2020dreaming}, a pioneering optimization-based method that establishes the foundational paradigm of Batch Normalization (BN) statistics matching; \textbf{Fast}~\cite{fang2022up}, a widely recognized generator-based framework serving as our efficiency-focused baseline; \textbf{CMI}~\cite{fang2021contrastive}, which introduces contrastive learning to explicitly mitigate mode collapse and represents diversity-driven synthesis; and \textbf{NAYER}~\cite{tran2024nayer}, a recent state-of-the-art approach that incorporates semantic embeddings to further enhance generative quality.
This diverse baseline selection provides a comprehensive testbed to validate the superiority and generalization capability of our framework.

\paragraph{Teacher Preparation.}
All teachers are trained by us from random initialization on the real training split of each dataset, without external data or pre-trained weights.
For RN34 we follow the protocol of prior DFKD works~\cite{fang2021contrastive,tran2024nayer}: SGD (momentum $0.9$, weight decay $5\times10^{-4}$, initial learning rate $0.1$ with cosine annealing, batch size $128$) for $200$ epochs with padded random cropping and horizontal flipping.
For the LN-based CNX, ViT, and Swin teachers we adopt the official \texttt{torchvision} recipes: AdamW (weight decay $0.05$, peak learning rate $1\times10^{-3}$, cosine schedule with linear warm-up), label smoothing of $0.1$, RandAugment, and Mixup/CutMix for $300$ epochs.
The same recipes produce the from-scratch students reported as the real-data upper bound.
All teachers are frozen in \texttt{eval()} mode and shared across all competing methods.

\paragraph{Training Details.}

With the teachers fixed, we first construct the semantic bank that drives synthesis.
Diverse yet category-faithful captions are collected by querying DeepSeek-V4~\cite{deepseekai2026deepseekv4} with the template below, yielding $200$ captions per class.
\begin{quote}\small\ttfamily
Your task is to write me 10 different image captions that include and visually describe a scene around a concept. Your concept is \{object\_text\}. Output one single grammatically correct caption that is no longer than 20 words. Do not output any notes, word counts, facts, etc. Do not use fluffy, poetic language. Output 10 single sentence only, beginning with `A photo of'. Note: Be creative and avoid repetition.
\end{quote}
Captions are encoded by a frozen OpenCLIP ViT-H/14 text encoder~\cite{cherti2023reproducible}, and the per-class embeddings are filtered by spherical $k$-means with $K=7$ and $r=15$ retained embeddings per cluster.
For SSA, the Gaussian prior is resampled per batch in units of the attribution grid $H_p\times W_p$, whose center is $(c_x,c_y)=((W_p-1)/2,(H_p-1)/2)$.
The Gaussian center is jittered symmetrically around it, $\mu_x\sim\mathcal{U}[c_x-\rho W_p,\,c_x+\rho W_p]$ and $\mu_y\sim\mathcal{U}[c_y-\rho H_p,\,c_y+\rho H_p]$, while the widths are drawn proportionally to the grid, $\sigma_x\sim\mathcal{U}[\beta_{\min}W_p,\beta_{\max}W_p]$ and $\sigma_y\sim\mathcal{U}[\beta_{\min}H_p,\beta_{\max}H_p]$, with $\rho=1/7$ and $(\beta_{\min},\beta_{\max})=(1/7,3/14)$ shared by all architectures.
Since the jitter is symmetric, the prior stays centered in expectation while its exact position and extent vary across batches; on the $14\times14$ grid of ViT this instantiates to $\mu_x,\mu_y\in[4.5,8.5]$ and $\sigma_x,\sigma_y\in[2,3]$, and the intervals scale accordingly for architectures with a different grid size.
The amplitude is fixed to $A=1$, since the cosine objective is scale-invariant and only the offset-to-amplitude ratio is identifiable; the offset is resampled per batch as $B\sim\mathcal{U}[0,0.3]$, leaving a uniform floor of up to $30\%$ of the Gaussian peak.
In the prior ablation of the main paper, the uniform prior $U(0.5)$ is a spatially uniform constant map filled with $0.5$ at every grid position, and the inverted Gaussian is $\max_{i,j}\boldsymbol{\pi}[i,j]-\boldsymbol{\pi}$.

During the data synthesis phase, we employ the Adam optimizer with a learning rate of $0.1$ to optimize each batch of $200$ images.
The number of synthesis iterations per batch is set to $40$, $60$, and $120$ for CIFAR-10, CIFAR-100, and ImageNet-100, respectively.
To ensure a stable and diverse initial data distribution, we conduct a synthesis-only warm-up for $20$ epochs to populate the replay pool prior to student training.
During synthesis, the balancing weights for the cross-entropy loss and our proposed SSA regularization are empirically set to $\lambda_{\mathrm{CE}}=1.0$ and $\lambda_{\mathrm{SSA}}=5$, respectively.
The stochasticity of the generator is handled in the same way as in NAYER~\cite{tran2024nayer}.

We emphasize that being ``architecture-agnostic'' does not imply being ``architecture-exclusive.''
The core objective of UniDFKD is to eliminate the reliance on BN statistics without preventing their utilization.
Therefore, our proposed semantic priors are designed to be universally applicable regardless of the teacher's architecture.
For BN-free teachers, these semantic priors alone provide the requisite guidance for effective synthesis, keeping the framework fully functional.
Conversely, when BN statistics are available, we retain them as an auxiliary distributional regularizer rather than discarding valuable teacher-side knowledge.
In short, BN statistics and our semantic priors are highly complementary, naturally combining to benefit BN-based teachers while maintaining robust performance for BN-free ones.

After the synthesis-only warm-up, the student is trained for $100$ effective epochs on the expanding replay pool, with $400$ updates per epoch and a batch size of $256$.
Following standard DFKD protocols~\cite{fang2021contrastive,tran2024nayer}, the overall optimization objective comprises a KL divergence loss to align the student's predictions with the teacher's soft labels.
We use AdamW ($\beta_1,\beta_2=0.9,0.999$; weight decay $0.05$) with a peak learning rate of $1\times10^{-3}$, a $20$-epoch linear warm-up followed by cosine decay to $1\times10^{-6}$.
Replay samples are augmented by random resized cropping (scale $[0.6,1.0]$, bicubic), horizontal flipping, RandAugment ($2$ operations, magnitude $5$), random erasing ($p=0.25$), and Mixup or CutMix with equal probability ($\alpha_{\mathrm{mix}}=0.8$, $\alpha_{\mathrm{cut}}=1.0$); the teacher and the student receive the same view.
We fix $T=20$ and $\lambda_{\mathrm{SSD}}=3$ in all experiments, and report the best top-$1$ validation accuracy averaged over three random seeds ($0$, $1$, and $2$).
To ensure a fair comparison, all baseline methods are trained with the same optimizer, learning rate schedule, data augmentation pipeline, and number of distillation iterations as UniDFKD.
Furthermore, we comprehensively sweep their respective method-specific hyperparameter combinations (e.g., the number of inversion steps per batch or the loss weights) on the evaluated benchmarks, reporting the best-performing configurations for each method.
All experiments were implemented in PyTorch 2.5.1 and conducted on two NVIDIA A800 GPUs (80GB).

\section{Extensive Quantitative Results}
\label{app:quantitative}

\begin{table*}[t]
\centering
\caption{Performance comparison of existing DFKD methods and UniDFKD.
The best results are highlighted in bold and the runner-up is \underline{underlined}.
The results of previous methods are taken from their original papers.
``RN'' indicates ResNet, ``W'' corresponds to WideResNet, and ``V'' stands for VGG in this paper.
``-'' denotes the result is not reported.}
\label{tab:performance_comparison}
\vspace{-4pt}
\renewcommand{\arraystretch}{1.1}
\resizebox{\textwidth}{!}{%
\begin{tabular}{lccccccccccc}
\toprule
& \multicolumn{5}{c}{\textbf{CIFAR-10}}
& \multicolumn{5}{c}{\textbf{CIFAR-100}}
& \multicolumn{1}{c}{\textbf{TinyImageNet}} \\
\cmidrule(lr){2-6}
\cmidrule(lr){7-11}
\cmidrule(lr){12-12}
\textbf{Method}
& \textbf{\shortstack{RN34\\RN18}}
& \textbf{\shortstack{W40-2\\W16-2}}
& \textbf{\shortstack{W40-2\\W16-1}}
& \textbf{\shortstack{W40-2\\W40-1}}
& \textbf{\shortstack{V11\\RN18}}
& \textbf{\shortstack{RN34\\RN18}}
& \textbf{\shortstack{W40-2\\W16-2}}
& \textbf{\shortstack{W40-2\\W16-1}}
& \textbf{\shortstack{W40-2\\W40-1}}
& \textbf{\shortstack{V11\\RN18}}
& \textbf{\shortstack{RN34\\RN18}} \\
\midrule
Teacher
& 95.70  & 94.87 & 94.87 & 94.87 & 92.25
& 78.05 & 75.83 & 75.83 & 75.83 & 71.32
& 66.44 \\
Student
& 95.20  & 93.95 & 91.12 & 93.94 & 95.20
& 77.10  & 73.56 & 65.31 & 72.19 & 77.10
& 64.87 \\
\midrule
DeepInversion~\cite{yin2020dreaming}
& 89.69 & 89.72 & 83.04 & 86.85 & 90.36
& 61.32 & 61.34 & 53.77 & 68.58 & 54.13
& - \\
DAFL~\cite{chen2019data}
& 92.22 & 81.55 & 65.71 & 81.33 & 81.10
& 74.47 & 43.70  & 20.88 & 42.83 & 54.16
& - \\
DFQ~\cite{choi2020data}
& 94.61 & 92.01 & 86.14 & 91.69 & 90.84
& 77.01 & 64.79 & 51.27 & 54.43 & 66.21
& - \\
ZSKT~\cite{micaelli2019zero}
& 93.32 & 89.66 & 83.74 & 86.07 & 89.46
& 67.74 & 54.59 & 36.60  & 53.60  & 54.31
& - \\
CMI~\cite{fang2021contrastive}
& 94.84 & 92.52 & 90.01 & 92.78 & 91.13
& 77.04 & 68.75 & 57.91 & 68.88 & 70.56
& 64.01 \\
Fast~\cite{fang2022up}
& 94.05 & 92.45 & 89.29 & 92.51 & 90.53
& 74.34 & 65.12 & 54.02 & 63.91 & 67.44
& - \\
MAD~\cite{do2022momentum}
& 94.90  & 92.64 & -     & -     & -
& 77.31 & 64.05 & -     & -     & -
& 62.32 \\
SpaceshipNet~\cite{yu2023data}
& \textbf{95.39} & 93.25 & 90.38 & 93.56 & 92.27
& 77.41 & 69.95 & 58.06 & 68.78 & 71.41
& 64.04 \\
SSD-KD~\cite{liu2024small}
& 94.26 & 93.11 & 89.96 & 93.23 & 90.67
& 75.16 & 65.28 & 55.61 & 64.57 & 68.77
& - \\
NAYER~\cite{tran2024nayer}
& 95.21
& 94.07
& 91.94
& 94.15
& \underline{92.37}
& \underline{77.54}
& 71.72
& 62.23
& 71.80
& 71.75
& 64.17 \\
CPNet~\cite{chen2025cpnet}
& 95.26
& \underline{94.09}
& \textbf{92.00}
& \textbf{94.22}
& 92.18
& \textbf{77.68}
& \textbf{71.90}
& \textbf{63.23}
& \underline{71.98}
& \underline{71.80}
& \textbf{64.91} \\
\midrule
\rowcolor{tableblue}
UniDFKD (Ours)
& \underline{95.33} & \textbf{94.10} & \underline{91.98} & \underline{94.19} & \textbf{92.44}
& 77.37 & \underline{71.77} & \underline{62.65} & \textbf{71.99} & \textbf{71.85}
& \underline{64.71} \\
\bottomrule
\end{tabular}%
}
\end{table*}

\subsection{Comparison on General CNN Architectures}
\label{app:general_cnn}
Table~\ref{tab:performance_comparison} compares existing data-free knowledge distillation (DFKD) methods with our UniDFKD on conventional CNN architectures.
As observed, previous approaches have already achieved remarkable success on these standard benchmarks, with their distillation performance closely approaching the accuracy of student models trained on real data.
The performance ceiling on these conventional CNNs is therefore highly saturated, making it exceedingly difficult for new methods to obtain large absolute accuracy gains.

Despite this saturation, our proposed UniDFKD remains highly competitive.
Most notably, compared with CPNet, the latest state-of-the-art method that heavily relies on architecture-specific BN priors, UniDFKD delivers entirely on-par performance.
Under several configurations, it even surpasses all existing baselines and establishes new state-of-the-art records.
For instance, UniDFKD achieves the highest accuracy in the W40-2$\to$W16-2 (94.10) and V11$\to$RN18 (92.44) settings on CIFAR-10, as well as in the W40-2$\to$W40-1 (71.99) and V11$\to$RN18 (71.85) settings on CIFAR-100.

These quantitative results indicate that the advantages of UniDFKD are not restricted to BN-free or heterogeneous scenarios.
While it fundamentally resolves the catastrophic degradation of existing methods on modern architectures, it simultaneously retains state-of-the-art efficacy in traditional BN-based settings, validating the universality and robustness of our explicit semantic prior framework.

\begin{table}[t]
\centering
\caption{Comparison of distillation accuracy (\%) and training time (hours) on CIFAR-10 and CIFAR-100 under the ViT-S$\to$ViT-T setting.
Our method not only achieves the highest distillation performance but also significantly improves synthesis efficiency.
The ``Average Speed Up'' is calculated relative to the DeepInversion baseline.}
\label{tab:training_time}
\vspace{-4pt}
\resizebox{\linewidth}{!}{
\begin{tabular}{lccccc}
\toprule
& DeepInversion & CMI & Fast & NAYER & \cellcolor{gray!15}Ours \\
\midrule

\textbf{CIFAR-10}
& 53.34 & 84.19 & 83.71 & 85.86 & \cellcolor{gray!15}\textbf{93.91} \\

\textbf{Time (hours)}
 &  36.87  &  33.43  &  5.97     & 5.69   &  \cellcolor{gray!15}\textbf{3.36} \\
\midrule

\textbf{CIFAR-100}
& 14.51 & 42.42 & 23.65 & 57.51 & \cellcolor{gray!15}\textbf{70.41} \\

\textbf{Time (hours)}
& 37.22 & 35.70 & 6.99 & 6.86 & \cellcolor{gray!15}\textbf{3.65} \\

\midrule

\textbf{Average Speed Up}
& $1.00\times$ & $1.07\times$ & $5.75\times$ & $5.95\times$ & \cellcolor{gray!15}\textbf{10.59$\times$} \\

\bottomrule
\end{tabular}
}
\end{table}

\subsection{Computational Efficiency Comparison}
Data-free knowledge distillation traditionally suffers from substantial computational overhead, primarily because the image synthesis process is iterative and time-consuming.
In Table~\ref{tab:training_time}, we therefore evaluate the computational efficiency of our method against existing baselines under the ViT-S$\to$ViT-T setting on the CIFAR-10 and CIFAR-100 datasets.

As shown in Table~\ref{tab:training_time}, earlier inversion-based methods such as DeepInversion and CMI are highly inefficient, requiring between 33 and 38 hours to complete the distillation process.
Recent methods such as Fast and NAYER improve efficiency by accelerating the synthesis phase, reducing the time to approximately 5.6 to 7 hours, yet UniDFKD pushes this boundary significantly further.
Our method requires only 3.36 hours on CIFAR-10 and 3.65 hours on CIFAR-100, achieving a remarkable average speedup of $10.59\times$ relative to the DeepInversion baseline.
This speedup stems from differences in sample efficiency across methods: guided by explicit semantic priors, UniDFKD reaches stronger distillation performance using substantially fewer synthesized samples.

Crucially, this exceptional efficiency does not come at the expense of distillation quality.
On the contrary, UniDFKD simultaneously achieves the highest distillation accuracy, outperforming the second-fastest method (NAYER) by a substantial absolute margin of 8.05 (93.91 vs. 85.86) on CIFAR-10 and 12.90 (70.41 vs. 57.51) on CIFAR-100.
This demonstrates that, by leveraging explicit architecture-agnostic semantic priors, UniDFKD not only provides stronger and more accurate guidance for high-fidelity data generation, but also facilitates much faster convergence.
Overall, UniDFKD yields a markedly superior trade-off between training efficiency and distillation performance compared with existing approaches~\cite{tran2024nayer}.

\begin{table}[t]
    \centering
    \caption{Accuracy (\%) on ImageNet-1K under the ViT-S$\to$ViT-T setting with limited synthesis budgets.
    The ``Data Ratio'' (1\% and 5\%) indicates the number of synthesized images relative to the full scale of the original ImageNet-1K training set.
    UniDFKD significantly outperforms both CMI and NAYER across all extremely low data regimes.}
    \label{tab:IN-1K}
    \vspace{-4pt}
    \renewcommand{\arraystretch}{1.05}
    \small
    \begin{tabular}{l|cc|cc}
        \toprule
        \textbf{Data Ratio}
        & \multicolumn{2}{c|}{\textbf{1\%}}
        & \multicolumn{2}{c}{\textbf{5\%}} \\
        \midrule
        \textbf{Accuracy}
        & \textbf{Top 1}
        & \textbf{Top 5}
        & \textbf{Top 1}
        & \textbf{Top 5} \\
        \midrule
        CMI
        & 3.61
        & 12.44
        & 16.22
        & 42.71 \\
        NAYER 
        & 2.68
        & 9.45
        & 18.27
        & 46.55 \\
\rowcolor{gray!15}        UniDFKD
        & 19.18
        & 41.72
        & 36.32
        & 66.24 \\
        \bottomrule
    \end{tabular}
\end{table}

\subsection{Evaluation on Large-Scale Datasets}
Extending DFKD to large-scale datasets such as ImageNet-1K (IN-1K) inherently poses severe computational and storage challenges.
Synthesizing the full scale of the IN-1K training dataset (over 1.2 million images at a high resolution) consumes prohibitive amounts of GPU memory and disk storage, rendering it practically infeasible in standard resource-constrained environments.
Consequently, to demonstrate the superiority and practicality of our method on large-scale datasets, we evaluate its robustness under extremely limited synthesis budgets, specifically synthesizing only 1\% and 5\% of the original IN-1K dataset scale.

Table~\ref{tab:IN-1K} presents the top-1 and top-5 distillation accuracy on IN-1K under the ViT-S$\to$ViT-T setting.
In the extremely stringent 1\% data regime, existing baseline methods suffer from catastrophic performance degradation: NAYER and CMI achieve a mere 2.68 and 3.61 top-1 accuracy, respectively, indicating their inability to extract effective task-relevant distributions with so few synthesis iterations.
In stark contrast, UniDFKD exhibits exceptional robustness, securing 19.18 top-1 and 41.72 top-5 accuracy under the same 1\% budget.

As the synthesis budget increases to 5\%, the performance gap widens further.
UniDFKD achieves 36.32 top-1 and 66.24 top-5 accuracy, doubling the top-1 accuracy of NAYER (18.27) and CMI (16.22).
These results demonstrate that our unified semantic prior framework effectively maximizes the information density of each generated sample.
By enforcing aligned semantic structures and class-discriminative spatial localization, UniDFKD achieves highly efficient knowledge transfer even with extremely sparse synthetic data, highlighting its scalability and robustness for large-scale applications.

\begin{table}[t]
    \centering
    \caption{Ablation on the text encoder used to build the class-semantic bank.
    We report student accuracy (\%) on CIFAR-10 and CIFAR-100 under the ViT-S$\to$ViT-T setting for two vision--language encoders (CLIP ViT-H/14 and ViT-B/32) and a vision-agnostic, language-only encoder (Qwen3-Embedding-0.6B), each evaluated without (w/o Cent.) and with (w/ Cent.) mean-centered prompt embeddings.}
    \label{tab:text_encoder_ablation}
    \vspace{-4pt}
    \renewcommand{\arraystretch}{1.15}
    \small
    \setlength{\tabcolsep}{2pt}
    \begin{tabular}{lc>{\columncolor{black!8}}cc>{\columncolor{black!8}}c}
        \toprule
        \multirow{2}{*}{Embedding Model}
        & \multicolumn{2}{c}{CIFAR-10}
        & \multicolumn{2}{c}{CIFAR-100} \\
        \cmidrule(lr){2-3}
        \cmidrule(lr){4-5}
        & w/o Cent. & w/ Cent.
        & w/o Cent. & w/ Cent. \\
        \midrule
        CLIP ViT-B/32
        & 93.50 & 93.88
        & 69.10 & 70.40 \\
        CLIP ViT-H/14
        & 93.91 & \textbf{93.91}
        & 69.79 & \underline{70.41} \\
        Qwen3-Emb.-0.6B
        & 93.79 & \textbf{93.91}
        & 68.53 & \textbf{70.44} \\
        \bottomrule
    \end{tabular}
\end{table}

\subsection{Robustness to Vision-Agnostic Text Encoders}
\label{subsec:textenc}
To construct the categorical semantic bank, our primary experiments utilize CLIP, a vision--language model natively aligned with visual representations.
A compelling question therefore arises: \textit{does the success of our explicit semantic prior rely strictly on this pre-existing cross-modal alignment?} To address it, Table~\ref{tab:text_encoder_ablation} investigates the robustness of UniDFKD by substituting CLIP with a purely vision-agnostic language encoder (Qwen3-Embedding-0.6B~\cite{zhang2025qwen3embedding}).

As illustrated in the ``w/o Cent.'' (without mean-centering) columns, directly utilizing raw embeddings from the pure language model yields highly competitive results on CIFAR-10 (93.79).
It nevertheless exhibits a noticeable performance drop on the more challenging CIFAR-100 dataset (68.53) compared with uncentered CLIP ViT-H/14 (69.79).
This discrepancy is expected, as embeddings from pure language models typically suffer from representation anisotropy (i.e., vectors occupying a narrow cone in the latent space) and lack explicit geometric connections to visual structures.

Crucially, with mean-centering (``w/ Cent.'') applied to calibrate the embedding space---shifting the textual embeddings by their global mean to mitigate modality bias---the performance of the vision-agnostic encoder recovers dramatically.
The centered Qwen3 embeddings achieve 93.91 on CIFAR-10 and 70.44 on CIFAR-100, which matches and even marginally surpasses our default vision-aligned CLIP ViT-H/14 model (93.91 and 70.41, respectively).

These empirical findings substantiate a core mechanism of UniDFKD: the efficacy of Categorical Semantic Conditioning (CSC) does not stem from any leakage of cross-modal visual memory.
Instead, it leverages the rich and structurally consistent relational hierarchies---such as inter-class semantic distances and intra-class descriptive diversity---that are intrinsically embedded within pure language models.
This structural alignment effectively guides the generative process and confirms the generalization and robustness of our framework across text encoders.

\section{Qualitative Visualizations}
\label{app:qualitative}

\subsection{Synthetic Image Comparisons}
\label{app:qualitative_synth}
\begin{figure*}[t]
    \centering

    \begin{subfigure}[t]{0.33\linewidth}
        \centering
        \includegraphics[width=\linewidth]{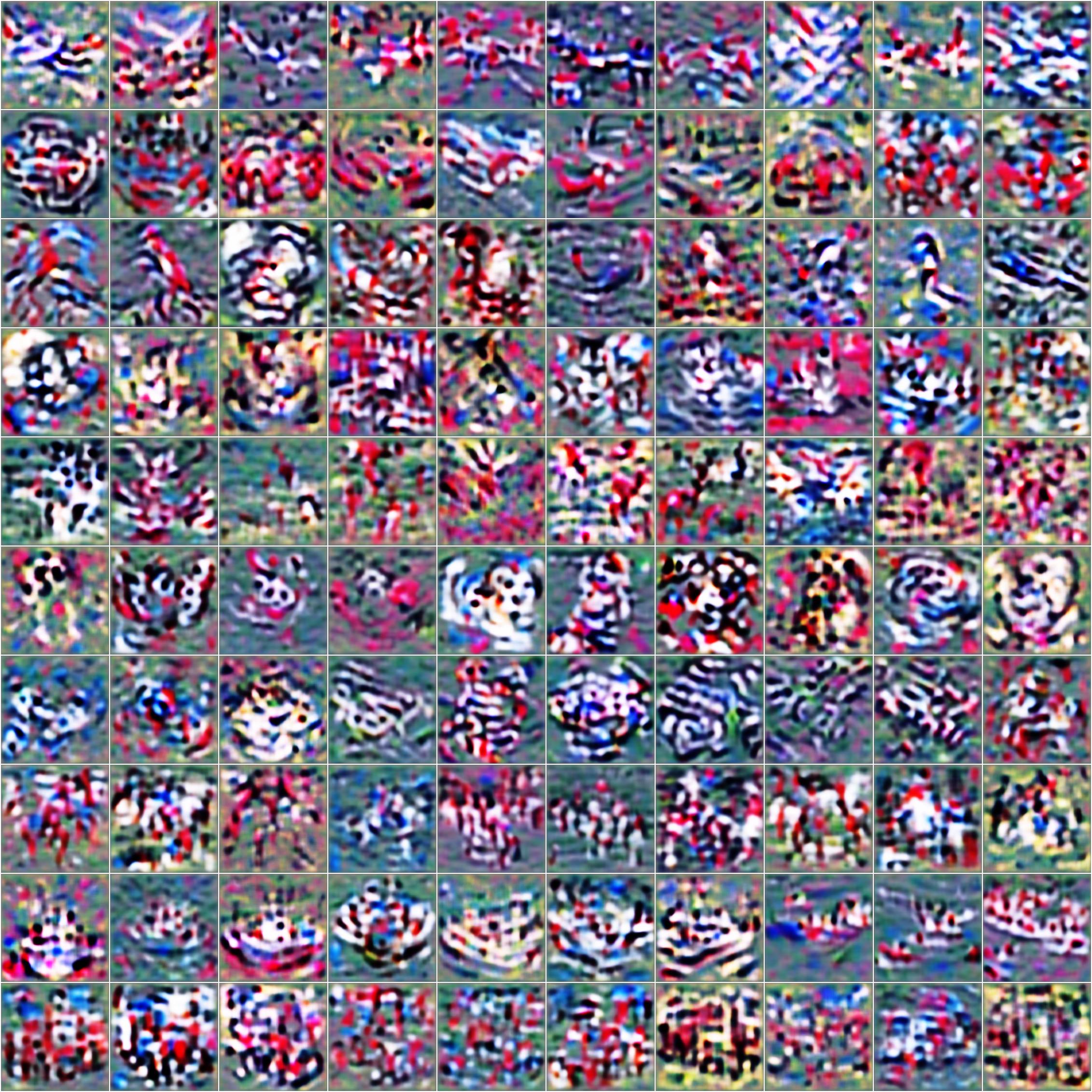}
        \caption{CMI (CIFAR-10)}
        \label{fig:appendix_img_ci10_cmi}
    \end{subfigure}
        \hfill
    \begin{subfigure}[t]{0.33\linewidth}
        \centering
        \includegraphics[width=\linewidth]{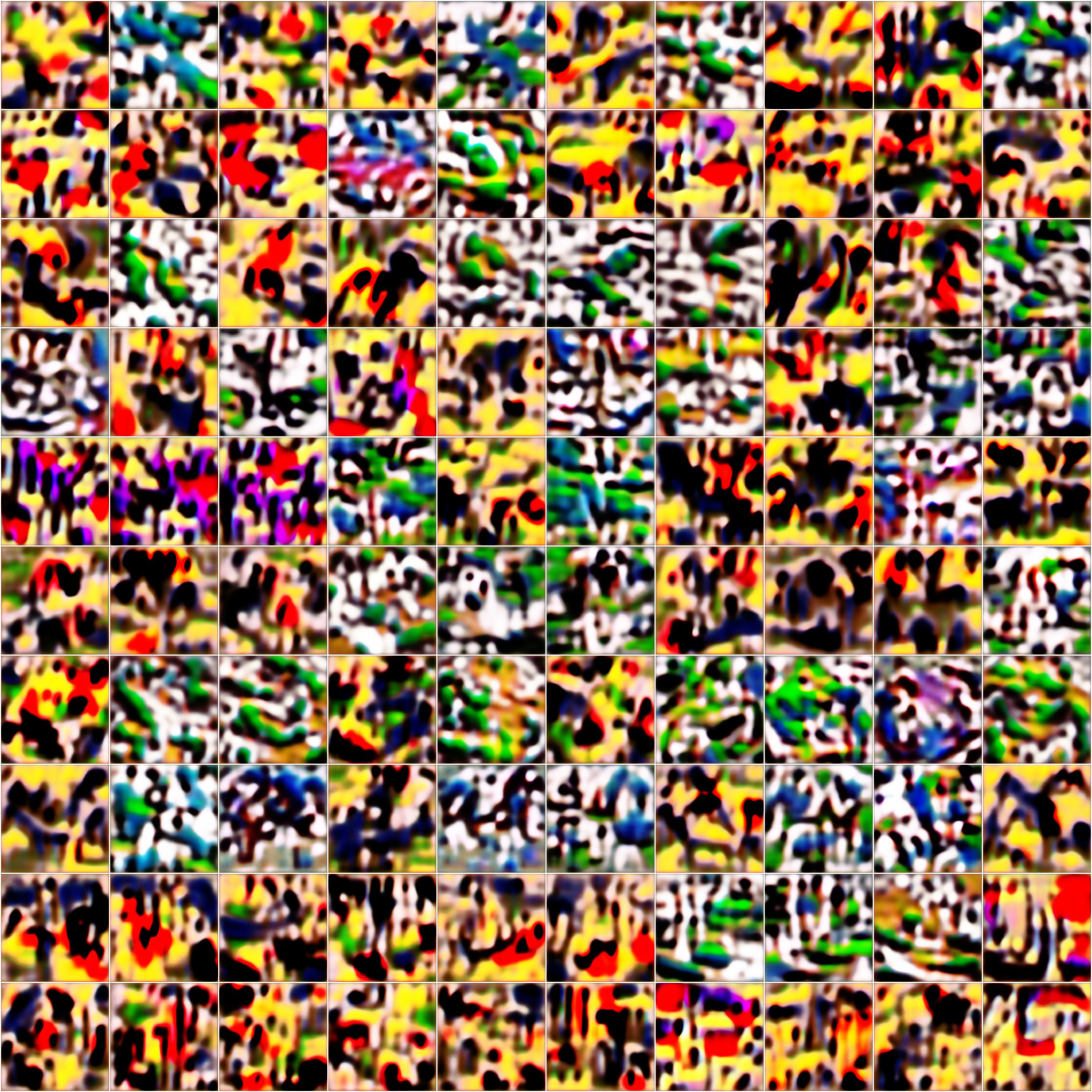}
        \caption{NAYER (CIFAR-10)}
        \label{fig:appendix_img_ci10_nayer}
    \end{subfigure}
    \begin{subfigure}[t]{0.33\linewidth}
        \centering
        \includegraphics[width=\linewidth]{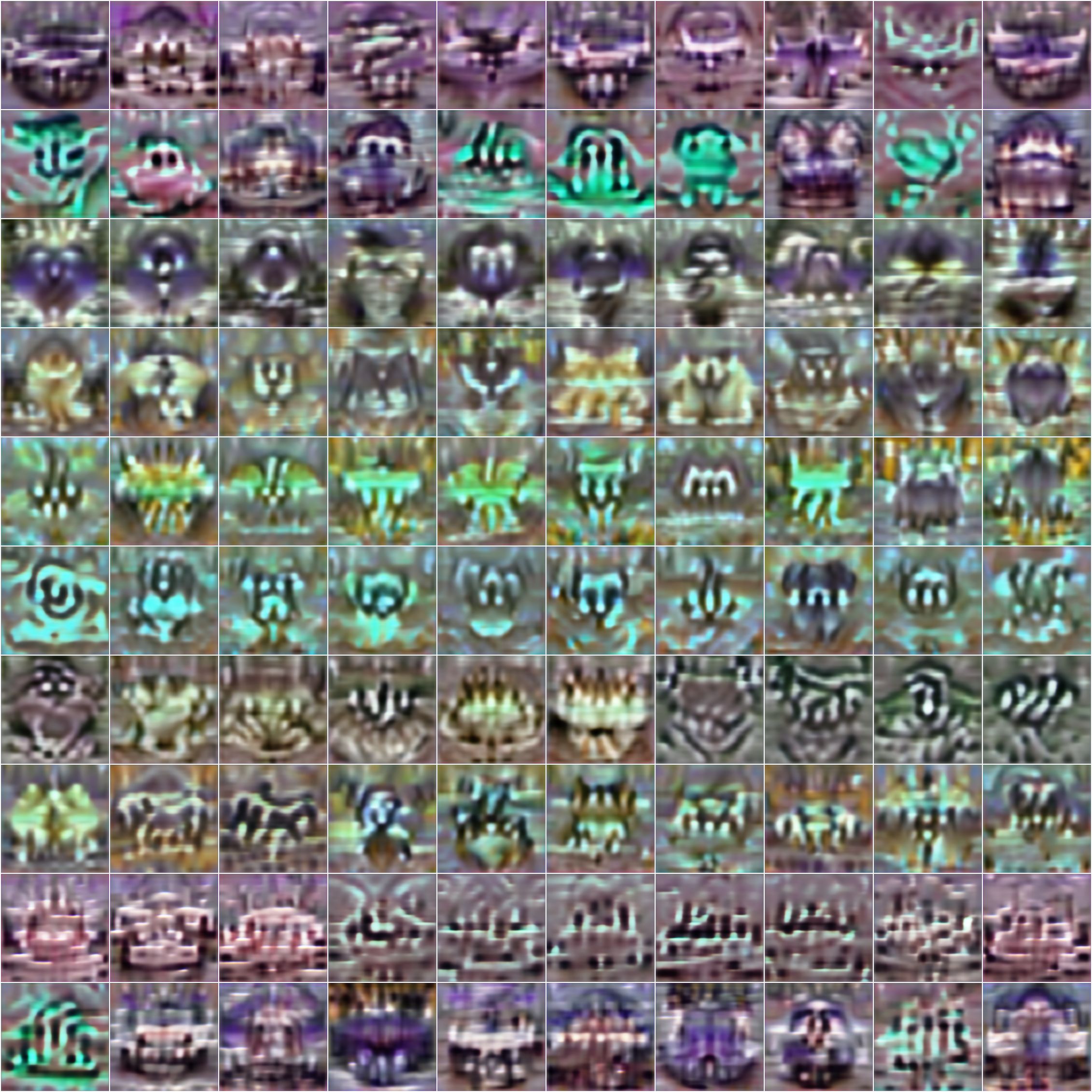}
        \caption{UniDFKD (CIFAR-10)}
        \label{fig:appendix_img_ci10_unidfkd}
    \end{subfigure}
    \begin{subfigure}[t]{0.33\linewidth}
        \centering
        \includegraphics[width=\linewidth]{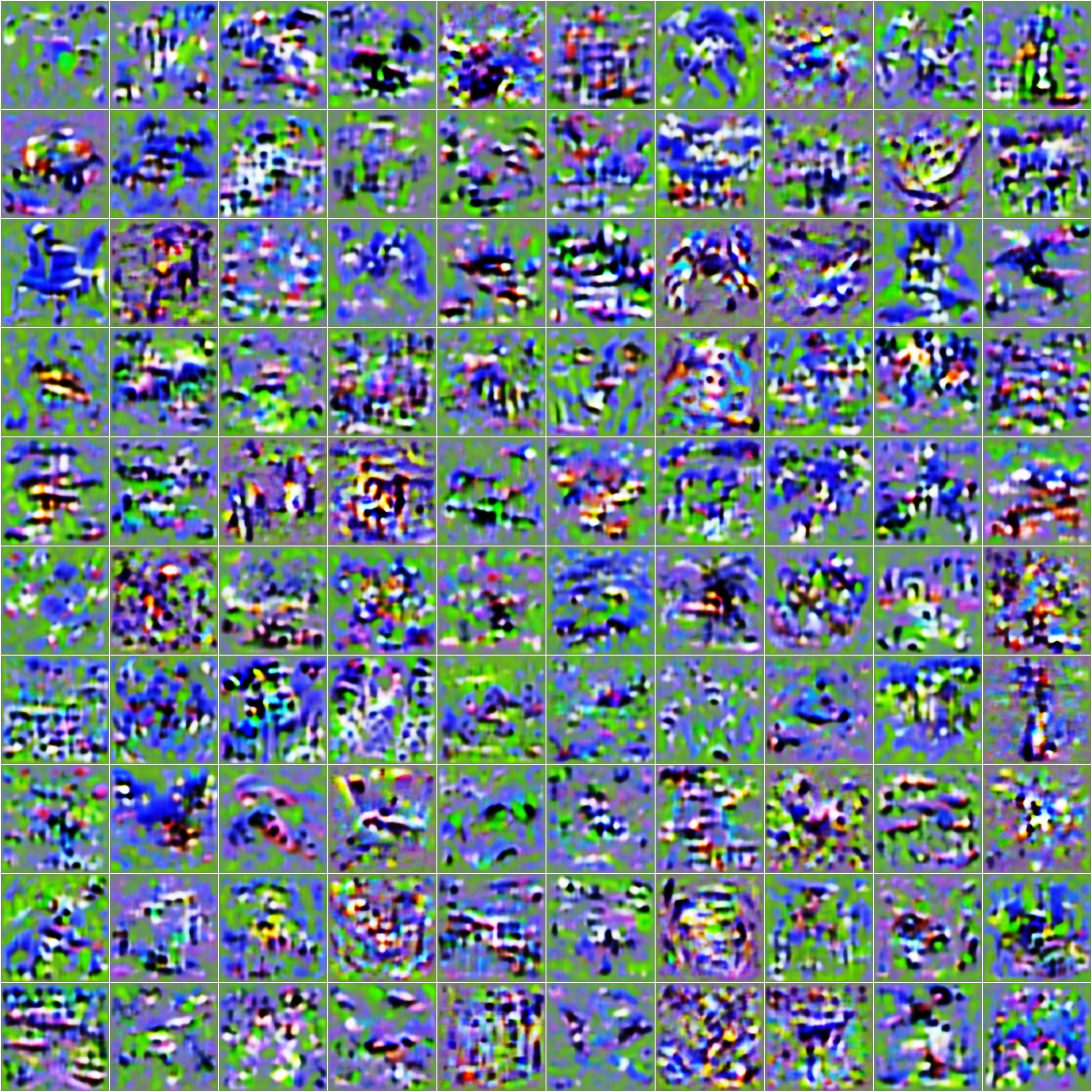}
        \caption{CMI (CIFAR-100)}
    \end{subfigure}
    \begin{subfigure}[t]{0.33\linewidth}
        \centering
        \includegraphics[width=\linewidth]{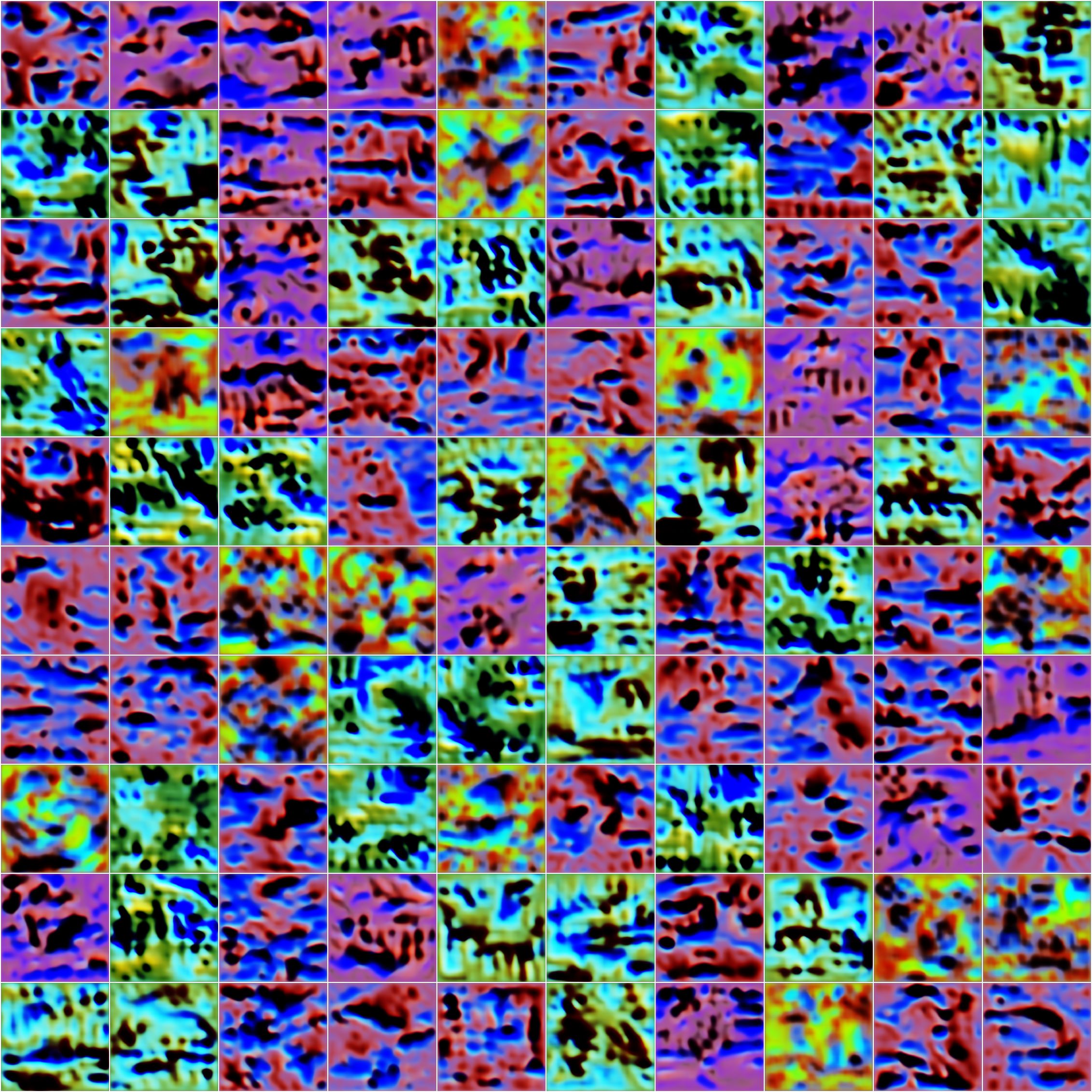}
        \caption{NAYER (CIFAR-100)}
    \end{subfigure}
    \begin{subfigure}[t]{0.33\linewidth}
        \centering
        \includegraphics[width=\linewidth]{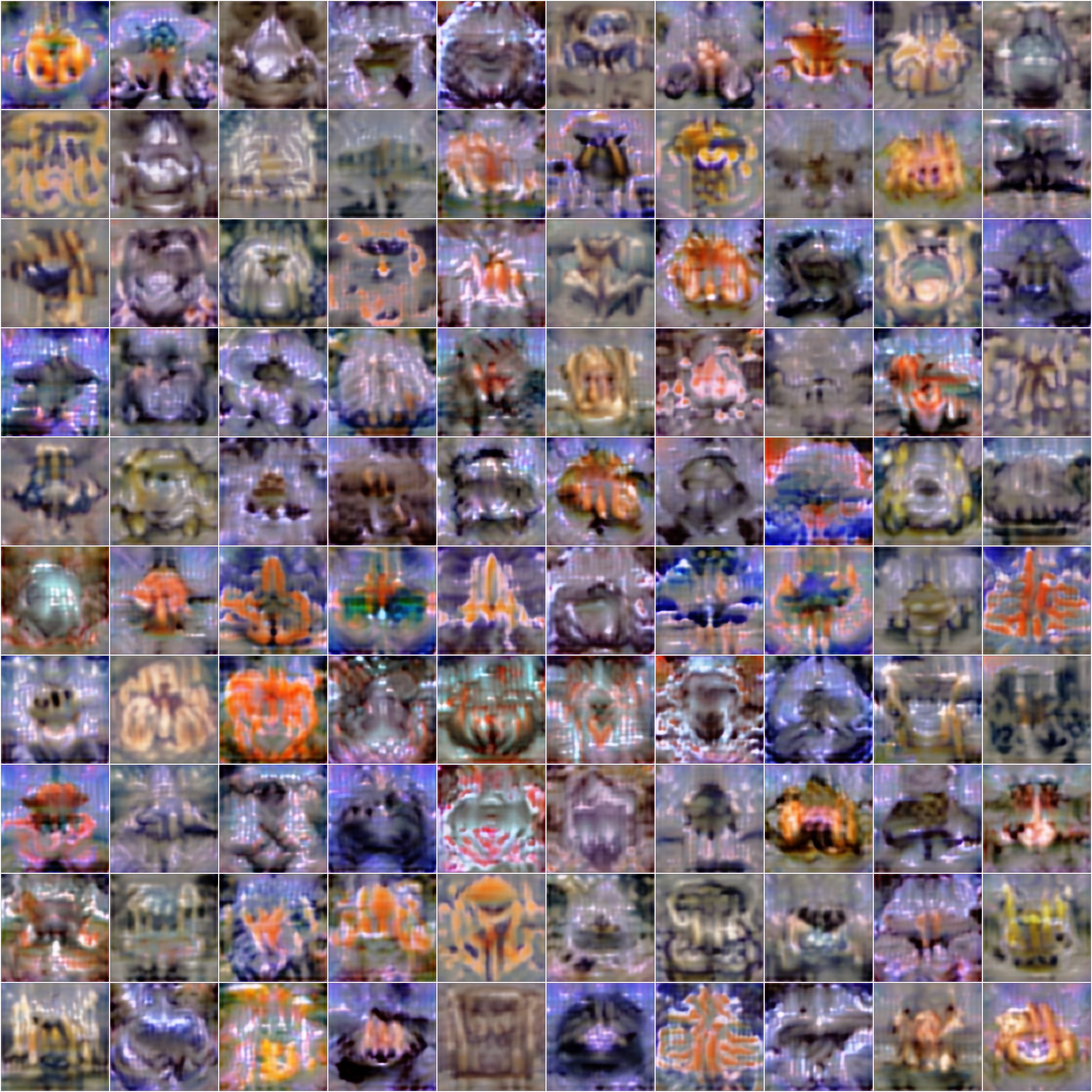}
        \caption{UniDFKD (CIFAR-100)}
    \end{subfigure}
    \begin{subfigure}[t]{0.33\linewidth}
        \centering
        \includegraphics[width=\linewidth]{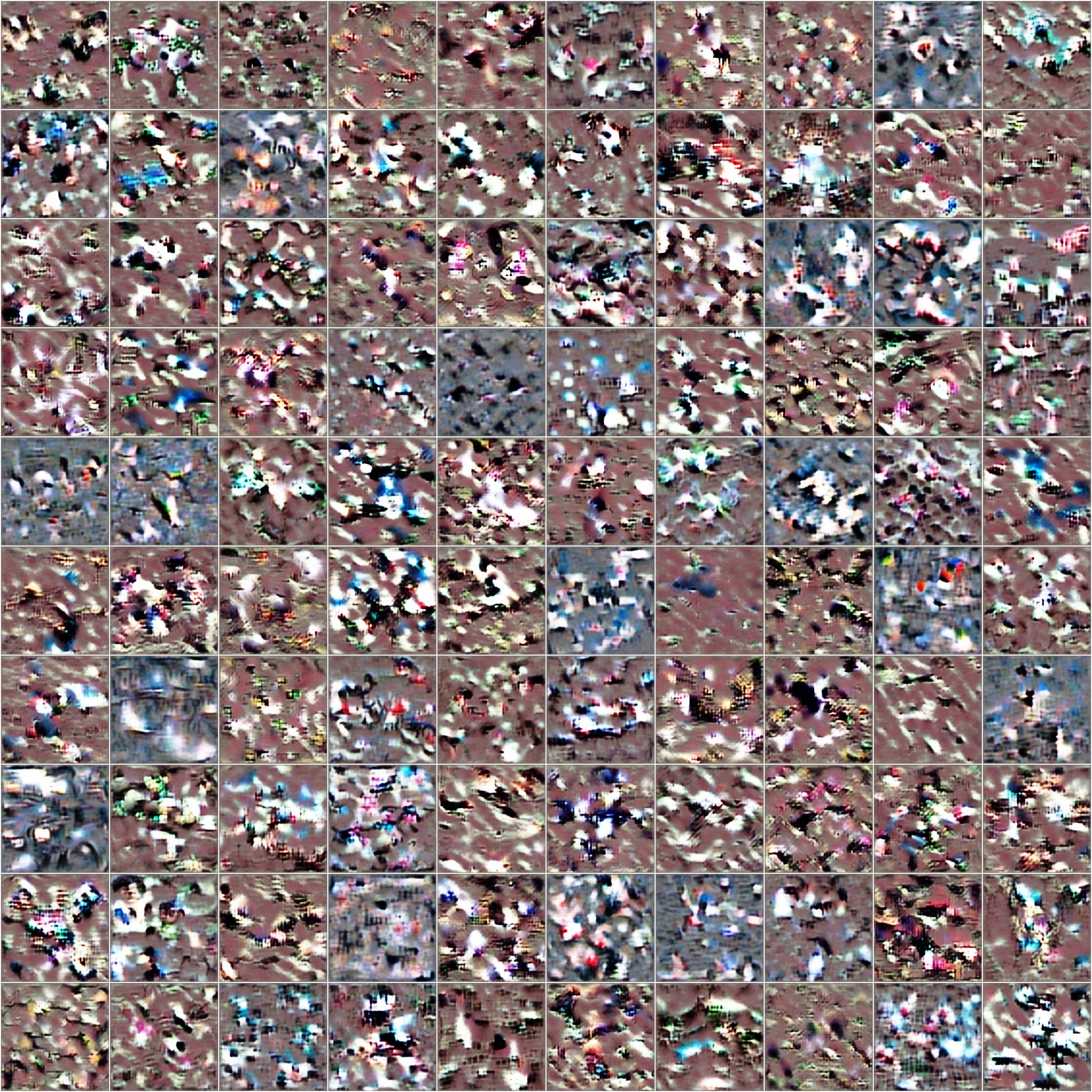}
        \caption{CMI (ImageNet-100)}
    \end{subfigure}
    \begin{subfigure}[t]{0.33\linewidth}
        \centering
        \includegraphics[width=\linewidth]{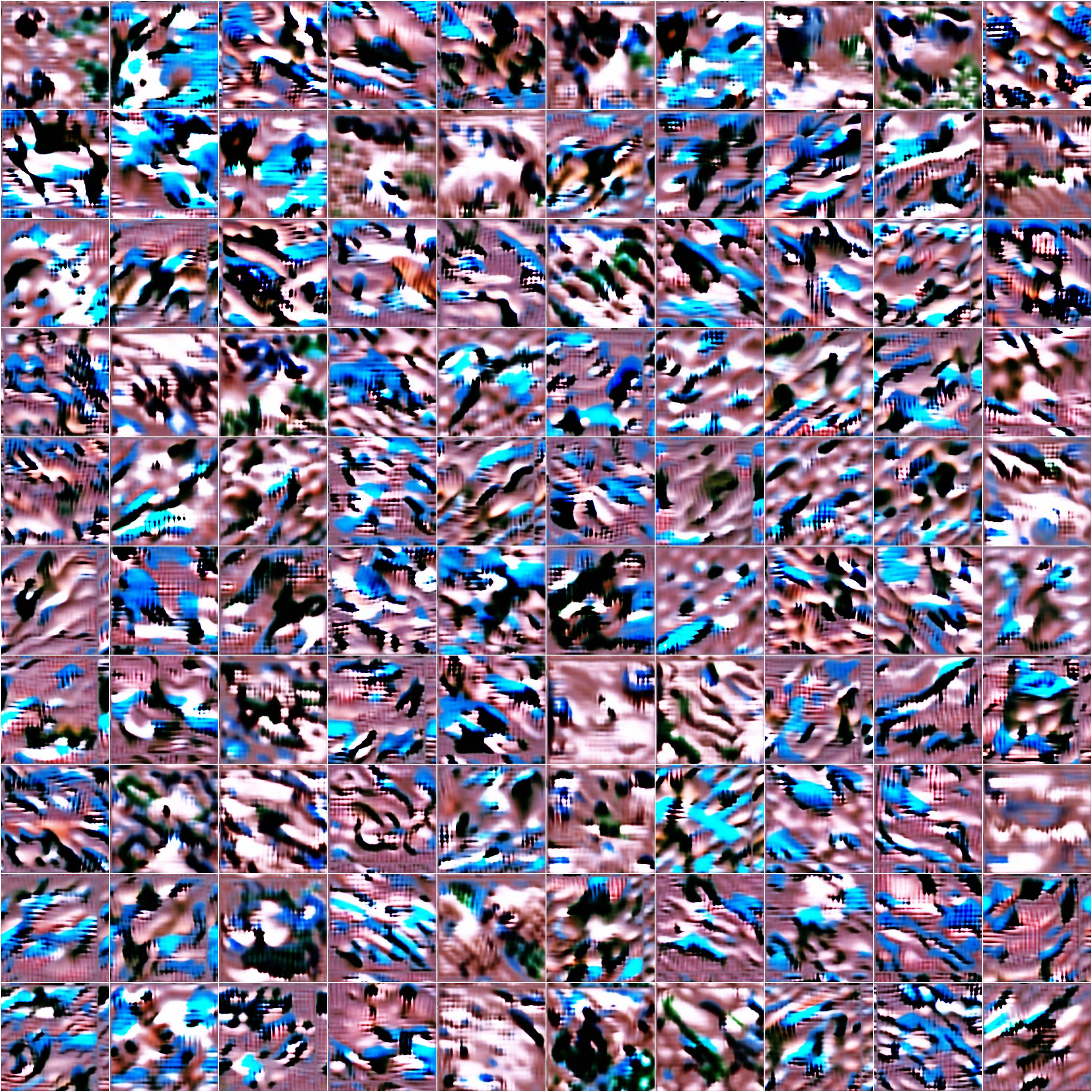}
        \caption{NAYER (ImageNet-100)}
    \end{subfigure}
    \begin{subfigure}[t]{0.33\linewidth}
        \centering
        \includegraphics[width=\linewidth]{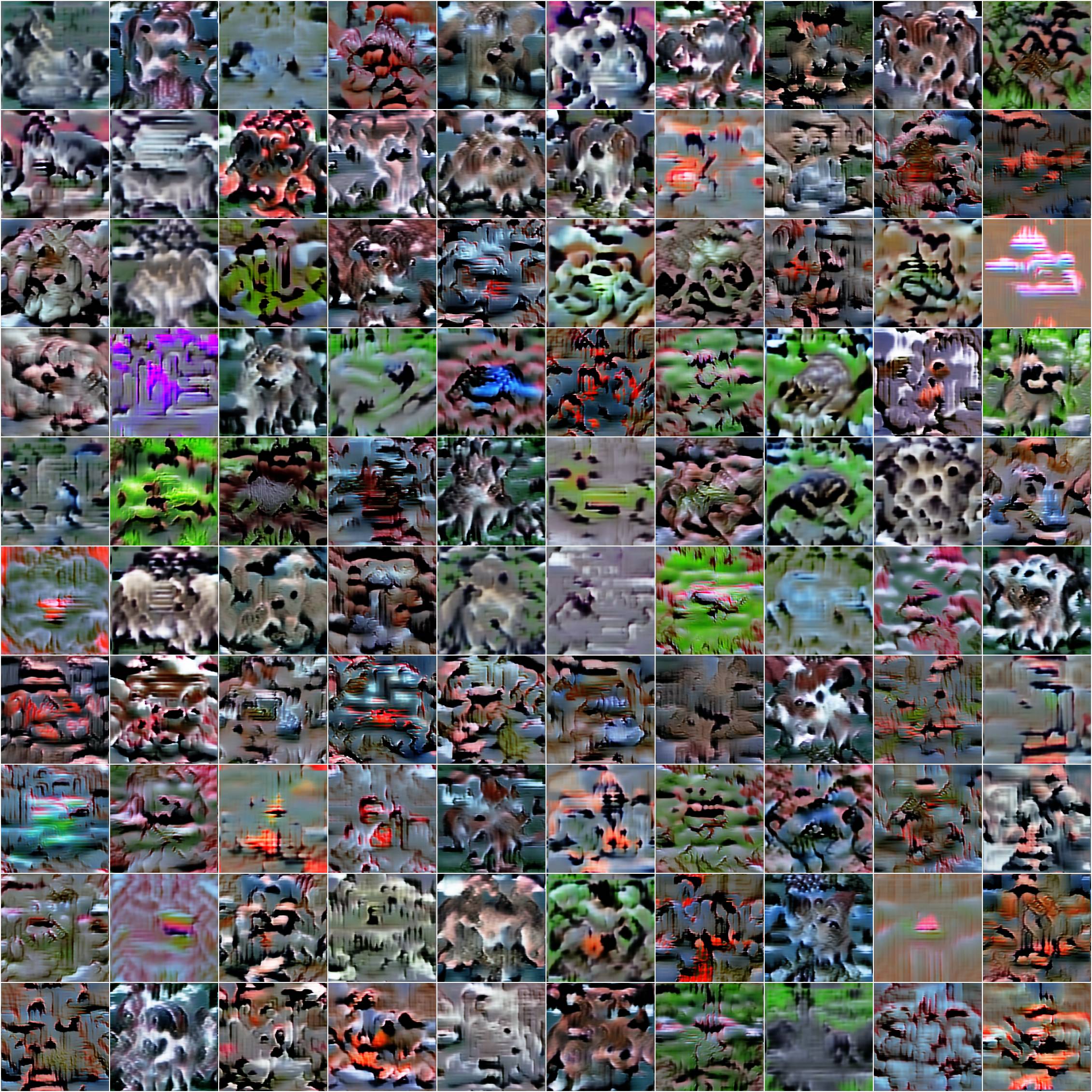}
        \caption{UniDFKD (ImageNet-100)}
    \end{subfigure}
\caption{Representative synthetic images generated using ViT-S as the teacher on CIFAR-10, CIFAR-100, and ImageNet-100.
Rows correspond to datasets, and columns compare CMI, NAYER, and UniDFKD.}
\label{fig:qualitative_visualization}
\end{figure*}

Figure~\ref{fig:qualitative_visualization} compares synthetic samples generated by CMI, NAYER, and UniDFKD using ViT-S teachers on CIFAR-10, CIFAR-100, and ImageNet-100.
CMI mainly produces fragmented high-frequency patterns, while NAYER improves semantic diversity but still exhibits strong texture artifacts and irregular spatial layouts.
In contrast, UniDFKD generates samples with more coherent class-related structures and better-localized discriminative regions.
This advantage becomes more evident on ImageNet-100, where UniDFKD recovers richer object- and scene-level patterns rather than relying on repetitive local textures.
These observations suggest that CSC provides category-consistent semantic guidance, while SSA organizes discriminative evidence into spatially meaningful regions without sacrificing diversity.
Overall, the qualitative results indicate that UniDFKD synthesizes more informative and structurally coherent samples, consistent with its superior distillation performance.

\subsection{Student Spatial Attribution Alignment under SSD}
\label{app:qualitative_student_cdsa}
\begin{figure*}[t]
    \centering
    \includegraphics[width=\linewidth]{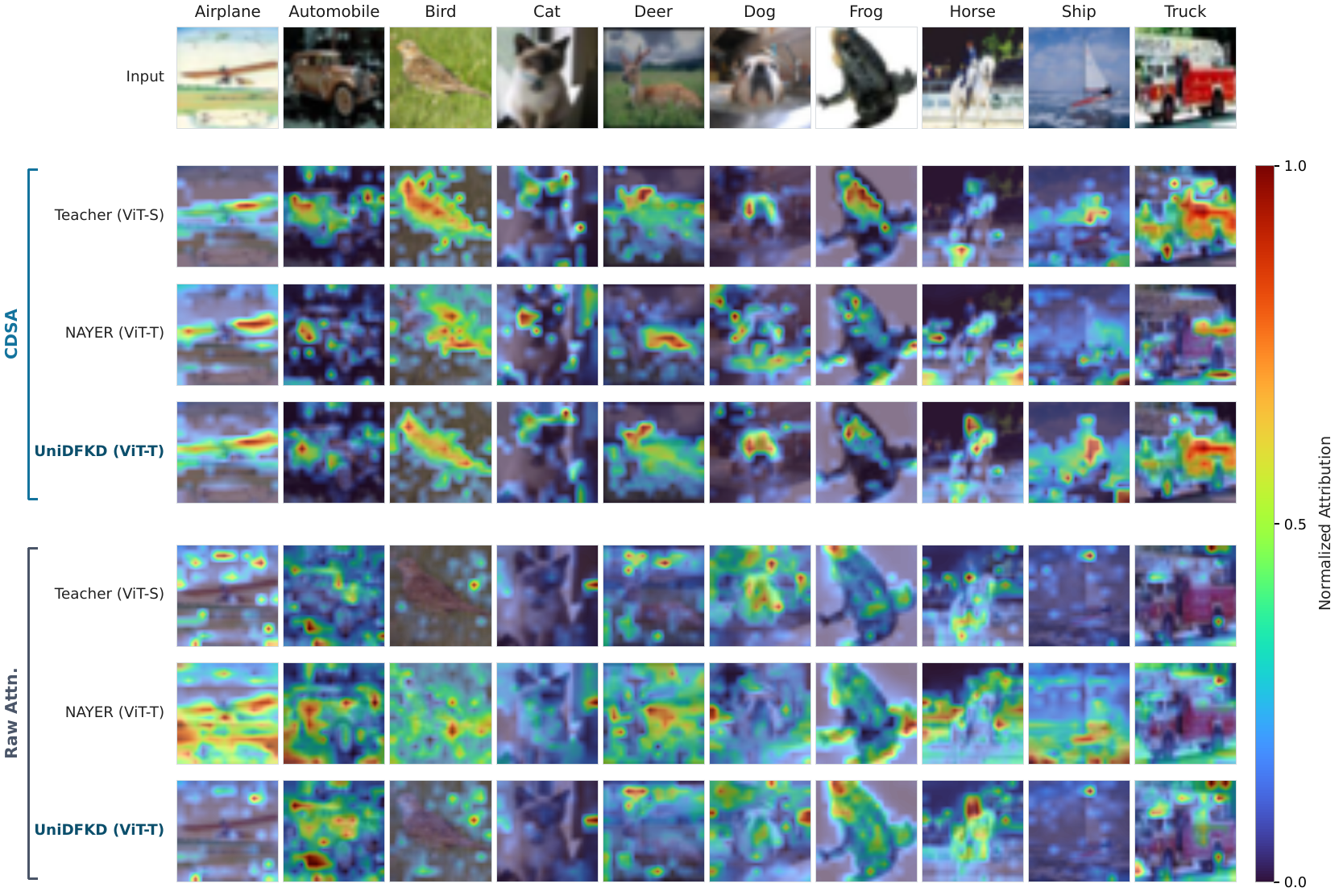}
    \caption{Qualitative spatial attribution comparison between the ViT-S teacher and ViT-T students distilled on CIFAR-10 without spatial attribution supervision (NAYER) and with SSD (UniDFKD).
    We evaluate ten representative real CIFAR-10 test images (one per category, from Airplane to Truck).
    For each sample, non-negative $14\times 14$ target-class CDSA maps and last-block raw CLS-to-patch attention maps are bilinearly upsampled to $224\times 224$, min--max normalized per map for visualization, and overlaid onto the input image.
    NAYER scatters attribution onto background regions and peripheral textures, whereas UniDFKD closely mirrors the teacher's object-centric spatial evidence in both CDSA and raw attention.}
    \label{fig:appendix_student_cdsa}
\end{figure*}

\begin{figure}[t]
    \centering
    \includegraphics[width=\linewidth]{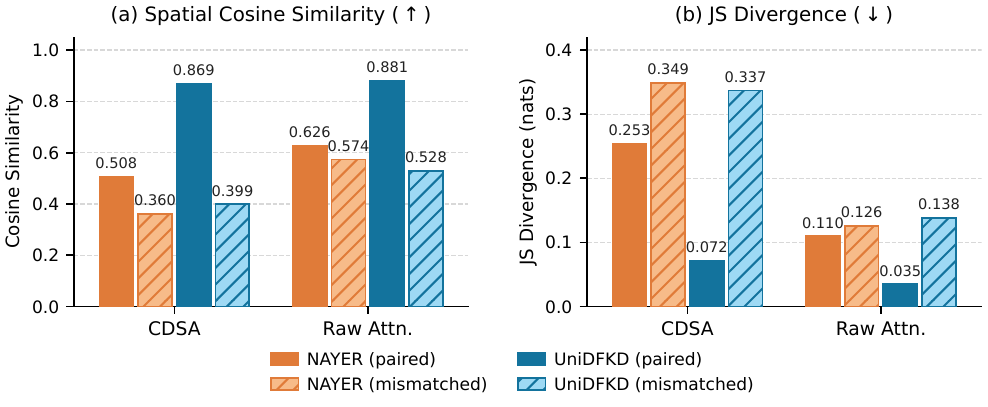}
    \caption{Quantitative teacher--student spatial alignment on 1,000 stratified real CIFAR-10 test images (100 per class, fixed seed, zero excluded samples) under ViT-S$\to$ViT-T distillation.
    Both (a) spatial cosine similarity ($\uparrow$) and (b) Jensen--Shannon divergence in nats ($\downarrow$) are computed on the native $14\times 14$ non-negative maps for the ground-truth class.
    Solid bars denote exact per-image pairing between teacher and student, while hatched bars denote the within-class mismatched control formed by cyclically pairing each teacher map with a student map from another image of the same class.}
    \label{fig:appendix_student_alignment}
\end{figure}

While Tab.~\ref{tab:ablation} of the main paper shows that SSD improves student accuracy, a central question is whether a student distilled exclusively from synthetic images genuinely inherits the teacher's instance-specific spatial reasoning on unseen real test images.
To answer this, we compare the ViT-S teacher against two ViT-T students trained on CIFAR-10 under identical architectures: the strongest logit-only baseline (NAYER, which lacks spatial attribution supervision) and UniDFKD (which adds SSD).
Fig.~\ref{fig:appendix_student_cdsa} presents side-by-side overlays across ten representative real CIFAR-10 test images spanning all ten categories.
Across these examples, NAYER fails to localize the target object reliably and frequently disperses its CDSA and raw attention responses onto background pixels or peripheral textures (e.g., Automobile, Bird, Cat, Deer, Dog, Horse, Ship, and Truck), directly illustrating the spatial shortcut ambiguity established in Sec.~\ref{app:theory_ssd} (Theorem~\ref{thm:ssd-necessity}).
In contrast, UniDFKD consistently concentrates its attribution on the same foreground object parts as the teacher.

Fig.~\ref{fig:appendix_student_alignment} quantifies this alignment across 1,000 stratified real CIFAR-10 test images (100 per class), using the UniDFKD and NAYER checkpoints from Tab.~\ref{tab:main} (93.91\% and 85.86\% test accuracy, respectively).
On native $14\times 14$ target-class CDSA maps, UniDFKD raises spatial cosine similarity from $0.508 \pm 0.148$ to $0.869 \pm 0.064$, outperforming NAYER on 99.8\% ($998/1{,}000$) of test images, while reducing spatial Jensen--Shannon (JS) divergence by more than $3.5\times$ from $0.253 \pm 0.075$ to $0.072 \pm 0.031$ nats.
Notably, although SSD supervises only CDSA maps rather than internal attention weights, this spatial fidelity transfers directly to the student's last-block raw CLS-to-patch attention: UniDFKD boosts attention cosine similarity from $0.626 \pm 0.115$ to $0.881 \pm 0.055$ and lowers JS divergence from $0.110 \pm 0.041$ to $0.035 \pm 0.015$ nats, surpassing NAYER on all $1{,}000/1{,}000$ ($100.0\%$) test samples.

To verify that this agreement reflects genuine per-image correspondence rather than shared class semantics or a common center bias, Fig.~\ref{fig:appendix_student_alignment} also reports a within-class mismatched control that cyclically pairs each teacher map with a student map from a different test image of the same category.
Breaking instance-level pairing causes UniDFKD's alignment to drop precipitously, from $0.869$ to $0.399$ in CDSA cosine similarity (with JS divergence rising to $0.337$ nats) and from $0.881$ to $0.528$ in raw attention cosine similarity (with JS divergence rising to $0.138$ nats).
By contrast, NAYER's paired raw attention sits only marginally above its within-class mismatched baseline ($0.626$ vs.\ $0.574$), confirming that logit-only distillation captures mostly coarse category-level or center-biased layouts.
Overall, both qualitative overlays and population-level alignment metrics confirm that SSD transfers instance-specific spatial attribution structure from teacher to student across the synthetic-to-real gap.

\section{Extended Discussions and Limitations}
\label{app:discussion}

\paragraph{On the spatial prior of SSA.}
The center-biased Gaussian in SSA is a simple prior and may not hold for cluttered scenes, multi-object images, or non-object-centric domains such as pathology, remote sensing, and industrial inspection.
Nevertheless, the contribution of SSA is not tied to this particular form.
Our analysis suggests that the effectiveness of BN priors in DFKD primarily comes from constraining semantic distributions rather than from normalization itself, while CDSA provides an architecture-agnostic interface for imposing such constraints.
The Gaussian is therefore only an instantiation suited to object-centric benchmarks.
Other domains may adopt a flat prior, a multi-modal prior, or a prior estimated from teacher attribution statistics.
Moreover, SSA is applied as a batch-level regularizer weighted by $\lambda_{\mathrm{SSA}}$, rather than as a hard per-sample constraint, and its performance remains stable over a broad range of prior parameters.
We thus regard the semantic prior interface, rather than the center-biased Gaussian, as the transferable component.

\paragraph{Relation to BN-free model inversion and data-free quantization.}
BN-free model inversion has also been explored in data-free quantization, often through transformer-specific objectives based on patch similarity or attention~\cite{li2022patch,ramachandran2024clamp,zhong2024semantics}.
These methods differ from ours in two aspects.
First, their priors depend on transformer-internal quantities and generally do not transfer to convolutional teachers, whereas CDSA abstracts a shared class-discriminative spatial response and therefore supports both BN- and LN-based architectures.
Second, quantization calibration mainly requires representative activation ranges, while DFKD must recover sufficiently rich decision knowledge to train a student from random initialization.
Priors effective for calibration are therefore not necessarily sufficient for distillation.
Our work focuses on the stronger requirement of constructing an architecture-agnostic prior for knowledge transfer.

\paragraph{Relation to diffusion-based synthesis.}
Our language-conditioned synthesis is related to methods that generate training data with text-to-image diffusion models~\cite{li2024diverse,qi2025data,he2025prism,li2024genq}, but the two settings differ in their information assumptions.
We consider an auxiliary module data-free only when it cannot independently generate target-class images.
A frozen text encoder satisfies this criterion because it provides only categorical geometry, while the image evidence must still be recovered from the teacher.
In contrast, a pretrained diffusion model already contains a strong natural-image prior and can generate class-conditional samples with limited dependence on the teacher.
Such approaches are valid, but effectively transfer knowledge from an external generative model under a stronger assumption than standard DFKD.
We further show that replacing the vision--language encoder with a mean-centered, vision-agnostic language encoder matches and even slightly surpasses the vision--language encoder (Sec.~\ref{subsec:textenc}), indicating that CSC mainly contributes categorical structure rather than external visual knowledge.

\paragraph{On the data-free setting.}
DFKD prohibits access to the training data of the teacher.
CSC satisfies this constraint in the same way as the label-text embedding of NAYER~\cite{tran2024nayer}: the semantic bank is built once from class names by frozen, publicly available language models and is stored before training.
No language model is queried during synthesis or distillation, none of them is fine-tuned, and no real or external image is involved at any stage.
All image evidence is recovered from the teacher.

%% file: fig/theory_schematic.tex
\definecolor{thBlueL}{HTML}{E3F4FC}
\definecolor{thBlue}{HTML}{9ED9F4}
\definecolor{thBlueD}{HTML}{13739D}
\definecolor{thOrL}{HTML}{FDECDF}
\definecolor{thOr}{HTML}{F7BB89}
\definecolor{thOrD}{HTML}{E07B39}
\definecolor{thRedL}{HTML}{FCE9E7}
\definecolor{thRedD}{HTML}{D65B52}
\definecolor{thGrL}{HTML}{C7FFD4}
\definecolor{thGrD}{HTML}{2E9E5B}
\definecolor{thSlate}{HTML}{343C49}
\definecolor{infA}{HTML}{000004}
\definecolor{infB}{HTML}{420A68}
\definecolor{infC}{HTML}{932667}
\definecolor{infD}{HTML}{DD513A}
\definecolor{infE}{HTML}{FCA50A}
\definecolor{infF}{HTML}{FCFFA4}
\tikzset{
  th/.style={font=\fontfamily{ComicNeue-TLF}\selectfont\scriptsize, line width=0.7pt, line cap=round, line join=round},
  sketch/.style={decorate, decoration={random steps, segment length=3.5pt, amplitude=0.3pt}, rounded corners=1.2pt},
  tharrow/.style={-{Latex[length=1.7mm, width=1.5mm]}, line width=0.9pt},
  thok/.style={circle, fill=thGrD, text=white, inner sep=0.5pt, font=\tiny},
  thbad/.style={circle, fill=thRedD, text=white, inner sep=0.5pt, font=\tiny},
  thcell/.style={minimum size=#1, inner sep=0pt, outer sep=0pt, draw=white, line width=0.3pt},
}
\begin{figure*}[t]
\centering
\pgfmathsetseed{11}%
\begin{subfigure}[t]{0.325\linewidth}
\centering
\begin{tikzpicture}[th]
  \fill[thOrL] (2.35,0.15) rectangle (5.25,3.25);
  \draw[thOrD, dashed, line width=0.7pt] (2.35,0.15) -- (2.35,3.25);
  \draw[sketch, draw=thBlueD, fill=thBlueL] (1.5,1.7) circle (1.25);
  \draw[draw=thGrD, fill=thGrL, line width=0.8pt] (2.35,0.7835) arc[start angle=-47.16, end angle=47.16, radius=1.25] -- cycle;
  \draw[thSlate, line width=0.5pt, -{Latex[length=1.3mm]}] (0.1,1.7) -- (5.3,1.7) node[above left, inner sep=1pt, text=thSlate] {$\mathring{\mathbf{w}}_y$};
  \draw[thSlate, line width=0.5pt, -{Latex[length=1.3mm]}] (1.5,0.15) -- (1.5,3.3) node[above, inner sep=1pt, text=thSlate] {kernel of $\mathring{\mathbf{W}}^{\top}$};
  \fill[thRedD] (2.62,1.7) circle (1.7pt);
  \node[anchor=west, inner sep=1pt, text=thRedD] at (2.82,1.92) {$\alpha\mathring{\mathbf{w}}_y$};
  \node[align=center, text=thBlueD] at (0.97,1.2) {variance\\cap\\[-1pt]{\tiny($\mathcal{L}_{\mathrm{stat}}$)}};
  \node[align=center, text=thOrD] at (3.72,2.83) {$\mathcal{L}_{\mathrm{CE}}$ needs\\$\langle\mathring{\mathbf{w}}_y,\tilde{\mathbf{z}}\rangle\ge\psi^{-1}(\mathcal{L}_{\mathrm{CE}})$};
  \draw[thOrD, dashed, line width=0.6pt, {Latex[length=1.2mm]}-{Latex[length=1.2mm]}] (4.6,0.25) -- (4.6,2.4) node[pos=0.32, sloped, below, inner sep=1.5pt] {free for $\mathcal{L}_{\mathrm{CE}}$};
  \node[align=center, text=thGrD] (conc) at (3.62,0.62) {synthetic features\\concentrate here};
  \draw[thGrD, line width=0.6pt, -{Latex[length=1.2mm]}] (conc.north west) ++(0.1,-0.02) -- (2.58,1.2);
\end{tikzpicture}
\caption{Deep layer (Sec.~\ref{app:theory}, Prop.~\ref{prop:deep-concentration}).}
\label{fig:theory_schematic_a}
\end{subfigure}\hfill
\begin{subfigure}[t]{0.325\linewidth}
\centering
\begin{tikzpicture}[th]
  \draw[thSlate, line width=0.5pt, -{Latex[length=1.3mm]}] (0.4,0.45) -- (5.3,0.45);
  \node[anchor=north east, inner sep=1pt, text=thSlate] at (5.3,0.19) {``What'' $\mathcal{A}(\mathbf{M})$};
  \draw[thSlate, line width=0.5pt, -{Latex[length=1.3mm]}] (0.4,0.45) -- (0.4,3.3) node[right, inner sep=2pt, text=thSlate] {``Where'' $\mathcal{P}_{\mathrm{where}}\mathbf{M}$};
  \foreach \x in {1.15,2.25,3.35}{
    \draw[thSlate, dashed, line width=0.7pt] (\x-0.3,0.28) -- (\x+0.3,0.28);
    \draw[thSlate, line width=0.6pt] (\x-0.3,0.22) -- (\x-0.3,0.34) (\x+0.3,0.22) -- (\x+0.3,0.34);}
  \node[anchor=north, inner sep=1pt, text=thSlate] at (2.25,0.19) {$K$ real modes};
  \draw[thOrD, dashed, line width=0.7pt] (0.4,2.05) -- (5.2,2.05);
  \foreach \i/\j/\c in {0/0/infB,1/0/infC,2/0/infB,0/1/infC,1/1/infF,2/1/infC,0/2/infB,1/2/infC,2/2/infB}{
    \node[thcell=0.16cm, fill=\c] at (4.43+0.16*\i,2.9-0.16*\j) {};}
  \draw[thOrD, line width=0.6pt, -{Latex[length=1.1mm]}] (4.59,2.48) -- (4.59,2.12);
  \node[align=right, anchor=east, text=thOrD] at (4.3,2.74) {SSA: batch-mean\\map $\propto\boldsymbol{\pi}$};
  \foreach \x in {1.15,2.25,3.35}{
    \draw[sketch, draw=thGrD, fill=thGrL] (\x,2.05) ellipse (0.3 and 0.2);}
  \node[align=left, anchor=west, text=thGrD] at (0.5,1.52) {CSC: green modes\\cover the real modes};
  \draw[thRedD, dashed, line width=0.6pt, {Latex[length=1.2mm]}-{Latex[length=1.2mm]}] (4.85,0.58) -- (4.85,1.72);
  \fill[thRedD] (4.85,0.85) circle (1.7pt);
  \node[align=right, anchor=east, text=thRedD] at (4.72,0.8) {$\mathcal{L}_{\mathrm{CE}}$ alone: one point,\\``Where'' left free};
\end{tikzpicture}
\caption{What/Where (Sec.~\ref{app:theory_what_where}, Props.~\ref{prop:ce-degeneracy}--\ref{prop:csc-what}).}
\label{fig:theory_schematic_b}
\end{subfigure}\hfill
\begin{subfigure}[t]{0.325\linewidth}
\centering
\begin{tikzpicture}[th]
  \foreach \i in {0,1,2}{\foreach \j in {0,1,2}{
    \draw[fill=thBlue, draw=thBlueD, line width=0.5pt] (0.05+0.4*\i,1.25+0.4*\j) rectangle ++(0.4,0.4);}}
  \draw[fill=thOr, draw=thOrD, line width=0.6pt] (0.45,1.65) rectangle ++(0.4,0.4);
  \node[text=thSlate] at (0.65,1.85) {obj};
  \node[align=center, text=thSlate] at (0.65,2.68) {synthetic image};
  \node[align=center, text=thBlueD] at (0.65,0.92) {background\\shows the class};
  \draw[sketch, draw=thRedD, fill=thRedL] (2.05,2.2) rectangle (5.3,3.3);
  \node[align=left, anchor=west, text=thSlate] at (2.12,2.75) {KL only: $\bar{\mathbf{f}}\parallel\bar{\mathbf{g}}$, $\kappa_{\mathrm{pool}}=\infty$\\follows the background\\real accuracy $1/C$};
  \node[thbad] at (5.1,2.4) {\ding{55}};
  \draw[tharrow, thSlate] (1.3,2.2) -- (2.0,2.62) node[midway, sloped, above, inner sep=1pt] {pool};
  \draw[sketch, draw=thGrD, fill=thGrL] (2.05,0.05) rectangle (5.3,1.95);
  \foreach \m/\x in {teacher/2.2,student/3.05}{
    \foreach \i in {0,1,2}{\foreach \j in {0,1,2}{
      \pgfmathsetmacro{\isobj}{(\i==1 && \j==1) ? 1 : 0}
      \ifdim\isobj pt>0.5pt \def\cc{infF}\else\def\cc{infB}\fi
      \node[thcell=0.2cm, fill=\cc] at (\x+0.1+0.2*\i,1.62-0.2*\j) {};}}
    \node[text=thSlate] at (\x+0.3,0.96) {\m};}
  \node[text=thSlate] at (2.925,1.42) {$\approx$};
  \node[align=left, anchor=west, text=thSlate] at (3.72,1.38) {$+$SSD:\\$\kappa_{\mathrm{patch}}=1$\\$\mathbf{W}_{\mathrm{bg}}=\mathbf{0}$};
  \node[align=center, text=thGrD] at (3.67,0.42) {evidence on the object patch\\real accuracy $1$};
  \node[thok] at (5.12,1.77) {\ding{51}};
  \draw[tharrow, thGrD] (1.3,1.35) -- (2.0,1.1) node[midway, sloped, above, inner sep=1pt] {patches};
\end{tikzpicture}
\caption{SSD (Sec.~\ref{app:theory_ssd}, Thms.~\ref{thm:ssd-bound}--\ref{thm:ssd-necessity}).}
\label{fig:theory_schematic_c}
\end{subfigure}
\vspace{-3pt}
\caption{The three theoretical analyses at a glance.
(a)~At the deepest layer, let $\tilde{\mathbf{z}}$ be the synthetic feature measured from the real feature mean.
A classification loss $\mathcal{L}_{\mathrm{CE}}$ requires the alignment $\langle\mathring{\mathbf{w}}_y,\tilde{\mathbf{z}}\rangle$ with the class direction $\mathring{\mathbf{w}}_y$ to reach $\psi^{-1}(\mathcal{L}_{\mathrm{CE}})$ on average, the smallest target score that this loss allows, and it leaves the kernel of the centered head $\mathring{\mathbf{W}}^\top$ free.
The variance cap from the channel statistics bounds every direction, so $\tilde{\mathbf{z}}$ concentrates near the prototype $\alpha\mathring{\mathbf{w}}_y$.
(b)~Each attribution map $\mathbf{M}$ splits into its spatial mean $\mathcal{A}(\mathbf{M})$, the ``What'' component that the classifier reads, and its zero-mean spatial pattern $\mathcal{P}_{\mathrm{where}}\mathbf{M}$, the ``Where'' component.
The classification loss collapses the centered logits of each class, the part that Softmax reads, to one point, and it leaves the ``Where'' component free.
CSC draws the synthetic images from $K$ semantic clusters, which lets each cluster match its own mode of the real centered logits, and SSA aligns the batch-averaged map with the center-biased template $\boldsymbol{\pi}$ up to scale, which gives its ``Where'' component a nonzero layout.
(c)~When the background of every synthetic image shows the class, the pooled object and background features are collinear ($\bar{\mathbf{f}}\parallel\bar{\mathbf{g}}$), so the pooled features cannot separate them (shift factor $\kappa_{\mathrm{pool}}=\infty$), and KL-only distillation follows the background.
The patch-level maps separate them ($\kappa_{\mathrm{patch}}=1$), and SSD sets the background weights $\mathbf{W}_{\mathrm{bg}}$ to zero, which places the evidence on the object patch.}
\label{fig:theory_schematic}
\end{figure*}